\documentclass{article}

\usepackage{arxiv}

\usepackage[utf8]{inputenc}
\usepackage[T1]{fontenc}
\usepackage{hyperref}
\usepackage{xurl}
\usepackage{booktabs}
\usepackage{amsmath}
\usepackage{amssymb}
\usepackage{amsfonts}
\usepackage{dsfont}
\usepackage{latexsym}
\usepackage{microtype}
\usepackage{graphicx}
\usepackage{natbib}
\usepackage{doi}
\usepackage{times}
\usepackage{inconsolata}
\usepackage{multirow}
\usepackage{array}
\usepackage{makecell}
\usepackage{xcolor}
\usepackage{caption}
\usepackage{subcaption}
\usepackage{fvextra}
\usepackage{placeins}
\usepackage{needspace}
\DefineVerbatimEnvironment{Artifact}{Verbatim}{fontsize=\small,
  breaklines=true,breakanywhere=true,breakautoindent=true,
  breaksymbolleft={},frame=leftline,rulecolor=\color{gray},
  xleftmargin=1em,xrightmargin=0.5em,framesep=0.7em}

\graphicspath{{figures/}}

\newcolumntype{R}[1]{>{\raggedleft\arraybackslash}p{#1}}
\newcommand{\up}{$\uparrow$}
\newcommand{\down}{$\downarrow$}

\title{MUSE: A Theory-Harnessed Story Engine for Vibe Narrativizing}
\author{%
  \begin{minipage}{0.97\textwidth}
  \centering
  {\normalsize\bfseries
  Jianxiang Ma\textsuperscript{1,2}\quad
  Xiaocui Yang\textsuperscript{1}\quad
  Daling Wang\textsuperscript{1}\quad
  Mingfu Zhang\textsuperscript{2,3,*}\\[4pt]
  Zijing Wang\textsuperscript{1}\quad
  Yuesong Hou\textsuperscript{1}\quad
  Yichen Gao\textsuperscript{1}\quad
  Junzhao Huang\textsuperscript{1}\endgraf}
  \vspace{7pt}
  {\small\normalfont
  \textsuperscript{1}School of Computer Science and Engineering, Northeastern University,\\
  Shenyang 110819, China\\[3pt]
  \textsuperscript{2}OranAI, Shenzhen 518057, China\\[3pt]
  \textsuperscript{3}OranAI Ltd., City of Industry, CA 91748, USA\endgraf}
  \vspace{6pt}
  {\footnotesize\normalfont
  Jianxiang Ma: \href{mailto:jianxiangma020518@gmail.com}{\nolinkurl{jianxiangma020518@gmail.com}}\\
  Yuesong Hou: \href{mailto:yuesonhou3-c@my.cityu.edu.hk}{\nolinkurl{yuesonhou3-c@my.cityu.edu.hk}}\\
  \textsuperscript{*}Corresponding author: Mingfu Zhang,
  \href{mailto:cto@oran.cn}{\nolinkurl{cto@oran.cn}}\endgraf}
  \end{minipage}
}
\date{October 13, 2026}

\renewcommand{\shorttitle}{MUSE: A Theory-Harnessed Story Engine}

\hypersetup{
pdftitle={MUSE: A Theory-Harnessed Story Engine for Vibe Narrativizing},
pdfsubject={Story generation, agent harnesses, narrative theory},
pdfauthor={Jianxiang Ma, Xiaocui Yang, Daling Wang, Mingfu Zhang, Zijing Wang, Yuesong Hou, Yichen Gao, Junzhao Huang},
pdfkeywords={story generation, agent harness, narrative theory, computational creativity, human--AI co-writing},
}

\begin{document}
\maketitle

\begin{abstract}
LLMs have been able to generate fluent prose, but creating a high-quality story also requires coordinated decisions about plot, character, and language across planning, drafting, and revision. Guiding these decisions presents two bottlenecks: rule quality, whether story knowledge gives the model clear guidance, and rule realization, whether that guidance and the decisions it informs continue to shape the work. We formulate Vibe Narrativizing as the task of turning natural-language writing requirements into a finished story and present MUSE, a Theory-Harnessed Story Engine that addresses both bottlenecks by combining story theory with a practical agent harness. Knowledge engineering organizes Robert McKee's story theory through rule atomization, semantic consolidation, mechanism abstraction, a single source of truth, and layered disclosure, with typical examples for principles that depend on context and aesthetic judgment. The harness organizes design, character performance, scene composition, and revision around intermediate deliverables that preserve story decisions; context engineering supplies each role with the guidance and decisions its task requires, and a masterwork corpus provides inspiration and prose references. A worked example follows one requested object from its thematic role to the characters' climactic actions. Across four base models, MUSE improves WritingBench by 1.1--6.2 points over zero-shot generation; it is the only multi-stage system in our comparison to do so. It also raises LongStoryEval by more than ten points on three of the four models. ConStory-Bench consistency error density remains in the low single digits for all four models, below every reproduced story-system baseline on three of the four models. Component ablations locate the largest quality contribution in structural design, voice-specific effects in the character path, and further gains in revision. Code is available at \url{https://github.com/RoadtoAGI/MUSE}.
\end{abstract}

\section{Introduction}

Story generation brings together problems of computational creativity and practical support for human writers. A writing request can specify a genre, character relationships, required events, and a desired prose style; the resulting work must fulfill these requirements in a coherent reading experience~\citep{colton2012computational,mirowski2023cowriting}. Large language models (LLMs) can turn such requests into fluent prose. This prose can still contain generic characterization, weakly connected events, and language that does not fit the intended work. A conspicuous form of this mismatch is the migration of software-engineering terminology and explanatory habits into other kinds of writing. In creative prose, such language can replace concrete action and perception with abstract descriptions, weakening readability and disrupting the story's register.

The quality of characterization, plot development, and prose depends on how creative decisions work together. A character's desire shapes their actions under pressure; the resulting consequences change the situation in the next scene. Viewpoint, dialogue, and detail determine how the reader understands those changes. Story theory and masterwork analysis connect these decisions to their narrative effects, providing a basis for guidance throughout creation.

The first bottleneck is \emph{rule quality}: story knowledge needs to provide clear, appropriate guidance for the decisions a model must make. Principles such as developing subtext, revealing character under pressure, or expressing a theme through a climax describe desirable effects at different narrative scales. To act on them, the model needs to understand what each principle asks it to decide, how related requirements fit together, and which guidance applies to its current task. Clear structure, consistent meanings, and explicit requirements make that guidance easier to understand and follow. For judgments that depend on context and aesthetic purpose, typical examples supply distinctions that an abstract instruction leaves implicit.

The second bottleneck is \emph{rule realization}: the guidance and the decisions it informs must continue to shape the work as creation proceeds. Recent systems generate stories in multiple stages, using recursive planning and revision, specialized agents, or playwriting knowledge~\citep{yang2022re3,yang2023doc,huot2024agentsroom,chen2024hollmwood,wu2025playwriting}. Each stage produces decisions that later stages need: an outline establishes a turn, a character design establishes a way of speaking, and a scene plan establishes the consequence of an action. Later writing and revision calls must recover these decisions from the context they receive. When an outline turn is buried in accumulated prose, a voice constraint is omitted, or an action's consequence is absent from the next scene plan, structure drifts, voices converge, and consequences disappear between otherwise fluent scenes. Sustained guidance requires both a record of what has been decided and an organized way to use that record in subsequent work.

\begin{figure}[t]
\centering
\includegraphics[width=0.95\textwidth]{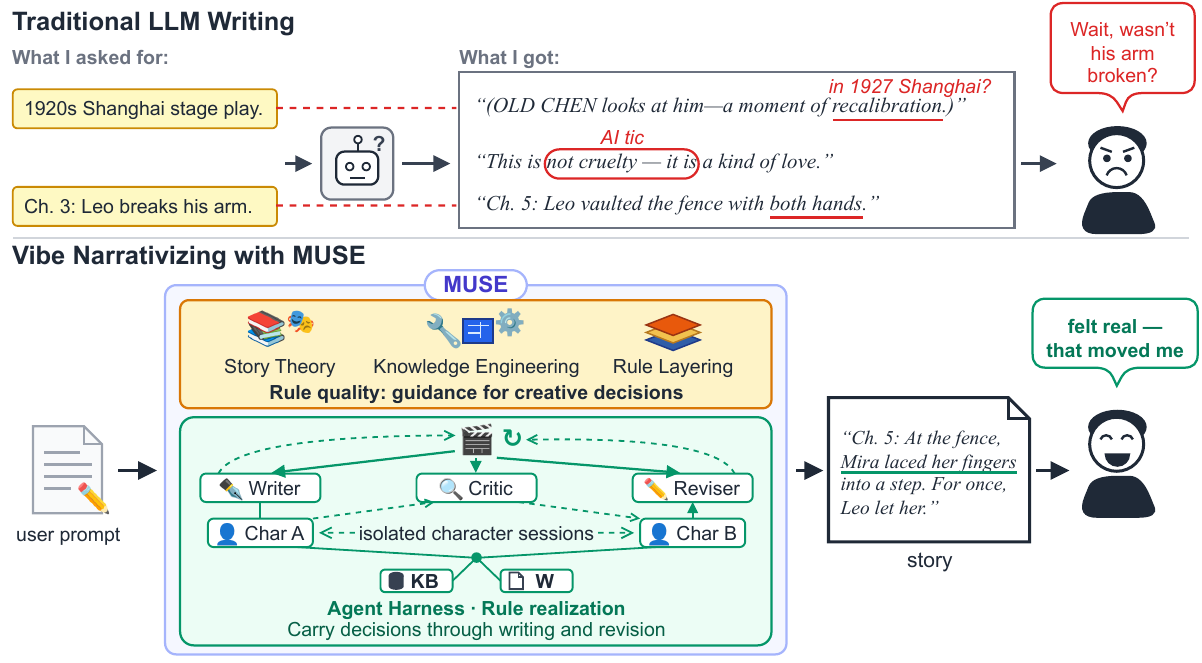}
\caption{MUSE motivation and overview. The upper panel illustrates prompt-only story generation. The lower panel shows how MUSE addresses rule quality and rule realization together: knowledge engineering organizes story theory into layered guidance for creative decisions, and an agent harness sustains the use of this guidance and the resulting decisions through writing, character performance, critique, and revision. KB and W denote the knowledge base and world state.}
\label{fig:motivation}
\end{figure}

MUSE addresses both bottlenecks through theory-harnessed story generation (Figure~\ref{fig:motivation}): story theory supplies the rules, and a practical agent harness puts them to work. Following the knowledge-engineering tradition of acquiring, representing, and applying expert knowledge~\citep{feigenbaum1977art}, we organize McKee's story theory~\citep{mckee1997story} into rules through atomization, semantic consolidation, and mechanism abstraction. A single source of truth and layered disclosure determine where each rule is maintained and when it is disclosed; typical examples clarify their use when decisions depend on context and aesthetic judgment. The harness structures the creative process: stages, roles, tools, persistent story decisions, and feedback for revision. Context engineering determines which rules and recorded decisions enter each model call and in what form. A scene writer, for example, receives the planned turn, the participating characters' intentions, the preceding scene's state, and a relevant prose reference. These materials guide the actions, dialogue, and narration that develop the turn; review then compares the resulting scene with its design to guide revision.

We call the task of turning a finite set of natural-language writing requirements into a finished story \emph{Vibe Narrativizing}. The name follows vibe coding: a user expresses an intent in natural language, and the system carries out the production. Here, that production is a story. MUSE produces the story together with intermediate records of its conception, design, and revision. We evaluate the final outputs using WritingBench~\citep{wu2025writingbench}, LongStoryEval~\citep{yang2025longstoryeval}, and ConStory-Bench~\citep{li2026constory} across Claude Sonnet 4.6, Claude Opus 4.7, GPT-5.4, and GPT-5.5. MUSE improves WritingBench by 1.1--6.2 points over the corresponding zero-shot models and raises LongStoryEval by more than ten points on three of them. Its ConStory-Bench consistency error density remains in the low single digits, below every reproduced story-system baseline on three models and roughly an order of magnitude below them on Sonnet and Opus.

This paper makes three contributions. \textbf{(1) Task formulation.} We define Vibe Narrativizing through intent reflection, beat fidelity, voice differentiability, and narrative coherence. \textbf{(2) Method.} We introduce MUSE, a theory-harnessed story engine. Its knowledge engineering organizes story theory into rules and examples for specific creative decisions; its five-module harness and context engineering carry that guidance and the resulting decisions through planning, performance, writing, and revision. \textbf{(3) Evaluation.} We report final-output scores and component ablations across four base models, together with a worked example that follows a writing requirement into the actions of the completed story.

\section{Related Work}

\paragraph{Story knowledge and narrative planning.}
Computational narratology connects story structure to causal progression and character behavior through plot grammars~\citep{propp1968morphology}, narrative planning~\citep{riedl2010narrativeplanning}, learned event representations~\citep{martin2018event}, and emotional arcs~\citep{reagan2016emotional}. Dramatron~\citep{mirowski2023cowriting} uses a hierarchical story grammar for co-writing with industry professionals. Playwriting-guided generation~\citep{wu2025playwriting} incorporates craft knowledge and reflection into interactive drama. These approaches make narrative knowledge available as structures or instructions for generation. Knowledge engineering examines how expert principles can be represented and organized to guide a system's decisions~\citep{feigenbaum1977art}. MUSE applies this perspective to McKee's account of scene change, dialogue, character, and climax.

LLM planning systems preserve plot over longer spans. Re$^3$~\citep{yang2022re3} recursively plans, drafts, and revises; DOC~\citep{yang2023doc} strengthens detailed outline control; and DOME~\citep{wang2024dome} combines dynamic hierarchical outlining with memory. StoryWriter~\citep{wang2025storywriter}, SuperWriter~\citep{wu2024superwriter}, and heterogeneous recursive planning~\citep{xiong2025heterogeneous} further develop planning, refinement, and retrieval. MUSE builds on this progression by relating each level of design to the creative decisions it prepares: a controlling idea shapes the climax, the climax shapes the preceding conflicts, and scene plans guide action and dialogue.

\paragraph{Agent collaboration and the creative environment.}
Agents' Room~\citep{huot2024agentsroom} assigns specialized agents to plot, character, and scene writing. HoLLMwood~\citep{chen2024hollmwood} organizes screen-oriented narratives through roles such as writer, director, and actor. Persistent character agents use persona descriptions, memory, and role-specific context to maintain behavior~\citep{park2023generative,shao2023characterllm,wang2023rolellm,zhou2024characterglm,ran2025bookworld}. MUSE combines specialized writing roles with character performance: an isolated call develops one character's possible actions and lines, and a writer composes the scene from the resulting materials. This handoff requires an environment that organizes tools, shared knowledge, and feedback around each role's work, as developed in research on agent--environment interfaces~\citep{yang2024sweagent} and harness engineering~\citep{lopopolo2026harness}. MUSE preserves the materials exchanged between roles in intermediate deliverables.

\paragraph{Examples and context engineering.}
Dense retrieval~\citep{karpukhin2020dense} and in-context example selection~\citep{liu2022incontext,rubin2022learning,min2022rethinking} provide generation-time guidance through relevant demonstrations. Context engineering extends this concern to the selection and maintenance of the information available throughout an agent's work~\citep{anthropic2025context}. MUSE uses examples for three purposes: teaching rules whose application depends on context, supplying plot inspiration during Design, and providing a scene's prose register during Creation. Its knowledge organization supports layered disclosure, while its runtime context combines the current guidance with persistent story decisions and task-specific references.

\paragraph{Evaluation of generated stories.}
WritingBench~\citep{wu2025writingbench} scores query-specific writing criteria, LongStoryEval~\citep{yang2025longstoryeval} applies a craft rubric, and ConStory-Bench~\citep{li2026constory} measures consistency errors. LLM-based evaluation provides structured judgments of generated text~\citep{liu2023geval,zheng2023judging}.

\section{Method}
\label{sec:method}

\begin{figure}[t]
\centering
\includegraphics[width=1.0\textwidth]{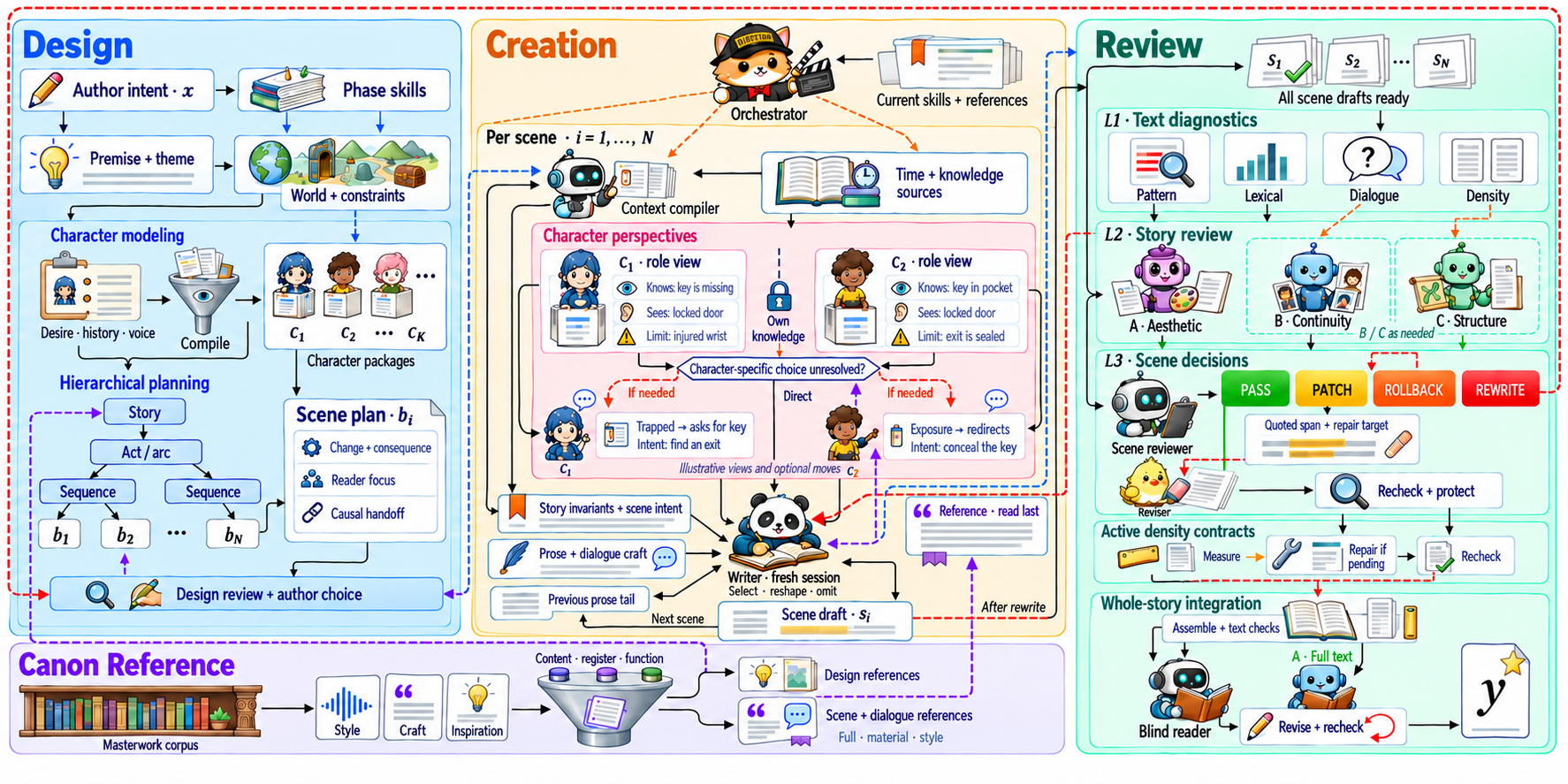}
\caption{MUSE execution flow. Design (blue) develops character packages and hierarchical scene plans. Creation (orange) assembles scene context, character perspectives, and performance material for a writer starting in a fresh session. Canon Reference (purple) supplies design inspiration and scene references. Review (green) combines text diagnostics, story review, scene-level revision, and whole-story integration; feedback returns to the relevant writing or design stage.}
\label{fig:pipeline}
\end{figure}

MUSE separates two kinds of material. Skills and references hold reusable story guidance; intermediate artifacts hold what a particular story has decided: its premise, world, characters, structure, drafts, and feedback. The five modules in Figure~\ref{fig:pipeline} produce and use these artifacts so that each stage develops the preceding work. We first formulate the task, then describe how the guidance is developed for rule quality and how the harness applies it for rule realization.

\subsection{Problem Formulation for Vibe Narrativizing}
\label{sec:problem}

Let $x$ denote a prompt with finite requirements, such as setting, character relations, required events, and prose style. A run $f$ maps $x$ to an ordered trace $\mathcal{A}=(a_0,\ldots,a_7)$ and a final story $y=(s_1,\ldots,s_N)$ of $N$ scenes: $f:x\mapsto(\mathcal{A},y)$. An \emph{artifact} is an intermediate file in this trace, such as a world description, character profile, scene plan, or review verdict. The trace preserves decisions at successive stages so later calls can develop the same story.

The scene plans in this trace use a \emph{beat contract} to specify the intended value transition and the pressure that drives it:
\begin{equation}
b_i=(v_i^{-},\,v_i^{+},\,p_i).
\label{eq:beat}
\end{equation}
Here $v_i^{-},v_i^{+}\in\mathcal{V}$ are the states entering and leaving scene $s_i$, and $p_i$ is the pressure that forces the transition. Narrative values in $\mathcal{V}$ include trust, safety, and betrayal. The map $\sigma:\mathcal{V}\rightarrow\{-,0,+\}$ expresses their negative, neutral, or positive charge in the scene. A forced choice, for example, can move a relationship from trust ($+$) to betrayal ($-$). Each $b_i$ records the value turn that the beats of scene $s_i$ build together; Section~\ref{sec:schema} defines the beat itself.

Character profiles record each character's motives, relationships, and planned development. Character packages $\mathcal{C}=\{c_1,\ldots,c_K\}$ distill these designs into guidance on desire, voice, behavior, and prohibitions for performance and writing. The rules used to produce artifact $a_t$ are denoted by $\mathcal{R}_t$.

A run is acceptable when its trace and story jointly satisfy four requirements:
\begin{equation}
\begin{aligned}
Q(y,\mathcal{A};x)={}&I(x,\mathcal{A},y)\,\wedge\,B(\{b_i,s_i\})\\
&\wedge\,V(\mathcal{C},y)\,\wedge\,H(y).
\end{aligned}
\label{eq:vibe-quality}
\end{equation}
\emph{Intent reflection} $I$ requires the intermediate decisions and finished story to reflect the prompt's requirements. \emph{Beat fidelity} $B$ requires each scene to realize its planned transition~\citep{wang2023pacing}. \emph{Voice differentiability} $V$ requires characters to remain distinct in word choice and behavior. \emph{Narrative coherence} $H$ requires plot, character, and world facts to remain continuous across scenes. These requirements guide design, composition, and review: they determine what to preserve in the artifacts and what to examine in the prose. The final-output evaluation in Section~\ref{sec:experiments} covers query satisfaction, craft, and consistency.

\subsection{Story Theory Knowledge Engineering}
\label{sec:schema}

Rule quality begins with how story knowledge is organized. We organize McKee's framework around four concepts \citep{mckee1997story}. A \emph{Beat} is a continuing action--reaction relationship; a change in that relationship begins a new beat, and connected beats develop a scene's value change. \emph{Dialogue} connects what a character says to what is withheld or cannot be acknowledged. A \emph{Character Arc} concerns what choices under pressure reveal or change about the character. The \emph{Plot Peak} is the climactic action that expresses the controlling idea, the story's central claim about how and why a value changes. Together, these concepts connect local exchanges, character development, and the meaning of the whole story.

MUSE develops McKee's principles into rules and organizes their use through five operations. \emph{Rule atomization} identifies the complete decision a compound principle asks for. Related rules are grouped by the narrative function or mechanism they support. \emph{Semantic consolidation} merges formulations that prescribe the same decision under the same conditions and responsibilities, preserving their requirements and permitted choices, so the model meets one requirement where it would otherwise meet several near-duplicates. \emph{Mechanism abstraction} identifies the relationship through which more specific rules achieve an effect; scale-specific responsibilities and alternative methods remain attached to that relationship. This organization adapts goal refinement and responsibility assignment from requirements engineering to narrative decisions~\citep{vanlamsweerde2001goal}. A \emph{single source of truth} (SSOT) gives each shared rule one authoritative location, and \emph{layered disclosure} makes task guidance and deeper references available where they are used. Atomization clarifies what a principle asks the model to decide; consolidation and abstraction show how related requirements fit together; the single source of truth and layered disclosure determine which guidance applies to the current task.

For example, conflict connects a character's pursuit to an opposing response and its consequences. When pursuit organizes the narrative, the resulting rule requires consequential responses to change the character's strategies, costs, or available choices, thereby supplying conditions for what happens next. Plot-spine design develops the pursuit and its major turns; sequence and scene design determine local changes and their consequences; the writer chooses the actions, dialogue, and narrative order that realize them. The shared mechanism supports different implementations while preserving the causal relationship. This separation of an intended effect from its realization also applies when evidence changes an interpretation or when recurrence and variation develop a motif. Appendix~\ref{app:rule-engineering} traces this conflict abstraction and uses Why, What, When, and How (WWWH) to show how objectives, conditions, and methods are organized in the rule library.

Typical examples supply the contextual distinctions needed to use such guidance. A quiet line after an argument can simultaneously convey care and withhold an apology; its subtext depends on the relationship, the preceding conflict, and the speaker's desired response. The rule library therefore pairs abstract guidance with short cases that connect an utterance or action to its narrative function. Canon craft analyses add sustained examples. In a night battle from Jin Yong, five finger-shaped holes in a skull make a threat tangible through touch; concrete wounds and their accumulating consequences then carry the changing balance of the fight (Appendix~\ref{app:canon-corpus}). By showing how an action or detail produces an effect, these examples give the model distinctions to apply when composing new material \citep{min2022rethinking}.

The guidance also determines which story decisions should be saved. Scene plans record the conflict or organizing relationship, entering and leaving states, and consequences for the next scene; character designs record desires, contradictions, and voice; climax designs connect a decisive action to the controlling idea. Review reads these records alongside the resulting text. Appendix~\ref{app:schema} specifies the representations and associated narrative judgments. These records make decisions available to later work, while the rules explain how to develop and assess those decisions.

\subsection{Story-Generation Agent Harness}
\label{sec:harness}

Rule realization depends on how the creative process is organized. The harness builds that process around a system of deliverables. Design establishes the story's commitments; Performance develops material from individual characters' perspectives; Creation composes scenes; Review directs revision and integration. Canon Reference supplies material for both structural and prose decisions. The orchestrator schedules these roles, provides tools and access to the relevant files, and maintains their handoffs. The persistent artifacts let one role build on another's work: a scene writer can use a character's established motive, and a reviewer can compare the resulting choice with the intended scene turn.

\paragraph{Design.}
Design builds six artifacts in sequence. Phases 0--5 develop the premise and controlling idea, world facts, character profiles, plot spine and climax, sequence structure, and scene plans. Each phase uses the prompt, earlier decisions, and its current rule bundle:
\begin{equation}
a_t=G_t\!\left(x,\,a_{<t},\,\mathcal{R}_t\right),\qquad t=0,\ldots,5,
\label{eq:phase-recurrence}
\end{equation}
where $a_{<t}=(a_0,\ldots,a_{t-1})$. Successive artifacts refine these decisions for scene writing: the climax determines what preceding conflicts must establish, and scene arrangement specifies the actions and consequences that make that progression possible. The Beat Tree organizes this development across four narrative scales:
\begin{equation}
\mathcal{T}:\quad \text{story}\;\longrightarrow\;\text{act}\;\longrightarrow\;\text{sequence}\;\longrightarrow\;\text{scene}.
\label{eq:beat-tree}
\end{equation}
Each level develops a consequential change at its own scale. The representations follow their respective tasks: plot-spine artifacts record the central conflict and climax; sequences record conflict, escalation, and a sequence climax; scene plans record entering and leaving values and the causal links between scenes. The tree's leaves are scenes; at each leaf, the writer develops the turn $b_i$ through a sequence of beats. Appendix~\ref{app:pipeline} inventories the artifacts, and Appendix~\ref{app:walkthrough} follows the choices and consequences of one case.

\paragraph{Performance.}
Performance gives the writer material shaped by individual characters' goals and voices. It compiles the Phase~2 character profiles into scene-specific role briefs, each containing an immediate objective and a suppressed pressure the character cannot state. For scenes with at least two characters, the harness typically dispatches separate LLM calls to roleplay the participating characters and propose decisions, physical actions, lines that address a specific listener and convey subtext, reactions, and visible tells. The materials for scene $i$ form $M_{\rho(i)}$, where $\rho(i)$ denotes its character roster. Before dispatching the writer, the harness checks that each character has a material file or a recorded reason for omission. The writer may adopt, revise, or discard these candidates; the character's prohibitions remain binding and are examined during review.

\paragraph{Creation.}
Creation turns the accumulated design and performance materials into prose. Each scene is composed in a separate writer session:
\begin{equation}
s_i=W\!\left(b_i,\,\mathcal{C}_{\rho(i)},\,M_{\rho(i)},\,\kappa_i\right),
\label{eq:writer}
\end{equation}
where $\mathcal{C}_{\rho(i)}$ contains the participating characters' packages and $\kappa_i$ contains the remaining scene context. The writer chooses how to stage the planned turn, integrate the characters' proposals, and distribute attention across action, dialogue, and description. Pacing follows the scene plan, and surplus performance material is cut during composition. Because each session starts fresh, continuity with earlier scenes comes from the selected context $\kappa_i$ (Eq.~\ref{eq:context}).

\paragraph{Canon Reference.}
The reference corpus contains 31 masterworks (28 novels and 3 plays), segmented into 572 scenes. Three annotation layers serve different creative needs. The style layer profiles five register dimensions and stores a style card for each work. The craft layer explains scene development beat by beat, linking each writing decision to a quoted source span, a common model failure at the same point, and a transferable technique. The inspiration layer groups recurring narrative patterns by their dramatic function and conditions of use. Inspiration cards enter Design while plot decisions are being formed; scene exemplars enter Creation to guide narrative voice, sentence rhythm, diction, dialogue form, and omission.

Retrieval ranks content and register separately. Reciprocal rank fusion~\citep{cormack2009rrf} combines dense-embedding~\citep{karpukhin2020dense} and lexical~\citep{robertson2009bm25} rankings of scene descriptions. A register channel then reorders the candidates using a short hint about narrative distance, interiority, and tempo~\citep{wegmann2022style}. Separating the two rankings matters: generation with a semantically matched exemplar in the wrong register scores lower on every dimension than generation without a reference, and register-aligned re-ranking reverses this result. Either channel can return no usable reference, in which case generation proceeds with the story's other materials. A selected scene exemplar is supplied with its usage protocol and quote-anchored craft analysis.

\paragraph{Context engineering across stages.}
The knowledge layers determine how guidance is stored; context engineering determines when and how it is used. On entering a phase, the harness loads the corresponding Skill bundle. Deeper references are read when the current task calls for them. Design receives earlier story commitments and relevant inspiration cards; Performance receives the active character's situation and pressures; Creation receives the scene plan, character materials, prior prose state, and an optional exemplar; Review receives the text together with the design it should realize. This layered disclosure concentrates attention on information that can change the current decision~\citep{anthropic2025context}.

For scene writing, the context is assembled as
\begin{equation}
\begin{aligned}
\kappa_i={}&\big\langle\mathcal{R}_{6},\mathrm{core}_i,\mathrm{comp}(\mathrm{ref}_i),\mathrm{tail}_{i-1},\mathrm{ex}_i\big\rangle,\\
&\mathrm{ex}_i\in\mathcal{E}\cup\{\varnothing\}.
\end{aligned}
\label{eq:context}
\end{equation}
where required world and plot facts $\mathrm{core}_i$ enter in full, secondary reference notes are compressed by $\mathrm{comp}$, and $\mathrm{tail}_{i-1}$ preserves the preceding scene's immediate state. The optional $\mathrm{ex}_i$ is drawn from the canon exemplar set $\mathcal{E}$. The writer reads the design constraints and character materials before the preceding scene tail and the exemplar. This order establishes what must happen and whose choices drive it, then supplies the local continuity and prose reference needed to compose it. The character packages guide character voice, while the exemplar informs the work's narrative register.

\paragraph{Review and revision.}
Review compares the composed text with its design and directs revision. Its three tiers combine $L_1$ pattern, lexical, dialogue, and paragraph-density checks; $L_2$ aesthetic, continuity, and structural review; and $L_3$ scene-level revision decisions. The density check compares each scene with its reference, and review also checks character prohibitions. Each finding identifies the affected scene and quoted text. Findings with whole-story consequences are escalated to story-level revision. The scene-level actions are ordered by the scope of change:
\begin{equation}
\nu(s_i,b_i)\in\{\textsc{pass},\textsc{patch},\textsc{rollback},\textsc{rewrite}\}.
\label{eq:verdict}
\end{equation}
A non-\textsc{pass} verdict triggers revision against the same planned scene transition:
\begin{equation}
s_i^{(r+1)}=\mathrm{Rev}\big(s_i^{(r)},\,b_i,\,\nu(s_i^{(r)},b_i)\big).
\label{eq:revision}
\end{equation}
The loop returns the first passing revision within $r_{\max}$ attempts, or the final revision when that budget is reached. A reviser distinct from the writer applies changes anchored to quoted spans and reports the batch as complete, partial, or failed. A finding that requires rollback returns to the writer at scene scope.

After scene revision, Phase~7 assembles the story and adds a reader review of immersion, expectations, and the development of information across the whole text.

Automated checks and reviewers divide the verification work: structural and cross-file checks cover format, roster coverage, and references, while reviewers judge realized value change, subtext, and continuity (Appendix~\ref{app:pipeline}).

\subsection{From a Requested Object to a Climactic Choice}
\label{sec:worked-example}

WritingBench query 189 requests a wuxia version of a Quidditch match featuring Jin Yong characters and an object equivalent to the Golden Snitch. The case develops this object into a black dragon pearl and embeds it in a complete scoring system: jade scales earn ten points when thrown through a hoop; returning the Pearl earns thirty and ends the match. The Pearl follows a physical rule: it responds to outward discharges of inner force. Holding it requires control, while winning still depends on the score.

The conception artifact records the requested object as REQ-4 and frames mastery as what a hand holds back. This framing links the Pearl's role in the match to the story's controlling idea: the object creates situations in which pursuing victory and protecting an opponent pull in different directions.

The encounters teach both the reader and the players how this rule works. Guo Jing's palm sends the Pearl away; it briefly rests on Zhou Botong's relaxed hand; Ouyang Feng traps it with force and freezes his knuckles. Ouyang Feng then changes his technique, contains his breath, and carries the Pearl on an open palm. His adjustment makes him a tactically capable opponent. The rule creates a shared problem that each character must solve through action.

The central choice arrives when a platform breaks. Guo Jing can reach the descending Pearl, but an opposing disciple is hanging beneath him. He catches the disciple's wrist instead. The rescue injures his arm and takes up Huang Rong's time while White Camel scores three times. A fourth goal as he secures the Pearl leaves its thirty points ten short. Peach Blossom needs two more goals, and Guo Jing must keep the Pearl on a damaged rope with a left hand that will not close.

The ending carries these consequences through. As the rope parts, Huang Rong abandons her scoring run to help him reach the referees' ledge. Guo Jing returns the Pearl and ends the contest at one hundred to ninety in White Camel's favor. The rescued disciple bows to him while still clutching a strip of his torn sleeve. Huang Rong uses that cloth to bind the injury.

The case shows how story guidance becomes concrete narrative decisions. REQ-4 becomes an object with a physical response and a scoring function; character choices alter the score, bodily condition, and available routes; those changes determine the final action. Guo Jing gives up the Pearl to save an opponent's disciple and loses the match, enacting the controlling idea through a costly choice. Appendix~\ref{app:walkthrough} follows the rule, the rescue, and the finish in the story's own words.

\section{Experiments}
\label{sec:experiments}

\subsection{Setup}
\label{sec:setup}

We evaluate final outputs for query satisfaction, narrative craft, and consistency using three benchmarks. \textbf{WritingBench} (WB)~\citep{wu2025writingbench} tests whether an output satisfies the query-specific writing criteria; we use its novel-creation subset spanning seven domains (Figure~\ref{fig:wb-gpt55}). \textbf{ConStory-Bench}~\citep{li2026constory} tests whether facts remain consistent across a long story, using Consistency Error Density,
\begin{equation}
\mathrm{CED}(y)=\frac{E(y)}{|y|}\times 10^{4},
\label{eq:ced}
\end{equation}
where $E(y)$ counts consistency errors over five categories and $|y|$ is the story's word count, so CED is the number of consistency errors per ten thousand words; lower is better.

\textbf{LongStoryEval} (LSE)~\citep{yang2025longstoryeval} tests craft quality with an eight-dimension rubric applied to the same WritingBench outputs. Dataset statistics, query-id lists, and scoring categories appear in Appendix~\ref{app:dataset}.

These evaluations correspond to the requirements in Eq.~\ref{eq:vibe-quality}. WritingBench's query-specific criteria assess how the final text reflects the prompt's requirements, linking it to intent reflection $I$, while CED quantifies the cross-scene consistency required by $H$. The character ablation examines the packages used to maintain voice differentiability $V$, with domain-level effects reported for Character Design. The outline-design and review ablations examine beat fidelity $B$ through the stages that establish scene transitions and check their realization.

We compare MUSE with prompt-only zero-shot base models (\S\ref{sec:problem}) and four reproduced story-system baselines, namely Agents' Room~\citep{huot2024agentsroom}, HoLLMwood~\citep{chen2024hollmwood}, playwriting-guided generation~\citep{wu2025playwriting}, and BookWorld~\citep{ran2025bookworld}. The four base models are Claude Sonnet 4.6, Claude Opus 4.7, GPT-5.4, and GPT-5.5. All systems use the same 58-query WritingBench novel-creation subset and 38-query ConStory-Bench sample (ids in Appendix~\ref{app:dataset}). An LLM judge scores each generation~\citep{liu2023geval,zheng2023judging}; the judge is \texttt{claude-sonnet-4-6}.

\subsection{Main Results}

Table~\ref{tab:main} reports WB, ConStory-Bench CED, and LSE overall score for all four base models.

\begin{table}[t]
\centering
\small
\setlength{\tabcolsep}{4pt}
\begin{tabular}{l ccc ccc ccc ccc}
\toprule
& \multicolumn{3}{c}{Claude Sonnet 4.6} & \multicolumn{3}{c}{Claude Opus 4.7} & \multicolumn{3}{c}{GPT-5.4} & \multicolumn{3}{c}{GPT-5.5} \\
\cmidrule(lr){2-4}\cmidrule(lr){5-7}\cmidrule(lr){8-10}\cmidrule(lr){11-13}
Method & WB \up & CED \down & LSE \up & WB \up & CED \down & LSE \up & WB \up & CED \down & LSE \up & WB \up & CED \down & LSE \up \\
\midrule
\multicolumn{13}{l}{\emph{Zero-shot}} \\
Zero-shot         & 86.79 & \textbf{0.520} & 69.33 & 86.09 & \textbf{0.199} & 65.77 & 81.77 & \textbf{0.193} & 57.85 & 83.04 & \textbf{0.067} & 62.04 \\
\midrule
\multicolumn{13}{l}{\emph{Reproduced story-system baselines}} \\
Agents' Room      & 72.51 & 10.520 & 58.90 & 75.10 & 3.180 & 66.02 & 73.69 & 2.720 & 65.13 & 74.51 & 0.840 & 65.10 \\
HoLLMwood         & 54.16 & 11.040 & 52.67 & 63.88 & 3.500 & 64.02 & 65.55 & 3.440 & 59.69 & 69.29 & 4.020 & 60.50 \\
Playwriting       & 49.08 & 13.410 & 50.57 & 58.39 & 7.040 & 63.60 & 59.91 & 6.670 & 59.48 & 64.44 & 3.587 & 61.78 \\
BookWorld         & 64.68 & 12.880 & 57.38 & 72.41 & 7.620 & 66.50 & 74.55 & 5.100 & 64.72 & 75.49 & 7.150 & 66.72 \\
\midrule
\multicolumn{13}{l}{\emph{Ours}} \\
MUSE              & \textbf{88.37} & 1.247 & \textbf{69.38} & \textbf{87.15} & 0.245 & \textbf{76.31} & \textbf{87.23} & 1.650 & \textbf{68.20} & \textbf{89.24} & 2.117 & \textbf{73.10} \\
\bottomrule
\end{tabular}
\caption{Main results on WritingBench (WB), ConStory-Bench consistency error density (CED), and LongStoryEval overall score (LSE). Higher is better for WB and LSE, lower for CED; per-column best in bold. Per-benchmark rows are reproduced in Appendix~\ref{app:perbench}.}
\label{tab:main}
\end{table}

\begin{figure}[t]
  \centering
  \includegraphics[width=0.95\textwidth]{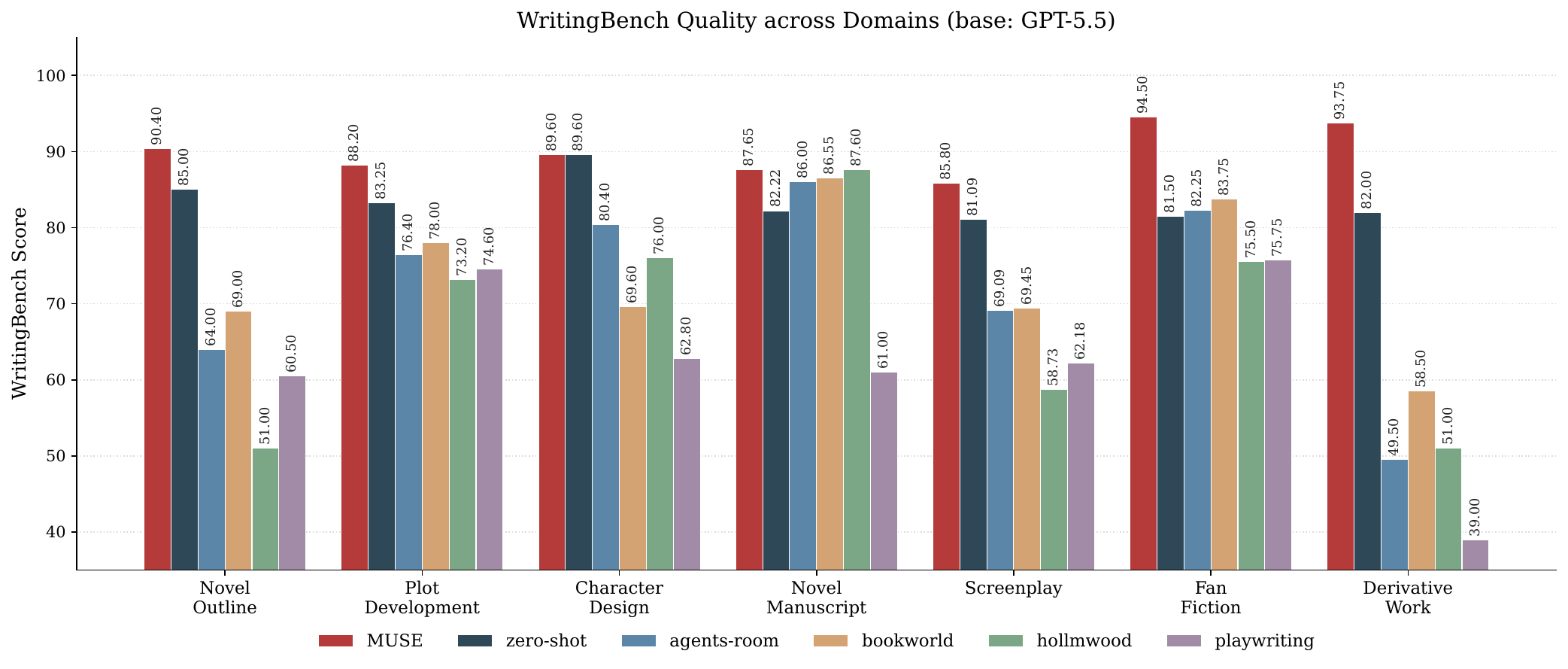}
  \caption{WritingBench scores across seven novel-creation domains under GPT-5.5. The plot compares MUSE (red), zero-shot, and four reproduced baselines. MUSE's two largest gains over zero-shot are on Fan Fiction and Derivative Work.}
  \label{fig:wb-gpt55}
\end{figure}

Every reproduced multi-stage baseline scores below its zero-shot base model on WritingBench and has higher consistency error density. MUSE is the only multi-stage system in Table~\ref{tab:main} with higher WritingBench scores than zero-shot on all four models, improving by 1.1--6.2 points while keeping CED between 0.245 and 2.117. Its stages carry theory-derived guidance together with the story decisions recorded in shared artifacts. The craft gains are model-dependent: LongStoryEval rises by more than ten points for Opus 4.7, GPT-5.4, and GPT-5.5, while Sonnet 4.6 scores 69.38 versus 69.33 for zero-shot. The following analyses separate query satisfaction, craft quality, and consistency.

\paragraph{WritingBench quality and domain variation.}
Each reproduced baseline scores 7.2--37.7 points below its corresponding zero-shot base model. The highest reproduced-baseline score in Table~\ref{tab:main}, 75.49 for BookWorld under GPT-5.5, is 11.7--13.8 points below the MUSE scores across models. We compare GPT-5.5 outputs across seven novel-creation domains in Figure~\ref{fig:wb-gpt55}. MUSE leads zero-shot on six of seven, with the largest gains on Fan Fiction ($+13.0$) and Derivative Work ($+11.8$), and matches it on Character Design (89.60). MUSE also leads every reproduced baseline on every domain. Fan Fiction and Derivative Work include continuation and adaptation tasks that require writers to develop supplied material while preserving relevant character, plot, and stylistic commitments. MUSE carries these commitments in its design artifacts and character packages, with prose references guiding the intended register. Appendix~\ref{app:perbench} reports MUSE's domain scores for each base model, and Appendix~\ref{app:case-study} analyzes cases across task families.

\paragraph{LongStoryEval craft rubric.}
LongStoryEval scores the WritingBench outputs for craft quality, complementing WB's query-dependent criteria. Craft gains are concentrated among models with lower zero-shot LSE scores. The three models starting at 57.85--65.77 gain more than ten points, while Sonnet 4.6 remains close to its zero-shot score of 69.33. MUSE's lowest LSE (68.20, GPT-5.4) also exceeds the highest reproduced-baseline score in the matrix (66.72, BookWorld under GPT-5.5).

\paragraph{ConStory-Bench consistency.}
CED scores for the reproduced baselines (Eq.~\ref{eq:ced}) range from 10.52--13.41 under Sonnet, 3.18--7.62 under Opus, 2.72--6.67 under GPT-5.4, and 0.84--7.15 under GPT-5.5. MUSE ranges from 0.245--2.117. It stays below every reproduced baseline on Sonnet, Opus, and GPT-5.4, by roughly an order of magnitude on Sonnet and Opus, and below all but one on GPT-5.5.

Across the three evaluations, MUSE combines higher query satisfaction with stronger craft scores and low single-digit consistency error density. The ablations below examine the contributions of structural design, character development, and revision.

\subsection{Ablation Study}
\label{sec:ablation}

Three ablation conditions, each run across the four base models, disable one capability path while retaining the artifact structures needed by the remaining stages.

\begin{itemize}\itemsep0pt
\item $-$ \textbf{character}. Disables character-distillation Skills and the distilled persona packages ($\mathcal{C}$ in Eq.~\ref{eq:writer}), while retaining the Phase~2 character YAML skeleton.
\item $-$ \textbf{outline-design}. Disables the Phase~0--5 design stages (Beat Tree, world building, character arc, Plot Peak design; Eqs.~\ref{eq:phase-recurrence}--\ref{eq:beat-tree}).
\item $-$ \textbf{review}. Disables the per-scene layered review inside Phase~6 ($L_1$--$L_3$, the verdict actions of Eq.~\ref{eq:verdict}, and reviser dispatch) and the Phase~7 reader audit; the final story is assembled directly from unreviewed scene drafts.
\end{itemize}

The ablations examine how each component affects writing quality and the AI-generated-content (AIGC) detection rate. For English queries, we measure the share of outputs flagged by Fast-DetectGPT~\citep{mitchell2023detectgpt,bao2024fastdetectgpt} with GPT-Neo-2.7B. Table~\ref{tab:ablation} reports the four-model averages for WB, LSE, and AIGC detection rate (lower is better). To evaluate the language problems discussed in Section~1, we use the AIGC rate as an external indicator of formulaic prose. Appendix~\ref{app:aigc} describes the governance module that addresses these prose habits during generation.

\begin{table}[t]
\centering
\small
\begin{tabular}{l R{1.0cm} R{1.0cm} R{1.0cm}}
\toprule
Condition & WB \up & LSE \up & AIGC \down \\
\midrule
Zero-shot                  & 84.42 & 63.75 & 49.9\% \\
\textbf{Full MUSE}         & \textbf{88.00} & \textbf{71.75} & \textbf{34.7\%} \\
$-$ character              & 87.33 & 70.60 & 44.6\% \\
$-$ outline-design         & 85.75 & 65.33 & 37.4\% \\
$-$ review                 & 86.60 & 68.73 & 42.5\% \\
\bottomrule
\end{tabular}
\caption{Four-model average ablation results. AIGC reports the share of outputs detected as AI-generated.}
\label{tab:ablation}
\end{table}

Full MUSE adds 3.58 WB and 8.00 LSE points over zero-shot and lowers the AIGC rate from 49.9\% to 34.7\%. The ablations isolate character packages, design, and layered review along the capability paths of \S\ref{sec:harness}. Outline-design has the largest quality effect (WB $-2.25$, LSE $-6.42$), consistent with the Beat Tree, character-arc, and Plot Peak artifacts carrying long-range structure.

Removing the character path changes aggregate WB by only $-0.67$ but raises AIGC from 34.7\% to 44.6\%. Its WB cost is concentrated in Character Design, where scores fall 6.0--8.6 points per model while other domains hold (Appendix~\ref{app:domain-ablation}). The character packages chiefly affect voice individuation.

Removing review lowers WB by 1.40 and LSE by 3.02 points and raises AIGC to 42.5\%. These changes identify the contribution of the verdict loop in Eqs.~\ref{eq:verdict}--\ref{eq:revision} to repairing drafts, removing formulaic language, and improving consistency. The three paths contribute complementary capabilities: structural development, character-specific expression, and revision of the composed story.

\section{Conclusion}

MUSE combines story theory with a practical agent harness. Knowledge engineering turns the theory into clear rules for creative decisions; the harness carries those rules and the resulting decisions through design, performance, writing, and revision. Context engineering supplies each role with the guidance and prior decisions its task requires. In the black dragon pearl case, a requested object becomes a rule the characters must learn, and the rescue it makes costly expresses the story's controlling idea. MUSE improves WritingBench over zero-shot generation on all four base models. It raises LongStoryEval by more than ten points on Opus 4.7, GPT-5.4, and GPT-5.5. Its consistency error density remains below every reproduced story-system baseline on Sonnet 4.6, Opus 4.7, and GPT-5.4. The ablations locate the largest quality contribution in structural design, voice-specific effects in the character path, and further improvements in revision. The same harness extends to open-ended serial fiction (Appendix~\ref{app:serial}).

\section*{Limitations}

\paragraph{Human evaluation.} All reported scores are automatic; evaluation by human readers is not included.

\paragraph{Token cost.} The multi-stage pipeline consumes far more tokens than zero-shot generation; end-to-end cost decomposition is not reported.

\bibliographystyle{unsrtnat}
\bibliography{custom}

\clearpage
\appendix
\numberwithin{table}{section}

\section{Reproducibility}
\label{app:repro}

Appendix~\ref{app:dataset} lists the WritingBench and ConStory-Bench query ids and the scoring categories, and Section~\ref{sec:setup} gives the judge model. Appendix~\ref{app:pipeline} describes the phases, the decisions each role reads, and the execution components of the harness; Appendices~\ref{app:canon-corpus}--\ref{app:reference-machinery} describe the canon corpus and the retrieval of inspiration cards and style anchors. The harness and evaluation procedures can be rerun on compatible platforms with separately supplied model access. Appendix~\ref{app:walkthrough} presents a detailed reading of query 189.

\section{Story Theory Knowledge Engineering and Representations}
\label{app:schema}

\subsection{Developing the Rule Library}
\label{app:rule-engineering}

The rule library connects expert principles to the decisions made during story creation. Each rule preserves the relationship among its intended effect, conditions of use, and the choices available to the role that uses it. The following examples trace this transformation from source principles to shared rules, task-specific guidance, and narrative realization.

\paragraph{Rule atomization: from subtext to character decisions.}
McKee's account of subtext distinguishes outward expression, withheld thought, and impulses the character cannot acknowledge. The prose-craft and dialogue-craft references use these distinctions to relate speech to the speaker's objective, knowledge, and relationship with the listener. When concealment or a cost of expression matters, Performance develops the character's immediate objective and what remains unspoken; Creation selects words and gestures that pursue that objective; Review examines whether the line reveals what the character would conceal. Because a direct admission can itself change the relationship, each rule retains the situation that makes its decision useful. The theoretical discussion and applications appear in \path{prose-craft/references/mckee-scene-craft.md} and \path{dialogue-craft/references/subtext-theory.md}.

\paragraph{Semantic consolidation: one decision for overlapping rules.}
MUSE merges formulations when they guide the same decision under the same conditions and responsibilities. The writer Skill supplies a concrete example. Two source instructions address overlapping inputs and repeated action steps, respectively (translated from \path{skills/writer/SKILL.md} in the MUSE-writing package):
\begin{quote}\small
\emph{Input overlap:} When multiple inputs serve the same narrative contribution, combine only restatements of the same result; retain repetition that changes action, relationships, danger, understanding, irreversible consequences, or formal effects, or that serves source reuse.

\emph{Action steps:} Combine action steps with the same \texttt{reader\_yield} only when they restate the same result; effective recurrence and repetition required by source material remain valid.
\end{quote}
The consolidated instruction reads:
\begin{quote}\small
When multiple inputs or action steps serve the same narrative contribution, combine only restatements of the same result; retain repetition that changes action, relationships, danger, understanding, irreversible consequences, or formal effects, or that serves source reuse.
\end{quote}
The merge puts both input types under one decision while preserving the conditions that justify repetition. Related rules with distinct mechanisms remain separate: concealing information and limiting a viewpoint can both sustain suspense, but their different conditions determine when each is useful.

\paragraph{Mechanism abstraction: conflict drives plot development.}
McKee connects a character's pursuit to opposing responses and consequential action~\citep{mckee1997story}. The rule library expresses this relationship at several scales: a behavioral beat changes an action--reaction pattern, a scene develops a consequential change of situation, and a sequence establishes conditions for subsequent events. These rules share a mechanism: the consequences of acting change the opportunities for further action. MUSE expresses the resulting abstraction as follows:
\begin{quote}
When a character's pursuit organizes narrative progression, develop opposition through consequential responses that change available strategies, costs, or choices. Let those consequences supply conditions for the next action.
\end{quote}
Opposition may arise from another character, the environment, an institution, or incompatible commitments. Table~\ref{tab:rule-abstraction} shows how this conflict mechanism guides distinct decisions at each narrative scale.

\begin{table}[htbp]
\centering
\small
\setlength{\tabcolsep}{5pt}
\begin{tabular}{@{}p{0.17\linewidth}p{0.39\linewidth}p{0.37\linewidth}@{}}
\toprule
Scale & Rule derived for this responsibility & Choice retained for realization \\
\midrule
Plot spine & Establish the pursuit and major turns that organize the story. & Sequences and scenes develop the intervening events. \\
Sequence & Make a local climax change the conditions for later development. & Scene design chooses the events and their presentation. \\
Scene & Preserve the intended change, character conditions, and handoff. & The writer chooses actions, dialogue, and narrative order. \\
Behavioral beat & Follow a continuing action--reaction relationship until the behavior changes. & An exchange may span several turns and any suitable paragraph organization. \\
\bottomrule
\end{tabular}
\caption{One conflict mechanism informs distinct decisions across narrative scales. The abstraction preserves the intended relationship while leaving its local realization to the corresponding creator.}
\label{tab:rule-abstraction}
\end{table}

\paragraph{WWWH: connect a rule to its point of use.}
MUSE uses Why, What, When, and How as questions for organizing a rule, drawing on goal refinement and interrogative elicitation in requirements engineering~\citep{vanlamsweerde2001goal,sultan2015w6h}. For the conflict rule, \emph{Why} states that progression becomes intelligible when pursuing an objective has consequences. \emph{What} identifies the intended change and the conditions left for the next scene. \emph{When} identifies pursuit under effective opposition as the situation in which this mechanism applies. \emph{How} relates pursuit, response, and consequence, while leaving the particular action and expression to the writer. An upstream How can become a downstream What: the scene designer specifies a consequential change, and the writer receives that change as an objective to realize.

These questions also determine where guidance appears in a Skill. Its \texttt{description} states where the Skill applies and what it can do, supporting task selection. The body retains required inputs, local activation conditions, intended results, and handoffs. A method's explanation accompanies the decision it informs; deeper theory and examples reside in references. Necessary dependencies retain their order, and alternative techniques retain their conditions of use.

\paragraph{Single source of truth: shared rules and current story decisions.}
A reusable rule has an authoritative home in the Skill or reference that explains it. Dependent guidance retains the local condition and points to that source for deeper treatment. For example, prose-craft refers dialogue development to dialogue-craft. For an individual story, Phase 2 supplies the character design and voice traits; scene-specific briefs develop current objectives and pressures; the writer receives the corresponding character information; Review uses the established boundaries to identify a voice violation. The work's global style comes from Phase 0, and the character's local voice develops within it.

\paragraph{Layered disclosure: prepare knowledge for its point of use.}
The prose-craft entry introduces scene development and the interpretation of creative materials. Its guidance relates scene development to narration, expression, and paragraph rhythm; deeper references explain beats, subtext, and contextual examples. The harness loads the phase Skill on entry and obtains detailed references when the active task requires them. A writer developing dialogue can consult the relevant exchange guidance while preserving the scene's intended outcome and the characters' knowledge. The required level of detail depends on the information available and the role's creative responsibility at that stage.

\paragraph{Examples for contextual judgment and narrative realization.}
The subtext examples connect ordinary utterances to their circumstances. After an argument, ``The food is on the table'' can convey continuing care while anger remains unspoken. Before a departure, ``Take care on the road'' can express attachment through concern for the journey. The relationship, preceding events, and anticipated response determine the utterance's function. The night-battle analysis in Appendix~\ref{app:canon-corpus} develops this connection through a sustained scene. Inspiration cards support Design's choice of dramatic relationships, while scene exemplars support Creation's prose register.

During the match in Appendix~\ref{app:walkthrough}, Guo Jing reaches for the falling disciple instead of the Pearl. The rescue injures his arm while the contest continues, changing his physical capacity and the competitive situation. He then takes a route whose knots his injured hand can use; the widening score deficit changes the significance of returning the Pearl. Later choices arise from earlier consequences. Pursuit, opposition, consequence, and changed opportunity form the reusable relationship; the Pearl, rescue, wound, and scoring conditions supply its realization in this story.

\subsection{Story Representations and Narrative Judgments}

The four McKee concepts organize fields distributed across design artifacts, role briefs, and review instructions. Table~\ref{tab:schema} summarizes the narrative judgments and stages that use them. The definitions below connect these judgments to artifact keys and their theoretical basis.

For a phase-$t$ artifact, let $\mathcal{K}_t$ denote its required keys. The predicate $\mathrm{fill}_k$ expresses whether key $k$ is present and well formed, while $\mathrm{fail}_k$ expresses a violation of its associated narrative requirement. Each design artifact records the fields its concept needs, and review checks the narrative problem each field guards against:
\begin{equation}
\Phi_t(a_t)=\bigwedge_{k\in\mathcal{K}_t}\big[\mathrm{fill}_k(a_t)\wedge\neg\,\mathrm{fail}_k(a_t)\big].
\label{eq:contract}
\end{equation}
The implementation distributes these judgments across format checks, specialized checks, and model-based review. The YAML hook handles structural properties such as parsing and required keys. Review examines semantic properties by reading the saved design and the prose.

For a scene's intended value transition, the McKee diagnostic identifies a \emph{non-event} when the entering and leaving values have the same charge:
\begin{equation}
\mathrm{fail}^{\mathrm{beat}}_i=\mathds{1}\!\big[\sigma(v_i^{-})=\sigma(v_i^{+})\big].
\label{eq:nonevent}
\end{equation}
The reviewer interprets the values in the scene's context and checks whether the action--reaction exchanges realize the planned change. Sequence and plot-spine artifacts express their progression through conflict, escalation, and climax fields, as described in Section~\ref{sec:harness}.

\begin{table}[htbp]
\centering
\small
\setlength{\tabcolsep}{3pt}
\begin{tabular}{@{}lll@{}}
\toprule
Concept & Narrative problem & Stages that read it \\
\midrule
Beat          & non-event (same charge) & Ph.~6, 7 \\
Dialogue      & on-the-nose             & Ph.~6 \\
Character Arc & static package          & Ph.~3, 4, 6 \\
Plot Peak     & removable peak          & Ph.~5, 6, 7 \\
\bottomrule
\end{tabular}
\caption{Narrative judgments for the four story theory concepts and the stages that use their representations.}
\label{tab:schema}
\end{table}

\paragraph{Beat.} The schema follows McKee's treatment of a beat as an exchange of action and reaction~\citep{mckee1997story}. It includes \texttt{value\_start}, \texttt{value\_end}, \texttt{beat\_direction}, and \texttt{value\_change}, plus a per-scene \texttt{tension\_curve} (\texttt{peaks}/\texttt{valleys}) and \texttt{scene\_causal\_chain}. A \texttt{non-event} occurs when \texttt{value\_start} and \texttt{value\_end} share a charge (Eq.~\ref{eq:nonevent}). Phase~6 consumes the fields during writing, and Phase~7 reads them during tension and causality review.

\paragraph{Dialogue.} McKee's subtext principle~\citep{mckee1997story} informs the dialogue schema. It encodes the said / unsaid / unsayable layering through \texttt{surface\_delivery}, \texttt{subtext}, \texttt{omission}, \texttt{immediate\_objective}, and \texttt{receiver\_belief\_state}. Its \texttt{on-the-nose} failure mode catches a character stating what the scene requires them to withhold. Phase~6 consumes these fields during dialogue composition.

\paragraph{Character Arc.} The character schema follows McKee's distinction between characterization and true character~\citep{mckee1997story}. Character fields comprise \texttt{desire\_system} (\texttt{conscious}, \texttt{unconscious}, \texttt{core\_flaw}), \texttt{character\_arc.mode} (transformative / revelatory / static / degenerative), \texttt{characterization\_vs\_truth} (\texttt{surface}, \texttt{deep\_truth}, \texttt{gap}), and \texttt{inner\_capacity} (\texttt{loss\_trigger}, \texttt{loss\_signal}). A \texttt{static package} marks a character who never acts under pressure. Phases~3, 4, and 6 consume these fields.

\paragraph{Plot Peak.} The plot-peak schema draws on McKee's controlling idea and obligatory climax~\citep{mckee1997story}. The \texttt{story\_climax\_design} field contains a \texttt{crisis} (\texttt{dilemma}, \texttt{option\_a}, \texttt{option\_b}), a \texttt{climax} (\texttt{action}, \texttt{value\_change}, \texttt{controlling\_idea\_expression}, \texttt{climax\_form}), and a counterfactual check through the scene-level \texttt{precedent\_mirror}. The \texttt{removable peak} failure mode identifies a high-affect scene whose removal leaves the controlling idea intact. Phases~5, 6, and 7 read these fields.

\section{Pipeline Components}
\label{app:pipeline}

\paragraph{Phases.} The pipeline runs eight phases: Phase~0 conception (premise, core value, controlling idea, genre); Phase~1 world-building (4D setting, world rules, genre conventions); Phase~2 character (desire system, character arc); Phase~3 spine (spine mode, inciting incident, arcs, climax design); Phase~4 structure (per-arc sequence expansion); Phase~5 scene arrangement (scene list, tension curve, beat direction); Phase~6 scene development (the only prose-producing phase, per-scene role briefs dispatched to the writer); Phase~7 integration (assembly plus reader-review pass).

\paragraph{Decisions and their readers.}
Saved artifacts give each story decision a persistent source. Later roles read the portions needed for their responsibilities, as Table~\ref{tab:context-handoff} shows. The writer, for example, receives the current scene card and each participant's knowledge view, a short plot-spine statement, and the preceding scene's prose tail. The full scene arrangement remains available to design and review. Phase entry loads the relevant Skill; detailed craft references are read when the task calls for them.

\begin{table}[htbp]
\centering
\small
\setlength{\tabcolsep}{4pt}
\begin{tabular}{@{}p{0.13\linewidth}p{0.29\linewidth}p{0.20\linewidth}p{0.30\linewidth}@{}}
\toprule
Stage & Saved decisions & Readers & Form in context \\
\midrule
Conception and world & Requirements, controlling idea, style, world rules and constraints & Later design; writer & Relevant fields in full; reference summaries and key details where specified \\
Character & Experiences, desires, voice, relationships and arc & Plot design; persona compiler; review & Author design for design and review; compiled persona for scene work \\
Spine and structure & Plot organization, climax, sequence events and causal links & Scene designer; writer; review & Design fields for scene planning; a one- or two-sentence spine statement for writing \\
Scene arrangement & Reading focus, necessary results, participants and handoff & Role-brief deriver; writer; review & Full scene design for derivation and review; a scene-card projection for writing \\
Performance & Each participant's knowledge, visible stimuli and constraints; candidate actions & Writer & Separate role views in full; authorized role moves when needed \\
Scene writing & Scene prose and its closing state & Next scene's writer; reviewers; assembler & Prose tail for continuity; full scenes for review and assembly \\
Canon reference & Adopted dramatic patterns and relevant prose examples & Design; writer & Selected cards and scene references on demand \\
Review & Located findings and revision decisions & Reviser; orchestrator & Current findings with the affected prose and relevant design \\
\bottomrule
\end{tabular}
\caption{Story decisions reach each role in the form required by its task. Input lists are maintained in the MUSE-writing phase Skills and \texttt{skills/writer/SKILL.md}.}
\label{tab:context-handoff}
\end{table}

\paragraph{Execution components by responsibility.}
Eight components support the five modules in Section~\ref{sec:harness}. Design uses the \emph{Beat Tree} to organize value movement across story, act, sequence, and scene. Canon Reference uses \emph{canon reference retrieval} to supply inspiration cards for design and scene exemplars for writing. Performance combines \emph{character distillation Skills}, which compile character designs into persona packages, with \emph{per-role performance materials}, which carry each character's local knowledge and candidate actions into scene composition. Creation uses \emph{layered context engineering} to assemble the scene-specific inputs in Table~\ref{tab:context-handoff}. Review uses \emph{reader-audit revision} to locate immersion breaks and send them to a separate reviser. Two components span these responsibilities: \emph{layered-disclosure Skills} load phase guidance and relevant references at their points of use, and \emph{orchestrator--subagent separation} assigns context selection and dispatch to the orchestrator while specialized roles perform the creative work.

\paragraph{Hook layer.} A plugin hook layer (\texttt{hooks.json}) couples rules to execution. \texttt{PreToolUse} hooks gate subagent dispatch (e.g., reviser patch-traceability and character-actor enforcement) and protect pipeline files. After an \texttt{Edit} or \texttt{Write} on a phase artifact, \texttt{PostToolUse} runs \texttt{check-yaml-contract}, \texttt{check-scene-card-compliance}, and \texttt{post-write-scene-lint}; \texttt{auto-phase6-index} derives the Phase~6 index after a Phase~5 write. Contract violations produce warnings that the orchestrator routes to the reviser during review. The hooks do not block the write inline.

\section{Canon Corpus Construction}
\label{app:canon-corpus}

The canon reference module of \S\ref{sec:harness} draws on masterworks reverse-engineered into annotated scenes.

\paragraph{Stance.}
The corpus uses finished masterworks as evidence of craft. The model already supplies generation capacity; the corpus supplies observations about design and prose. Analysis works backward from the ending to explain the order and rhythm of earlier events: these events must prepare the climax's value change~\citep{mckee1997story}. Verbatim scene slices retain prose texture, and line-anchored provenance keeps every excerpt traceable to its source.

\paragraph{Reverse pipeline analysis.}
Each work is reconstructed in the pipeline's own schema as conception, world, character, structure, and scene artifacts (Appendix~\ref{app:pipeline}). This makes masterwork decisions comparable with generated artifacts and usable as phase-aligned references. Scene slicing extracts eight to thirty-five verbatim scenes per work, with the count growing sublinearly with length. Every slice records genre, point of view, conflict type, content description, and the original line span.

\paragraph{Character distillation.}
Character distillation begins with scene evidence. A package is created after evidence accumulates across identity, desire, underlying truth, voice, boundaries, and arc; characters with sparse evidence remain outside the distilled roster. Every key assertion points to a verbatim scene locator (\path{scenes/scene_05.md:L10-L25}). A verification script requires all six sections, at least one locator per section, and an evidence line for every boundary rule. Irreversible endings such as death or descent into madness live in a separate file. Canon continuation loads that file, whereas alternate-universe writing omits it.

\paragraph{Three annotation layers.}
Annotation is organized by its consumers. Indexed craft guidance needs structured fields, style profiles must survive index rebuilds, and cross-work patterns need their own records. The corpus therefore has three consumer-specific layers. The \emph{style} layer records a five-field register profile per scene and a signature/anti-signature card per work. The \emph{craft} layer pairs each beat with the master's move, a common model failure at the same point, a transfer rule, and a short quote anchor. The \emph{inspiration} layer stores a dramatic function, its applicability conditions, and the source scenes that support it. During ingest, a script validates the schemas, enumerated values, and existence of every cited scene.

\paragraph{Craft annotation.}
A night battle from \emph{The Legend of the Condor Heroes} (Jin Yong; artifacts reported in English translation) is annotated as follows. One beat records how horror is built: a character verifies five holes in a skull by touch; five fingers fit exactly, as if the holes had been carved to the shape of a hand. The annotation notes that this passage contains no emotional adjective. The corresponding model default is to open with mood words and let a character cry out the killer's name. The beat's machine-readable sidecar distills the pattern: \emph{original move} --- every wound in the group fight is concrete to body part and tissue and irreversible, and the accumulating ledger of injuries is what measures the battle; \emph{model default} --- turn-based exchanges, vague injuries without consequence, a narrator announcing who is winning; \emph{transfer rule} --- measure a group fight by an accumulating ledger of concrete, irreversible injuries. A companion character package pins its voice rules and hard boundaries to the same slices: ``never harms Huang Rong'' carries the locator of the line that proves it.

\paragraph{Retrieval defaults.}
Two retrieval defaults complement the register channel in \S\ref{sec:harness}. A mandatory genre filter prevents cross-genre false positives, such as a wuxia query retrieving science-fiction dialogue on semantic similarity alone. Used as a pass/fail gate, the relevance threshold restricts reference coverage: in a ten-query audit, only one third of scenes received a reference under that rule. MUSE therefore uses the threshold to select a consumption tier: high-relevance matches supply narrative and style information, while lower-relevance matches supply style only.

\section{Reference Machinery for Inspiration Cards and Style Anchors}
\label{app:reference-machinery}

The corpus of Appendix~\ref{app:canon-corpus} serves two stages. Inspiration cards inform design when a plot lacks causal development, and style anchors guide prose register during scene writing.

\paragraph{Inspiration enters at design time.}
In flat, list-like plots, actions lack reader payoff and information has no carrier; late prose polishing cannot repair such a causal structure. During Phases 1--5, the orchestrator can promote a retrieved cross-work pattern card into a run-level ledger whose states are candidate, accepted, bound, and retired. The ledger separates dramatic \emph{patterns} from character \emph{archetypes}; a character may use at most two archetypes, with one dominant and an explicit merge boundary for the second. Once bound, a card names the design fields that consume it. Its disclosure plan specifies an early signal, a mid-story reframe, and a final confirmation, along with the scene, carrier object, permitted inference, and reserved information for each step.

\paragraph{Review closes the inspiration loop.}
Story-level review checks every bound card for a reader-visible carrier. A missing carrier is critical. A broken inference path or premature explanation produces a separate located finding. The review records what the text exposes and how the inference is staged.

\paragraph{Style anchors enter at writing time.}
Openings, climaxes, turns, high-pressure dialogue, group scenes, and scenes flagged by lexical lint receive a per-scene reference file. The file places the register profile, style card, craft patterns, and a small inspiration block before the verbatim exemplar. The writer reads it after the other scene inputs, allowing the exemplar to act as a local style anchor. The extracted dimensions are narrative voice, sentence-length rhythm, lexical texture, dialogue form, and omission. Each reference carries roughly four thousand characters of original prose and is scoped to one scene.

\paragraph{Retrieval.}
The retrieval query contains two sentences describing an ideal exemplar. Genre, language, and medium filters run first, including exclusion of stage-play scenes from prose-writing requests. Reciprocal rank fusion combines dense-embedding and lexical rankings; a register channel then scores point of view, scene type, rhythm, and dialogue density. The dense score alone determines whether a retrieved match supplies narrative and style information or style alone. Lexical or register similarity cannot promote a content-poor match into the narrative-plus-style tier.

\paragraph{Verbatim reuse requires explicit directives.}
A four-arm dispatch test identifies the reuse boundary. Permissive reuse clauses produced no shared five-grams with the source, whereas a directive block in the writer prompt produced measurable reuse. One unmanaged arm also imported 82 characters of storyteller-register narration into an epic-register story. For verbatim reuse, the writer contract specifies the permitted material by functional fit, normalizes voice, and records a reuse receipt. The full tier permits whole passages and requires high functional isomorphism; the material tier permits names, terms, and single sentences; lower-fit references enter the style tier and supply style only. Voice normalization removes source-narrator mannerisms, maps proper nouns, and gives the current design contract precedence. The writer lists reused material, and a verifier checks that receipt against the story.

\section{AI-Pattern Governance on Masterwork-Calibrated Baselines}
\label{app:aigc}

The schema layer of \S\ref{sec:schema} specifies required story properties. AI-pattern governance addresses recurring prose habits: vague reference, formulaic phrasing, fragmented rhythm, and excessive punctuation. Both draw on the same masterworks: the schema captures craft concepts, while this module measures recurring surface patterns. The AIGC rate in \S\ref{sec:ablation} is an external post-hoc metric; the governance module operates during writing and revision.

\paragraph{Locating recurring prose habits.}
Deterministic scripts detect roughly twenty Chinese pattern families, including lexical clich\'es, fragmented rhythm, dash overuse, dummy pronouns, demonstrative-classifier padding, unsupported contrastive negation, and micro-clause chains. Each hit includes a quoted span locator, and the report gives each family's density per thousand characters. The scripts locate patterns for revision without making a model call.

\paragraph{Distinguishing excess from legitimate style.}
Per-family thresholds come from percentiles of the masterwork corpus. At runtime, a family is excessive only when its density exceeds the 90th percentile of masterwork scenes. Placeholder thresholds raise cluster alerts on 96\% of those scenes; the P90 criterion yields 23\%. Legitimate high-density passages serve as hard-negative regression fixtures. Calibration also determines which passages a detector counts: in the reference works, most dashes occur in dialogue (65\% in \emph{Stoner}, 57\% in \emph{Jane Eyre}), so the detector exempts quoted spans and counts dashes in narration only.

\paragraph{Repairing repetition while preserving its function.}
Revision distinguishes \emph{protected}, \emph{legal-but-optional}, and \emph{pathological} instances. Protected instances remain when deletion would damage reference, identity, or a narrative device. Context-resolvable optional instances are usually elided; pathological instances require repair. The reviser checks whether a repair replaces one unwanted pattern with another. Scene revision handles spans named by patch directives, while distribution-level problems go to a manuscript-level reviser in a dedicated de-AI mode. Table~\ref{tab:aigc-release} summarizes how the harness records and checks these decisions.

\begin{table}[htbp]
\centering
\small
\setlength{\tabcolsep}{4pt}
\begin{tabular}{@{}p{0.22\linewidth}p{0.72\linewidth}@{}}
\toprule
Check & Implementation \\
\midrule
Family status & \emph{Observe} reports hits; \emph{enforced} adds a family to the machine ledger and release gate. Promotion requires model-output prevalence, masterwork hard negatives, a recorded human adjudication reference, and paired should-fix and must-not-fix fixtures. Manifest validation stops if an item is missing. \\
Density budget & An enforced family fails when its count exceeds $\lceil \mathrm{baseline}\times\mathrm{chars}/1000\rceil-1$. The report includes the maximum, required reduction and hit locations. Fixed-count severity remains separate from density. \\
Text identity & Admission clears stale verdicts when revision begins. The terminal check reruns detection and binds its verdict to the final text hash; a missing verdict withholds release. \\
Protected instances & If protected instances alone exceed the budget, the run escalates to an operator. A revision that does not converge ends in escalation or refusal to release. \\
Substitution & A cross-family check records whether repairs introduce the same problem in another form. \\
\bottomrule
\end{tabular}
\caption{Execution checks for AI-pattern revision.}
\label{tab:aigc-release}
\end{table}

\paragraph{Worked examples.}
Two WritingBench queries illustrate the module: 183 (first-person outbreak survival) and 367. Chinese passages are reported in English translation.

For query 183, one de-AI round brought the densities of dummy pronouns and demonstrative classifiers from above their corpus baselines to below them. Although the dispatch prompt named no pronoun target, the gate required these reductions to meet its density limits. The verdict changed from FAIL to PASS, and the revised text kept nearly the length of the source.

Two revisions of the same source illustrate the substitution problem. In one, suppressing the targeted pattern families left bare ``it'' frequent while ``they'' increased. A revision using the observation channel, substitution check, and density contract reduced bare ``it'' and removed the added ``they''.

A separate adjudication retained nearly all hits as licensed and barely changed density, whereas revision under the density contract brought the dummy-pronoun and demonstrative-classifier densities below their baselines. This comparison distinguishes whether an individual occurrence is licensed from whether the final distribution meets release limits.

For query 367, revision removed most whole-text dashes. This example makes the role of passage type concrete: punctuation density is interpreted alongside the dialogue and narration distinction used in corpus calibration.

\paragraph{Before and after.}
Two sentence pairs from query 183 (elided pronouns marked $\emptyset$):

\begin{quote}\small
\emph{Before:} When the outbreak began, \textbf{they} pounced on whoever they saw and would not stop even after dashing themselves dead; it is different now, \textbf{they} have learned to wait\,\ldots\ even I found \textbf{its} noise too loud.

\emph{After:} When the outbreak began, $\emptyset$ pounced on whoever came near and would not stop even with a skull staved in; now $\emptyset$ have learned to wait\,\ldots\ even I found the noise too loud.
\end{quote}

\begin{quote}\small
\emph{Before:} I stared at \textbf{them} for a beat, my mind reaching for \textbf{that set of things} --- compute the odds, find the exit\,\ldots\ \textbf{that} door to the corridor stood ajar.

\emph{After:} I stared for a beat, my mind reaching for \textbf{the skills I live on} --- compute the odds, find the exit\,\ldots\ the door to the corridor stood ajar.
\end{quote}

Chinese permits pronoun drop, allowing context to carry the reference after elision. Where the source used the vague object ``that set of things,'' revision supplied its concrete referent. The same pass kept every occurrence of ``that thing'' because the narrator's refusal to name the infected is a narrative device. The revision record stores this protected classification and the substitution check beside the counts.

\paragraph{Language and register.}
Enforced families cover Chinese; English families run in observation mode. Thresholds use corpus-wide percentiles. Pattern frequencies vary by register: bare ``it'' occurs near zero per thousand characters in the web-novel register and at 2.1 in translated science fiction.

\section{Dataset Details}
\label{app:dataset}

\paragraph{WritingBench.}
All systems use the same novel-creation query set, and LongStoryEval scores the outputs generated for it. The set contains 32 English and 26 Chinese queries. Query ids are listed by domain, with query counts in parentheses:
\begin{itemize}\itemsep0pt
\item Screenplay (12): 195, 196, 376, 377, 535, 598, 636, 696, 746, 861, 931, 990.
\item Plot Development (11): 186, 187, 188, 369, 370, 533, 634, 798, 858, 928, 987.
\item Novel Manuscript (11): 182, 183, 184, 366, 367, 531, 693, 742, 797, 926, 986.
\item Fan Fiction (8): 185, 368, 532, 594, 694, 743, 857, 927.
\item Novel Outline (7): 180, 181, 365, 530, 593, 796, 856.
\item Character Design (5): 204, 385, 599, 639, 933.
\item Derivative Work (4): 189, 190, 595, 799.
\end{itemize}

\paragraph{ConStory-Bench.}
The sample uses query ids 2, 10, 134, 188, 233, 694, 750, 792, 840, 860, 889, 993, 1076, 1084, 1092, 1101, 1109, 1144, 1207, 1272, 1311, 1361, 1380, 1451, 1505, 1519, 1584, 1596, 1662, 1726, 1801, 1821, 1868, 1952, 1960, 1985, 1993, 1998.

\paragraph{Scoring categories.}
ConStory-Bench scores consistency errors in characterization, factual\_detail, narrative\_style, timeline\_plot, and world\_building. LongStoryEval scores plot\_structure, characters, writing\_language, world\_building, themes, emotional\_impact, enjoyment, and expectation.

\FloatBarrier
\begingroup
\setlength{\intextsep}{6pt}
\captionsetup{font=footnotesize,position=bottom,skip=8pt}
\Needspace{30\baselineskip}
\section{Per-Benchmark Result Tables}
\label{app:perbench}

Tables~\ref{tab:wb}--\ref{tab:wb-domain} reproduce the per-benchmark rows of Table~\ref{tab:main} and add the MUSE per-domain breakdown across the four base models.

\begin{table}[!htbp]
\centering
\begin{minipage}[t]{0.49\linewidth}
\centering
\small
\setlength{\tabcolsep}{3pt}
\begin{tabular}{l r r r r}
\toprule
Method & Sonnet 4.6 & Opus 4.7 & GPT-5.4 & GPT-5.5 \\
\midrule
Zero-shot     & 86.79 & 86.09 & 81.77 & 83.04 \\
Agents' Room  & 72.51 & 75.10 & 73.69 & 74.51 \\
HoLLMwood     & 54.16 & 63.88 & 65.55 & 69.29 \\
Playwriting   & 49.08 & 58.39 & 59.91 & 64.44 \\
BookWorld     & 64.68 & 72.41 & 74.55 & 75.49 \\
MUSE          & \textbf{88.37} & \textbf{87.15} & \textbf{87.23} & \textbf{89.24} \\
\bottomrule
\end{tabular}
\caption{WritingBench overall scores (higher is better). Per-column best in bold.}
\label{tab:wb}
\end{minipage}\hfill
\begin{minipage}[t]{0.49\linewidth}
\centering
\small
\setlength{\tabcolsep}{3pt}
\begin{tabular}{l r r r r}
\toprule
Method & Sonnet 4.6 & Opus 4.7 & GPT-5.4 & GPT-5.5 \\
\midrule
Zero-shot     & 69.33 & 65.77 & 57.85 & 62.04 \\
Agents' Room  & 58.90 & 66.02 & 65.13 & 65.10 \\
HoLLMwood     & 52.67 & 64.02 & 59.69 & 60.50 \\
Playwriting   & 50.57 & 63.60 & 59.48 & 61.78 \\
BookWorld     & 57.38 & 66.50 & 64.72 & 66.72 \\
MUSE          & \textbf{69.38} & \textbf{76.31} & \textbf{68.20} & \textbf{73.10} \\
\bottomrule
\end{tabular}
\caption{LongStoryEval overall scores (higher is better). Per-column best in bold.}
\label{tab:lse}
\end{minipage}

\vspace{1.4em}
\begin{minipage}[t]{0.49\linewidth}
\centering
\small
\setlength{\tabcolsep}{3pt}
\begin{tabular}{l r r r r}
\toprule
Method & Sonnet 4.6 & Opus 4.7 & GPT-5.4 & GPT-5.5 \\
\midrule
Zero-shot     & \textbf{0.520} & \textbf{0.199} & \textbf{0.193} & \textbf{0.067} \\
Agents' Room  & 10.520 & 3.180 & 2.720 & 0.840 \\
HoLLMwood     & 11.040 & 3.500 & 3.440 & 4.020 \\
Playwriting   & 13.410 & 7.040 & 6.670 & 3.587 \\
BookWorld     & 12.880 & 7.620 & 5.100 & 7.150 \\
MUSE          & 1.247 & 0.245 & 1.650 & 2.117 \\
\bottomrule
\end{tabular}
\caption{ConStory-Bench consistency error density per 10{,}000 words (lower is better). Per-column best in bold. MUSE keeps CED below 2.2 on all base models and below every reproduced baseline on Sonnet, Opus, and GPT-5.4, by roughly an order of magnitude on Sonnet and Opus.}
\label{tab:ced}
\end{minipage}\hfill
\begin{minipage}[t]{0.49\linewidth}
\centering
\small
\setlength{\tabcolsep}{3pt}
\begin{tabular}{l r r r r}
\toprule
Domain & Sonnet & Opus & GPT-5.4 & GPT-5.5 \\
\midrule
Character Design  & 92.80 & 85.80 & 91.40 & 89.60 \\
Derivative Work   & 93.50 & 93.25 & 91.00 & 93.75 \\
Fan Fiction       & 92.25 & 91.80 & 92.50 & 94.50 \\
Novel Outline     & 86.20 & 88.30 & 84.70 & 90.40 \\
Novel Manuscript  & 87.10 & 86.50 & 86.50 & 87.65 \\
Plot Development  & 85.60 & 84.00 & 84.05 & 88.20 \\
Screenplay        & 87.20 & 85.40 & 85.80 & 85.80 \\
\bottomrule
\end{tabular}
\caption{MUSE WritingBench scores by domain.}
\label{tab:wb-domain}
\end{minipage}
\end{table}

\FloatBarrier
\Needspace{12\baselineskip}
\section{Ablation: WritingBench by Domain}
\label{app:domain-ablation}

Tables~\ref{tab:abl-sonnet}--\ref{tab:abl-gpt55} report domain-level WritingBench scores under the three ablation conditions across the four base models. These conditions match Table~\ref{tab:ablation} of the main paper. Two patterns recur across all four models. Removing outline-design collapses Novel Outline (73.20--77.60 against 84.70--90.40 under Full MUSE) and depresses Plot Development by 2.4--2.9 points. Both domains require structural plans, the artifacts removed by this ablation.

Removing the character module collapses Character Design (79.80--84.20 against 85.80--92.80) while changing scores in the remaining domains by only a few tenths, which localizes the character package's contribution to persona-centric work.

\begin{center}
\captionsetup{type=table}
\begin{minipage}[t]{0.49\linewidth}
\centering
\small
\setlength{\tabcolsep}{3pt}
\begin{tabular}{l r r r r}
\toprule
Domain & Full & $-$char & $-$design & $-$rev \\
\midrule
Character Design & 92.80 & 84.20 & 92.40 & 91.60 \\
Derivative Work  & 93.50 & 93.40 & 93.30 & 91.80 \\
Fan Fiction      & 92.25 & 92.20 & 91.65 & 90.70 \\
Novel Outline    & 86.20 & 86.10 & 75.50 & 85.00 \\
Novel Manuscript & 87.10 & 87.00 & 86.50 & 85.70 \\
Plot Development & 85.60 & 85.50 & 82.80 & 84.20 \\
Screenplay       & 87.20 & 86.90 & 86.20 & 85.95 \\
\bottomrule
\end{tabular}
\caption{Claude Sonnet 4.6 (Full = 88.37).}
\label{tab:abl-sonnet}
\end{minipage}\hfill
\begin{minipage}[t]{0.49\linewidth}
\centering
\small
\setlength{\tabcolsep}{3pt}
\begin{tabular}{l r r r r}
\toprule
Domain & Full & $-$char & $-$design & $-$rev \\
\midrule
Character Design & 85.80 & 79.80 & 85.70 & 84.30 \\
Derivative Work  & 93.25 & 93.15 & 93.15 & 91.40 \\
Fan Fiction      & 91.80 & 91.75 & 91.60 & 90.10 \\
Novel Outline    & 88.30 & 88.20 & 75.60 & 87.00 \\
Novel Manuscript & 86.50 & 86.50 & 86.10 & 85.30 \\
Plot Development & 84.00 & 83.95 & 81.30 & 82.60 \\
Screenplay       & 85.40 & 85.40 & 84.50 & 83.90 \\
\bottomrule
\end{tabular}
\caption{Claude Opus 4.7 (Full = 87.15).}
\label{tab:abl-opus}
\end{minipage}

\end{center}
\vspace{0.8em}
\begin{center}
\captionsetup{type=table}
\begin{minipage}[t]{0.49\linewidth}
\centering
\small
\setlength{\tabcolsep}{3pt}
\begin{tabular}{l r r r r}
\toprule
Domain & Full & $-$char & $-$design & $-$rev \\
\midrule
Character Design & 91.40 & 83.50 & 91.30 & 90.10 \\
Derivative Work  & 91.00 & 90.90 & 90.80 & 89.40 \\
Fan Fiction      & 92.50 & 92.35 & 92.30 & 91.00 \\
Novel Outline    & 84.70 & 84.60 & 73.20 & 83.40 \\
Novel Manuscript & 86.50 & 86.50 & 86.20 & 85.20 \\
Plot Development & 84.05 & 84.05 & 81.15 & 82.80 \\
Screenplay       & 85.80 & 85.75 & 84.90 & 84.55 \\
\bottomrule
\end{tabular}
\caption{GPT-5.4 (Full = 87.23).}
\label{tab:abl-gpt54}
\end{minipage}\hfill
\begin{minipage}[t]{0.49\linewidth}
\centering
\small
\setlength{\tabcolsep}{3pt}
\begin{tabular}{l r r r r}
\toprule
Domain & Full & $-$char & $-$design & $-$rev \\
\midrule
Character Design & 89.60 & 83.60 & 89.50 & 88.20 \\
Derivative Work  & 93.75 & 93.65 & 93.50 & 92.00 \\
Fan Fiction      & 94.50 & 94.40 & 94.40 & 92.80 \\
Novel Outline    & 90.40 & 90.40 & 77.60 & 89.10 \\
Novel Manuscript & 87.65 & 87.65 & 87.50 & 86.40 \\
Plot Development & 88.20 & 88.20 & 85.75 & 86.80 \\
Screenplay       & 85.80 & 85.80 & 85.50 & 84.35 \\
\bottomrule
\end{tabular}
\caption{GPT-5.5 (Full = 89.24).}
\label{tab:abl-gpt55}
\end{minipage}
\end{center}

\FloatBarrier
\endgroup
\section{Case Walkthrough: Rules, Choices, and Consequences}
\label{app:walkthrough}

WritingBench query 189 asks for a wuxia-style Quidditch match featuring Jin Yong characters, a Golden-Snitch equivalent, dangerous situations, and spectacular exchanges between martial-arts masters. The story places the contest on nine bamboo platforms suspended from a cliff of Peach Blossom Island above the sea. Huang Yaoshi, Huang Rong, and Guo Jing play for Peach Blossom; Ouyang Feng, his disciple Yang Kang, and Zhou Botong play for White Camel; Hong Qigong and Yideng referee. The passages below trace its scoring rules, the players' discoveries, and the consequences of a rescue.

\subsection{A Rule That Governs the Match}

Throwing a jade scale through a bronze hoop scores ten points. Returning the Pearl, a black dragon pearl, to its cage scores thirty points and ends the match. Its possessor must therefore consider both safe return and the current score.

\begin{quote}
\small
``When that incense is ash, the Reverend opens the cage,'' Hong Qigong reminded the players, chewing. ``Thirty for putting the Pearl back in it, and then the match is over. Until then, keep playing.''
\end{quote}

Huang Yaoshi's articles add one more rule: a player who draws an opponent's blood with his own hand loses the match for his side. The hoops, tally bowls, connecting ropes, and suspended platforms remain active throughout the scene. Ouyang Feng uses darts to disturb landing places, while Yang Kang, whom Guo Jing greeted that morning as his little brother, scores through the gaps. Huang Rong gives up two easy throws to steady Guo Jing's landings. The match establishes the price of helping a teammate before that price becomes decisive.

\subsection{Learning the Pearl's Response}

The Pearl flees outward force. It rises from Guo Jing's palm, briefly rides on Zhou Botong's loose hand, and escapes when he clenches his fists. Ouyang Feng pins it against the cliff with the Toad Skill and closes his hand on it; frost whitens two knuckles. He then adjusts:

\begin{quote}
\small
Huang Rong had expected another rush, but he waited, his breathing quiet, the staff laid along his forearm. Then he cupped his uninjured hand beneath the Pearl, and it settled there.

``Jing-gege,'' she called. ``He has it!''

Her father's flute shot toward Ouyang Feng's wrist. The Western Venom moved his open palm aside and parried with the staff, and the Pearl stayed on his skin. A master who kept his breath contained could carry it as readily as a child carries an egg; Ouyang Feng had learned that from one frozen touch, and now he had only to cross the field. But he could not strike. A blow with everything behind it rises through the whole body, through the carrying arm as well, and the Pearl would have felt it go.
\end{quote}

Ouyang Feng's adjustment shows how the Pearl's response constrains its carrier. Huang Rong taunts him until he answers aloud, and the breath of speech lifts the Pearl a hair's breadth. As Huang Yaoshi presses at that opening, Ouyang Feng tips the Pearl down a sloping rope toward Yang Kang, bringing the three younger players onto the same rope.

\subsection{The Rescue Changes the Contest}
\label{app:walkthrough-brief}

A platform already split by a dart breaks under the masters' palm exchange and strikes the sloping rope. Yang Kang falls and catches a splintered crosspiece. Guo Jing is close enough to reach either him or the Pearl:

\begin{quote}
\small
The Pearl was coming down within reach of Guo Jing's right hand. He turned under the rope and caught his brother's wrist instead.

Their names had been chosen on the same snowy night, before either of them was born, by a Taoist who wanted both boys to remember the Jingkang years, when the Jurchen took Kaifeng and carried two Song emperors off to the north: Jing for the first half of that era's name and Kang for the second. Guo Jing had never once thought of letting go of the second half.
\end{quote}

The rescue requires cooperation and costs Peach Blossom time. Yang Kang claws at the passing Pearl, which flees and frosts his fingertips; the jerk tears Guo Jing's sleeve and grinds the rope into his forearm. Guo Jing holds him while Huang Rong levers the broken frame away, Yang Kang cuts his trapped sash, and the two referees work a rescue line. Hong Qigong refuses Huang Rong's call to stop play, because the articles keep the match running until the Pearl is caged, and Zhou Botong keeps the Pearl moving with one hand while scoring with the other. White Camel's lead grows to ninety against sixty. Guo Jing reaches the platform with a left hand that will not close, and Huang Rong calls for one more goal before he returns the Pearl.

\subsection{Possession and the Final Decision}

Guo Jing reaches the Pearl along an inner route with knots that support his injured hand, and keeps his palm open as Zhou Botong presses him off balance. Above them, Ouyang Feng intercepts Huang Rong's throw and scores again, bringing White Camel to one hundred, so the Pearl can now reduce the deficit only to ten points. While Guo Jing holds it the match continues: Huang Yaoshi and Huang Rong go after the two goals Peach Blossom needs, and Ouyang Feng comes down the lowest rope to make Guo Jing lose the Pearl. Guo Jing meets his palm with the Overweening Dragon Has Regret, sending out ten parts of force and keeping twenty in reserve, none of it in the arm that carries the Pearl. The blood rule keeps Ouyang Feng from striking, so he drops his weight onto the rope instead.

The rope begins to part at its iron anchor, where the falling platform had struck it. His right hand carries the Pearl, his left is injured, and the nearest firm landing is the referees' ledge. Ke Zhen'e hears the fraying and shouts; Huang Rong leaves her run to bring him up, and her father pulls him onto the ledge. Guo Jing cages the Pearl. Huang Rong charges that Ouyang Feng broke the rope; Yideng finds that it parted where the platform struck it. The score follows from the accumulated choices:

\begin{quote}
\small
Hong Qigong looked from one of them to the other, and from the ruined platforms to the cage, chewing on nothing. ``Then nobody cut it,'' he said. ``It wore through.'' He rang the bell again. ``One hundred to ninety. White Camel wins.''
\end{quote}

\subsection{The Cost Remains After the Bell}
\label{app:walkthrough-review}

The ending returns to the rescue through the torn sleeve. Yang Kang bows to Guo Jing with a strip of the sleeve still in his fist; Huang Rong takes it to bind the injured arm. Ouyang Feng's frostbitten fingers cannot close on the silver bowl given to the winning captain, so Yang Kang carries it. Yideng treats Guo Jing's arm with the One Yang Finger, and Huang Yaoshi, as losing captain, draws the white powder down his own cheeks.

A shared rule organizes the pursuit, the rescue changes both Guo Jing's physical abilities and the score, and the damaged rope forces the final return. The match ends through the consequences of decisions made under pressure. The result and the characters' relationships on the ledge express the controlling idea that mastery shows in what a hand holds back: the palm that keeps its force in reserve can carry the Pearl and a brother's weight, while the grasping hand wins only the tally.

\section{Case Studies of Creative Requirements}
\label{app:case-study}

We examine ten queries from the WritingBench novel-creation subset, including query 189 in Appendix~\ref{app:walkthrough}. They cover distinct task forms: long-form outline, short-story plotting, derivative writing, character design, opera adaptation, historically constrained stage play, systematic worldbuilding, psychological progression, concept-driven screenplay, and absurdist theatre. Local scores are per-query WritingBench scores (1--10) from the judge used in Section~\ref{sec:setup}. Chinese queries and artifacts are presented in English translation, and each quotation gives the path of its artifact file.

Each case traces the query's \emph{intent} through its design and \emph{realization} in prose to the resulting \emph{outcome}. Table~\ref{tab:cases} groups them by analytical focus: \emph{full chains} trace a requirement through design decisions to its realization and consequences in the finished work, while \emph{capability evidence} focuses on how a completed response meets a particular set of writing requirements.

\begin{table}[htbp]
\centering
\small
\setlength{\tabcolsep}{4pt}
\begin{tabular}{r l r}
\toprule
Query & Domain & Local score \\
\midrule
\multicolumn{3}{l}{\emph{Full chains}} \\
180 & Novel Outline      & 10.0 \\
187 & Plot Development   & 9.0  \\
189 & Derivative Work (App.~\ref{app:walkthrough}) & 9.6 \\
204 & Character Design   & 9.6  \\
377 & Screenplay (opera) & 9.2  \\
535 & Screenplay (stage) & 9.2  \\
\midrule
\multicolumn{3}{l}{\emph{Capability evidence}} \\
365 & Novel Outline        & 10.0 \\
856 & Novel Outline        & 9.4  \\
990 & Screenplay (wuxia)   & 9.4  \\
196 & Screenplay (absurd.) & 7.8  \\
\bottomrule
\end{tabular}
\caption{Selected case queries.}
\label{tab:cases}
\end{table}

\paragraph{Shared mechanism.}
Each case starts with Phase~0, which records aesthetic intent, task boundaries, and hard requirements in fields for premise, core value, controlling idea, genre, originality, and the requirement register. Later phases develop these decisions into world rules, character systems, a plot spine, scene structure, and prose. The \texttt{core\_value} guides scene design, character choice, and review; \texttt{phase2\_character.yaml} supplies the writer with voice and action constraints; \texttt{phase5\_scenes.yaml} specifies the scene contract realized in Phase~6 (Appendix~\ref{app:schema}). The requested deliverable determines how far the pipeline runs: a character commission can stop after Phase~2, while a scene-writing task continues through drafting and review.

\paragraph{Query 180 (Novel Outline).}
\emph{Intent.} ``I want to create a post-apocalyptic sci-fi novel and need help with an outline. The protagonist is a resilient delivery courier who is introverted but has a strong sense of responsibility, continuing to deliver supplies to survivors even after the apocalyptic crisis breaks out,'' with required survival, emotional, and mystery elements: resource-scarce urban survival, care for elderly survivors and orphans, and forces secretly manipulating the disaster.
\emph{Design.} The outline links relief work to the investigation through ration accounts and delivery obligations. A rejected parcel leads Wen Yi to a household recorded as evacuated; maintaining the children's food supply consumes goods and displaces another courier's route. The final handover requires witnesses, records, transport, and people willing to receive the residents. These dependencies connect survival, care, and discovery across the fourteen chapters.
\emph{Realization} (\path{180/story.md}, source-language English):
\begin{quote}
\small
Wen has never made a speech, so he reads addresses instead: each note he wrote that the register copied, and the name of the person it describes, who steps forward as he reads. Liang Cheng identifies the receipts he signed and explains why he signed them. Han offers him treatment inside the Filtered District. Mrs.\ Liang, who has come from the school at first light, asks whether she may stay with him and whether the others will receive the same care, which turns Han's offer to one man into a question she has to answer in front of everyone on the quay.
\end{quote}
\emph{Outcome.} Local score 10.0. The climax brings the residents and the records before the same recipients. Wen is detained while his colleagues complete the handover. The ending continues through treatment, reunification, food rounds, and shared dispatch work.

\paragraph{Query 187 (Plot Development).}
\emph{Intent.} ``Please help me design a story of around 1{,}500 words with the theme of finding life's direction through a hometown Sichuan dish.''
\emph{Design.} The story is told in three parts named for the stages of twice-cooked pork: the first cooking, cooling, and the return to the wok. The cooking scene brings together the protagonist's breakfast shifts, a Shanghai contract due for signature by morning, her cousin's proposed shop in Chengdu, and memories of her father. Preparing and eating the dish changes what she notices about work she already does. She declines the contract and the shop and asks about a month of Friday suppers in a familiar kitchen.
\emph{Realization} (\path{187/story.md}, source-language English):
\begin{quote}
\small
I email Shanghai to decline and write separately to their chef, who taught me how a sauce changes when it is frozen, to thank him.

For Hao I record a voice message. I tell him I'm not coming, that he should let the landlord go, that I'm sorry I made him ask so many times. I say I'll come at Spring Festival and cook for his mother, with \emph{lianbai} if the garlic shoots are poor.

Mrs.\ Lau will be up at four, so I leave her a message asking whether her offer still stands: a month of Friday suppers, one dish, rice from the pot.

Near two Hao answers: \emph{All right. Call tomorrow anyway.} Then a second line: \emph{Leeks are fine.}

I divide the leftover pork between two containers, one for my lunch, one for Mrs. Lau to taste before we talk about the stove. The slices have set in their curl, each holding a little red oil, the way a lamp holds oil before anyone lights it. On the back of the contract's last page I write the first menu: \emph{Friday. Huiguorou, with whatever greens are good. Rice, as much as you want.}
\end{quote}
\emph{Outcome.} Local score 9.0. The decision follows sustained work experience and the meal. Distinct replies settle existing commitments, and a proposed month of suppers gives the new direction a schedule, a workplace, and a first menu.

\paragraph{Query 204 (Character Design).}
\emph{Intent.} ``I'm writing a campus light novel and need help designing a male protagonist who appears cheerful and outgoing but has some inner insecurities.''
\emph{Design.} Haruto's sociability grows from interests and relationships: he plays bass, enjoys games, and draws classmates into activities. His insecurity intensifies when plans change without him or friends he introduced begin making plans of their own. Agreeing to play bass for three bands and to help with the class café then creates conflicts between rehearsal, café shifts, and a magazine deadline, giving the character an ordinary situation in which to make a difficult choice.
\emph{Realization} (\path{204/story.en.md}, paired English version of the Chinese deliverable):
\begin{quote}
\small
In the week before the festival, his late arrival at rehearsal costs the band a run-through, and his unfinished table cards consume time Misaki had set aside for her friend. He starts with a joking apology. Misaki asks, ``Are you coming tomorrow or not? I need to put someone down.''

Now he has to choose, and someone will be disappointed whatever he picks. He stays with Before the Last Train, apologizes in person to the other two bands, and recommends Ono, the first-year he used to practice with, as their bassist. He gives up a café shift, hands over the work he has completed, and personally asks another classmate to cover for him. Misaki is still annoyed, and Ono sounds perfectly fine in someone else's band. What he has to resist that evening is the urge to take everything back on so that everyone will smile at him again right away.
\end{quote}
\emph{Outcome.} Local score 9.6. The design links a specific fear to a recurring mistake and a consequential choice. Haruto stays with his own band, passes the other two to a first-year bassist, gives up a café shift, hands over the work he has completed, and tolerates a friend's annoyance.

\paragraph{Query 377 (Peking Opera adaptation).}
\emph{Intent.} After a historical synopsis of the Battle of Red Cliffs, the prompt asks: ``Please adapt the [Battle of Red Cliffs] into a 2-hour Peking Opera script for use as educational material to promote traditional culture among young people,'' requiring the four arts of singing, speaking, acting, and martial arts; language suitable for students aged 13--18; 20--30 minute scenes; all four role types (\emph{sheng}, \emph{dan}, \emph{jing}, \emph{chou}); at least four arias; no more than two fight scenes; and detailed lyrics, dialogue, movement, and scene arrangements.
\emph{Design.} Five scenes of 23--25 minutes organize eight numbered arias and two fighting scenes; four of the arias are set pieces, one for the \emph{dan}, one for the \emph{laosheng}, and one for each of two \emph{jing} roles. Huang Gai proposes a fire attack carried in under a false surrender, and Zhou Yu must order a flogging convincing enough for Jiang Gan, Cao Cao's envoy, to witness. The command seal connects the private night agreement, sung in Zhou Yu's aria \emph{The Command Seal}, with his restraint at the morning muster. The agreement ties fast escape boats behind the fire-ships; Huang Gai still goes into the winter river, and at the southern landing Cheng Pu recognizes him by his voice among the unknown wounded.
\emph{Realization} (\path{377/script.md}, source-language English):
\begin{quote}
\small
At the fifth gong, the fiftieth stroke, he falls forward onto his palms, and his long beard swings out and down. Cheng Pu takes off his helmet and kneels before the desk; Lu Su throws himself across Huang Gai's back, and the rod stays raised above them.

Zhou Yu half rises, his hand reaching toward the edge of the desk. He stops, closes his hand on the seal, and sits again.
\end{quote}
\emph{Outcome.} Local score 9.2, with performance element integration at 10. The flogging is acted to percussion, with the rod striking empty air, so the visible punishment carries the concealed agreement. At the southern landing Zhou Yu tells the soldiers that Huang Gai's surrender was their plan and that he took the rod with the commander's agreement, returning public credit to the man whose disgrace made the attack possible.

\paragraph{Query 535 (Stage play).}
\emph{Intent.} ``Write a script for a stage play set in early 20th-century Shanghai, showcasing the urban landscape under the fusion of Chinese and Western cultures,'' with underground revolutionary activity as background, a disillusioned poet whose poetry becomes a tool for transmitting revolutionary messages, and a prologue, conflict, twist and climax, conclusion, intricate psychology, tense narrative, literary dialogue, historical-cultural texture, roughly 90 minutes, and a scene breakdown.
\emph{Design.} The play lets a printed alteration change what different readers do. Lin finds a comma he did not write in his magnolia poem, and Zhu, a schoolteacher and underground courier, buys the issue and pauses over the comma he has crossed out. A later offer of a teaching post in Kowloon comes with a statement declaring that political matter entered \emph{Spring Mud} without his knowledge, and, with a policeman present, Lin says a printer's name, which later brings the police back to that printer. The final poem, ``Errata,'' revisits the printer Qiao holding a door while others escape and the obligations of those who survived, while its publication requires a decision about space and payment.
\emph{Realization} (\path{535/script.md}, source-language English):
\begin{quote}
\small
LIN
It takes the space of two stanzas on page seventeen. I counted it against your layout. The advertisement goes to the back.

SU
The advertiser paid for seventeen.

LIN
Pay him back out of my fee.
\end{quote}
\emph{Outcome.} Local score 9.2, with character development at 10. The completed poem joins witnessed events to their human aftermath. Su agrees to print material that revises the ``we'' of his own editorial, Lin signs his full name and leaves Shanghai, and in the epilogue Larissa hands the last issue to a schoolgirl whose teacher, Miss Zhu, told the class to read it.

\paragraph{Query 365 (Novel Outline).}
\emph{Intent.} ``I want to write an exciting zombie-fighting novel. Here's my current zombie setting: Zombie Level Classification,'' followed by a Basic-Zombie-through-Zombie-King hierarchy keyed to eye color, mutation rarity, supernatural ability, and cultivation stage, with a request for a coherent chapter outline, one chapter per zombie level, cliffhangers after each defeated zombie, and avoidance of cliches.
\emph{Design.} The outline assigns one supplied tier to each chapter, from basic zombies through the Zombie King and the Earth Demon Immortal, and turns each tier's ability into a tactical problem. A retained work habit exposes the Level~1 driver; force carried through handrails, pipes, and stair stringers makes the Level~2 corpse a problem of routes; the Level~7 Cartographer runs three bodies, negotiates, and retaliates by releasing its prisoners into two killing grounds. Each victory opens part of the route and changes the enemy's next move. The fights draw on shared resources and people: a lost truck, a shelter of three hundred people, Shen Jingwu's death, and a city that must rebuild along a surface road.

\emph{Realization} (\path{365/story.md}, source-language English):
\begin{quote}\small
Han runs ahead as bait, burning his stolen stage to stay out of reach. All night the Immortal believes it is pacing the edge of the city. At what it takes for the last corner it finds five pillars holding up a haze of qi, scores its claim on them in the grave-sea script and leaves its mark at the foot of the first. Then it returns to its starting point to close the boundary.

Han, nearly burned out, shows it where it has been: the five pillars are the anchor columns of the Chorus Hall basement, three hundred metres from where it began. It has claimed one city block, and a claim binds both ways: it can draw only on what lies inside. Before it can walk a new boundary, the crews drop the five columns and the hall comes down on it in a mound. Yue sets the last ward plate on the top stone, and the rubble takes root and closes its seams. Han cuts the line beneath his feet and falls into the net Wei's crew has rigged below.

Unable to destroy the Immortal, the city keeps it under the rubble and posts a keeper at the mound. Now and then a hand pushes up between the stones and shakes.
\end{quote}
\emph{Outcome.} Local score 10.0. The supplied levels organize twelve chapters, each ending on a hook produced by the preceding fight. The final opponent cannot be harmed; the coalition holds it by leading it to claim a single city block and then bringing down the Chorus Hall, whose basement columns anchor that block, on top of it. The progression satisfies the chapter requirements, and the climax depends on the ward repairer Lin Yue's survey of the underground currents and on the city's work crews as much as on the swordsman Lin Han.

\paragraph{Query 856 (Novel Outline).}
\emph{Intent} (translated). A chapter-by-chapter outline about a lonely painter whose work has never been accepted, tracing psychological change from loneliness, confusion, and self-doubt to redemption through a soulmate, renewed confidence, and eventual recognition --- with chapter numbers and titles, plot summaries, relationship development, psychological detail, key events, pacing, setting, painter profile, soulmate arc, signature paintings, and at least 1{,}000 Chinese characters.
\emph{Design.} The outline follows Lin Zhiyao from 1992 to 2017 across fourteen chapters in three parts. His early paintings give every sitter the same heaviness. Rejections, a critic who offers a review in exchange for a fee, and his father's question about why he was painted at his worst first confirm Lin's belief that no one understands him. A rejection note saying that his figures have no relation to one another then sends him back to sketches from which he had removed the gestures connecting them. Su Wei, a blind music teacher, enters as a collaborator with her own recording project and her own mistakes: both artists have decided for others what is most valuable in their lives. Recognition arrives through concrete work: a library exhibition, reviews, a touring show negotiated with a commercial curator, a museum acquisition, and a reprinted catalogue.

\emph{Realization} (\path{856/story.md}, translated):
\begin{quote}\small
The commercial curator Jiang Bing reads the reviews of the small exhibition and comes to his door with a proposal for a three-city tour, setting its publicity theme as ``the miner's son no one understood for twenty years.'' Lin Zhiyao is tempted: he has waited twenty years, and what he most wants to hear is someone introducing him this solemnly. Su Wei reads the first draft of the publicity and asks him why he cut the exhibitions he has held, and cut Old Zhao and the nurse who bought his paintings. He quarrels with her, saying she does not understand how bitter his wait has been.

After the quarrel he rereads the draft and finds that, on this account, his father's ledger, Old Zhou's shirt and the two hundred yuan his senior classmate left on the table would all cease to exist. He goes back and negotiates with Jiang Bing point by point: he restores his real history, accepts a smaller scale, agrees to sell several works, and insists only that \emph{Ever Spring} be shown together with all its studies. The critic who named a price for a ``brush fee'' at 798 in 2007 now writes a long essay unasked, full of ``suffering'' and ``genius''; Lin replies courteously and asks him not to publish it.

Of the three cities, one stop is quiet, while another receives school group bookings for several days in a row; reviews are mixed. The provincial art museum acquires \emph{Old Zhou}, and the catalogue is reprinted.
\end{quote}
\emph{Outcome.} Local score 9.4. The outline turns ``eventual recognition,'' the prompt's least specified element, into a public process built from exhibitions, reviews, negotiation, and acquisition. Lin's psychological change appears in how he responds to that process: he learns to tell attention that looks into the paintings from attention that wants only a symbol, and to accept applause together with viewers who do not like his work.

\paragraph{Query 990 (Screenplay, wuxia calligraphy).}
\emph{Intent.} The prompt provides a sample script titled ``Shadow of the Sword: A Martial Arts Story,'' then asks for a wuxia-themed script in which the protagonist is a calligrapher-martial artist who ``engages in an unconventional battle with his opponents through the words he writes in calligraphy,'' blending calligraphy, martial arts, and philosophy across bamboo groves, marketplaces, and duel arenas, with three distinct enemies, a mentor, and a complex emotional arc.
\emph{Design.} The play treats written characters as techniques with reach, duration, and physical cost: \emph{zhi} (stop), \emph{feng} (edge), and \emph{jing} (still), with \emph{fang} (release) as the unfinished work. In the marketplace, Su Heng tries to send an edge stroke back at the archer; when a silk merchant rises behind him, he forces the stroke into an empty cart, severing a bolt of silk, and his wrist gives. Mo Liaoran's old silk holds only the left half of \emph{fang}. Su can write the whole character on a practice sheet, but there it is ``complete and ordinary,'' and its use in combat remains undiscovered. The three opponents have distinct motives: the swordsman Bai Liangzhou accepts the duel for a purse, a promise made to the schools, and an old bond; the archer Yan Wujiu is hired to stop any breach of the contest and watches whether its terms are kept; the administrator Han Jinglue needs the contest before the schools' representatives to bring six schools under a single banner. At the climax, release replaces the expected edge stroke.

\emph{Realization} (\path{990/script.md}, source-language English with character glosses):
\begin{quote}\small
Su stops before the last stroke of \emph{feng} (edge) and lowers the brush to the old silk instead.

The last two strokes of \emph{pu} (striking hand) leave the brush as one movement, the whole stalk at once, and Bai, who has read every character Su has made since they were boys, has no order to read. The third stroke runs under the pressure of the blade, and Su turns his whole body with it, giving up the center. The fourth, the golden knife, runs out to a needle's point in the empty air beside the pine. The completed \emph{fang} (release) spreads across the screen.

Su lets his stop go.
\end{quote}
\emph{Outcome.} Local score 9.4. Su abandons an edge stroke whose return would strike Han's men on the narrow stair. Release carries Bai over the drop; Su follows and pulls him back, and Bai lets his sword fall in order to climb. Bai's surrender, Yan's refusal to shoot a man who has yielded, and the unfinished compact in Han's case lead the schools to withdraw their forces.

\paragraph{Query 196 (Screenplay, absurdist campus play).}
\emph{Intent} (translated). Using provided research notes on Ionesco-style absurdist drama, campus employment, romance, family pressure, Kafka-like surrealism, youth internet culture, and campus drama examples, write an engaging absurdist play suitable for university performance: 4--6 main characters, 60--90 minutes in 2--3 acts, simple staging, absurd plus realistic elements, possible dream, stream-of-consciousness, or symbolic scenes, and language both youthful and literary.
\emph{Design.} The conception specifies a 60--90 minute campus play with symbolic figures, youth language, social-pressure choices, and alternating absurdist and realist scenes (\path{phase0_conception.yaml}). The structure plans the final confrontation to end in Mr.~System's ``complete victory'' through the graduation ceremony's capping ritual, Zhou Xiaobo's loss of speech at the podium, and an epilogue in which a new student puts on his jacket and Mr.~System says ``Next'' (\path{phase4_structure.yaml}). Zhou Xiaobo's character file, written at the character-design stage, reserves the nursery rhyme ``Dig-a dig-a dig'' for his graduation speech in the third act, the only moment in which he is allowed to fall silent, and makes his bright patchwork jacket the shell without which he loses his words.
\emph{Realization} (\path{196/script.md}, translated):
\begin{quote}
\small
\textsc{Zhou Xiaobo} (from the throat): ``\ldots Dig.'' The sound comes out. \textsc{Zhou Xiaobo}: ``Dig-a\ldots'' \textsc{Chorus} (mechanically, in unison, taking it up): ``\ldots dig-a dig.'' \textsc{Zhou Xiaobo}: ``Dig-a dig-a\ldots'' \textsc{Chorus}: ``\ldots plant tiny little seeds\ldots'' Zhou Xiaobo wants to say something else. Only this sound will come out. \textsc{Zhou Xiaobo}: ``Dig-a dig-a dig-a dig---'' \textsc{Chorus}: ``\ldots grow tiny little flowers.'' \textsc{Zhou Xiaobo} (into the microphone, louder and louder): ``Dig-a dig-a dig-a dig-a dig-a dig-a---'' He stops. The microphone slips from his hand and hits the table with a thud, which the speakers make enormous.
\end{quote}
\emph{Outcome.} Local score 7.8, the lowest in Table~\ref{tab:cases}. The collapse occurs as planned: everyone on stage freezes, the Chorus in their poses from Scene~2, no one laughs this time, and Mr.~System says, ``Thank you for your excellent speech. Next.'' The script's three acts, about 85 minutes, and five principal actors with a six-member chorus meet the prompt's constraints.

\paragraph{Cross-case findings.}
The cases connect broad creative requirements to decisions that shape an entire work. Courier obligations produce both an investigation and a witnessed handover (180); work experience and a familiar dish lead to a feasible next commitment (187); a scoring rule makes rescue costly (189); overlapping school commitments expose a cheerful character's insecurity (204). The stage works carry responsibility through public punishment and recognition (377), and through a printed correction, a conditional offer, and publication (535). The remaining cases develop a tier-by-tier campaign that ends in a shared containment (365), recognition built through exhibitions and negotiation (856), a released stroke that turns a duel into a rescue (990), and a campus character's loss of language (196).

The requested deliverable determines how these consequences unfold. An outline distributes them across chapters; a short story concentrates them in one evening; a character design supplies relationships and situations for later writing; a script coordinates dialogue, action, and performance.

\paragraph{Extended response excerpts.}
The excerpts below show how each response develops its central choices and their consequences. For long works, we select passages from key scenes. The table reports source and displayed character counts; Chinese-source passages are presented in English translation. Query 204 uses its paired English version for the source count. Chapter and scene labels identify nonconsecutive selections. A short commentary after each excerpt explains the craft choice the passages display.

\begin{center}
\scriptsize
\setlength{\tabcolsep}{3pt}
\begin{tabular}{@{}r r r p{.40\linewidth}@{}}
\toprule
Query & Source chars & Display chars & Excerpt mode \\
\midrule
180 & 45,148 & 9,302 & English source passages \\
187 & 8,219 & 4,832 & English source passages \\
189 & 35,150 & 8,350 & English source passages \\
204 & 28,696 & 5,652 & Paired English version \\
377 & 84,167 & 15,394 & English text and aria surtitles \\
535 & 86,071 & 23,982 & English source passages \\
365 & 36,852 & 8,461 & English passages; tier table as a list \\
856 & 10,674 & 7,629 & English translation of Chinese passages \\
990 & 68,919 & 19,224 & English passages with character glosses \\
196 & 29,391 & 11,069 & Direct English translation of selected passages \\
\bottomrule
\end{tabular}
\end{center}

\paragraph{Query 180 extended response excerpt.}
\emph{Novel outline: delivery notes that become an evacuation register, residents brought to the quay in person, and the thirty-first parcel delivered;} \path{180/story.md}.
\begin{quote}
\small
\textbf{1. The Closed Account}

\emph{Liang, 5B Tannery Lane. Knock twice. She feeds you whether you're hungry or not.}

At half past six Wen tapes a fresh coin to his handlebars and checks the chalk line on the lamp post under his window. He carries Mrs.\ Liang's food parcel across the middle district through a disused restaurant passage, and the owner takes a share of the cooking oil for the access. Wen argues, pays, and reaches Mrs.\ Liang with less than she expected. She makes him eat before he goes; he eats standing in the doorway. Then she asks why her son's medicine has come back again.

At the depot Wen learns that her household has been closed because Liang Cheng was transferred. The clerk offers to restore the allowance if Mrs.\ Liang reports for relocation herself. Wen has delivered to her flat often enough to know how little is in the cupboard. He takes a copy of the rejected slip and promises to find the office responsible, the first promise he has made that cannot be kept by handing something over at a door.

On his way back he takes a patient's request to the Meridian's clinic and watches Dr.\ Visser argue, without success, for permission to make home visits. The ship sails at dusk, and its next call is chalked on the quay wall. When Wen climbs to Mrs.\ Liang's landing that evening with the slip, the door opposite hers has a fresh red circle on it, and a radio is playing behind it.
\end{quote}
\medskip
\begin{quote}
\small
\textbf{10. The Register}

\emph{Assisted Evacuation Register, Eastern Shore and adjacent wards, entry 4471. Mobility: stairs, unassisted. Hearing: none. Source note: ``Fu, 3F, 22 Ferry Street. Deaf. Knock hard and wait a full minute. Lives alone.''}

Lin's letter is short. \emph{Know that I was asked to get your slips. Know that almost everything I told you about my manager was a lie.} Helion's security office gave her the task through the account dispute, and her report emptied the four shelters and described the wristband. She is writing because she wants to save what she can. With the letter are the complete transfer list, a weekly closure schedule that matches the Bulletin's figures exactly, and a copy of the Assisted Evacuation Register, dated three months before the Veil.

Wen reads the register at the dispatch table. It lists the households on the Eastern Shore and in the low wards that could not leave without help, each with a note copied from Swallow's records, and hundreds of the notes are his: \emph{Knock twice. She's slow on the stairs. Deaf. Knock hard and wait.} He had typed those phrases a hundred, a thousand times, for couriers he would never meet, and never once wondered who else might read them.
\end{quote}
\medskip
\begin{quote}
\small
\textbf{11. The Route Meeting}

\emph{Missing Letter to 22 Ferry Street, third floor: ``Grandma, they told us you were moved. If someone reads this to you, I am in the staff hostel in the Filtered District, room 16. Min.''}

The Meridian is due in three days, and Old Mail's members insist on deciding what happens next. Wen, who has never spoken at a meeting, says that the ship should receive people, because paper can be dismissed or turned against the people it describes, and people can be examined. Jiao asks who will keep the daily runs going if the best riders are all escorting his evidence. Mrs.\ Liang brings requests from the yard families through the Missing Letters. Ah-Sin wants an agreement about the children before he will help.
\end{quote}
\medskip
\begin{quote}
\small
\textbf{13. The Quay}

\emph{Missing Letter to Breakwater House, any room: ``My father is Gao Ming, eighty, from Canal Row. He can't read. Please read this to him: we are still at the old address.''}

At dawn the storm surge has broken the approach to the quay, and the way round runs along a road Helion controls. Wen cannot carry the injured, guide the children and reach the ship on time, so he splits the group. He sends Ah-Sin and an escort ahead with the children to the school, and the escort takes their names on to the doctor. Teng leads the walkers along the Helion road. Wen and Liang Cheng go back for two residents stranded at the broken approach.

Han has come to the quay to secure the equipment delivery. Old Hu and Lin have already brought copies of the papers to the medical team. Han offers to investigate the forged accounts internally and warns that any delay in unloading will endanger the hospitals. The warning is credible: one ward is already running on its generator. The captain will not hold the filtration parts indefinitely over papers he cannot authenticate.

Wen arrives with residents who appear in those papers as successfully resettled. Han does not dispute the papers. She explains the arithmetic instead: the surge that was coming, the cases she lost in court, the pumps that keep the hospitals running. Then she turns his own record against him. He fed eleven children at Pier 14 while a diabetic woman on Gull Road waited two days for her insulin, and he told those children to refuse beds; every route he has ever planned decided who would wait longest. She makes the same decision for four million people, and the grants that pay for it come from the doctor's own funders. Wen has never made a speech, so he reads addresses instead: each note he wrote that the register copied, and the name of the person it describes, who steps forward as he reads. Liang Cheng identifies the receipts he signed and explains why he signed them. Han offers him treatment inside the Filtered District. Mrs.\ Liang, who has come from the school at first light, asks whether she may stay with him and whether the others will receive the same care, which turns Han's offer to one man into a question she has to answer in front of everyone on the quay.

Dr.\ Visser tells the captain that the charity will not unload until its team has access to the yard and to Breakwater House and the people on the quay have been treated. The captain agrees to a short delay before the weather closes in again. Lulu's drawings show the medical team which landing the barges use, and Pan's copies and Lin's testimony go aboard. Han keeps command of her staff but loses exclusive control over whom the visitors can see.

Helion guards recognise Wen from the posters and detain him while the medical team receives the last residents. He has already decided not to run: arresting him cannot take back the people and records the doctor has received, and the others can complete the handover without him. Liang Cheng gives the doctor Wen's name, and Old Hu refuses to leave the quay until the place of detention is written down.

\textbf{14. Deliveries Resumed}

\emph{Caption on a drawing by Lulu: ``For the nurse who made Bao better. This is Bao. The fish is extra.''}

Two months later an outside inquiry has begun and Han is suspended pending its findings, while Helion still runs the waterworks, now under outside supervision. The Coastal Recovery Facility has frozen its grants while it works out what it paid for, which also delays the next shipment of filters. The clearance order is halted, though missing people remain unaccounted for, and thirty-seven deaths at Breakwater House are being examined again. The kits from Bonded Store 6 have finally been issued in the outer wards, and people wear the dated wristbands anyway. Food reaches more neighbourhoods than before, though not on any schedule people can rely on. There are arguments about which Helion staff should keep their jobs and who has profited from the new arrangements. In the Filtered District, where few people have read the inquiry's interim report, Wen is still the courier who sold the children's rice.

Wen is released after the medical team and the cooperative keep pressing his case. He comes back with an injured hand that keeps him off a loaded bicycle, and Jiao gives him dispatch work and asks him to stop correcting her routes from the table.

Liang Cheng lives with his mother again. He sleeps badly and at first will not talk about the receipts he signed. She has kept the letter Wen brought, including the parts that made her angry. Their reunion takes place over weeks of meals, appointments and ordinary disagreement. The Missing Letters now run from the school, and among the families they have traced is Gao Ming's, who brought him home from the breakwater to Canal Row.

The children are still together at the school. Ah-Sin goes to lessons and sometimes leaves the dinner arrangements to an adult, who manages them badly enough that he feels obliged to complain. He will not speak to Lin when she visits the school to help the inquiry match the children's names against the transfer list. Lulu asks Wen to deliver a picture to the ward nurse who treated Bao.

Wen walks the thirtieth parcel, the opened box from Bonded Store 6, to the inquiry's evidence clerk, who signs for it. The thirty-first takes him to a city ward, where Fu Lanying, brought off the breakwater by the medical team, is learning to use a hearing aid from the Meridian's clinic. Her granddaughter visits on Sundays. The batteries in Wen's parcel are the wrong size for the new aid, and Mrs.\ Fu tells him the parcel is six months late. He agrees that it is, and sits down. She keeps the box for her buttons.

On his way home he climbs to Mrs.\ Liang's flat to collect a parcel and stays to eat at her table while another courier takes the afternoon round. His bicycle is waiting downstairs for repair. For once, somebody else knows which deliveries are still outstanding.
\end{quote}

\emph{Commentary.} The outline makes the courier's delivery note the object that joins Wen's route, the mystery, and the resolution. Chapter~1 opens on a courier's note for Mrs.\ Liang, and chapter~10's epigraph prints Wen's note for Fu Lanying inside entry 4471 of the Assisted Evacuation Register, so the courtesies of the trade turn out to be the source of the list of households that could not leave without help. At the route meeting Wen argues that the ship should receive people, since paper can be dismissed and people can be examined; at the quay he answers Han in the same way, reading out each note he wrote and the register copied while the residents it describes step forward. The ending delivers the last of the thirty-one parcels he carried on the morning of the Veil, the chemical release that began the disaster: it reaches Fu Lanying six months late, and Wen, who before the Veil chose ``leave at door'' whenever the app allowed it, sits down with her.

\paragraph{Query 187 extended response excerpt.}
\emph{Short story: a hotel contract set against a father's twice-cooked pork, two offers to cook for people directly, and the meal that settles the choice;} \path{187/story.md}.
\begin{quote}
\small
\textbf{I. The First Cooking}

At eleven I come home from Eighth Avenue with a pound of pork belly, skin on, three clean layers of fat and lean, and a jar of Pixian doubanjiang I had no plan for. The Shanghai contract lies printed on the table, to be signed by six in the morning, and the salary would clear the last of the restaurant debt within a year. Tonight I read past the figure to the duties: menu development for thirty-one hotel dining rooms, supplier approval, quarterly tastings. Other people will cook what I write down, and I will taste it from a spoon.

[\ldots]

When I was nine he told me why the dish is cooked twice. In his grandmother's house, on the first and fifteenth of every month, a square of boiled pork was set before the ancestors with three cups of wine. They called it \emph{daotou}, the knife's head. When the incense burned down, the meat went back to the wok with bean paste and garlic shoots, and the children ate. ``The first cooking is for the ancestors,'' he said. ``They only look. The second is for whoever's hungry.''

\textbf{II. Cooling}

Since my West Village restaurant closed, I have worked breakfasts at Mrs.\ Lau's café in Sunset Park, a job I call temporary. I know the orders of men whose names I have never asked: congee without pork, eggs cooked hard, extra scallion packed separately for the drive. Last Sunday a bus driver who usually eats standing brought his daughter and sat down with her at a table. I warmed her bowl before filling it because she was taking so long with her coat.

Mrs.\ Lau saw me do it. Her café is dark five evenings a week, and twice she has offered me the kitchen for suppers. The second time she named a rent, because she could see I was treating the offer as kindness. I said I was waiting to hear about something.

``You still don't know?'' she said. ``Temporary people don't warm the bowls.''

[\ldots]

``With garlic shoots?''

``Leeks. Eighth Avenue had no garlic shoots.''

``Leeks are fine. Your dad used \emph{lianbai} in winter, remember? Big white pieces of cabbage. I used to pick them out.''

I don't remember the cabbage. For ten years I have told interviewers that my father's recipe never changed.

``The landlord will hold the shop until Friday,'' Hao says. He has sent me photographs of it, two lanes from where my father's shop stood, and one of himself under the extractor hood, pointing up, because he remembered my complaints about smoke. ``I can keep the accounts, but you'd have to cook. If you're coming, come. I can't keep asking.''
\end{quote}
\medskip
\begin{quote}
\small
\textbf{III. Back to the Wok}

Then I have it: the leeks. On Saturdays at two, with the shutters half down, my father used to fish the garlic shoots out of his own bowl and drop them into mine because, he said, he preferred the meat. Years later, when his teeth were bad, he went on giving them to me. And just as the grey slices curled in the fat and became lamps, the whole lane rises out of my bowl and grows solid: the bamboo chairs of the corner teahouse, ginkgo leaves stuck yellow to the wet pavement, the wooden rice barrel by his door with its lid tipped back and the chalked sign above it, \emph{Rice, as much as you want}. I see the builders who came on payday to \emph{da yaji}, to have meat, and stood up for a second bowl, and the one afternoon it snowed, when he left me with the washing-up and went out to look before it melted.

[\ldots]

I look at the contract. Is it wasting a life, to want a borrowed stove two nights a week and the same men coming back for a second bowl? And is it success to taste thirty-one versions of my own pork from a spoon and approve the ones that freeze well? My restaurant was a first cooking, meat set out to be looked at, and I don't think it was wasted: skip the boiling and you get salt-fried pork, a good dish, but not this one. I already know which burner on Mrs.\ Lau's stove is strong enough for a wok, though I have never cooked a supper on it.

I email Shanghai to decline and write separately to their chef, who taught me how a sauce changes when it is frozen, to thank him.

For Hao I record a voice message. I tell him I'm not coming, that he should let the landlord go, that I'm sorry I made him ask so many times. I say I'll come at Spring Festival and cook for his mother, with \emph{lianbai} if the garlic shoots are poor.

Mrs.\ Lau will be up at four, so I leave her a message asking whether her offer still stands: a month of Friday suppers, one dish, rice from the pot.

Near two Hao answers: \emph{All right. Call tomorrow anyway.} Then a second line: \emph{Leeks are fine.}

I divide the leftover pork between two containers, one for my lunch, one for Mrs. Lau to taste before we talk about the stove. The slices have set in their curl, each holding a little red oil, the way a lamp holds oil before anyone lights it. On the back of the contract's last page I write the first menu: \emph{Friday. Huiguorou, with whatever greens are good. Rice, as much as you want.}
\end{quote}

\emph{Commentary.} The story takes its decision frame from the method of huiguorou. In Part I the father explains the two cookings: the boiled pork is set before the ancestors, who ``only look,'' and the second cooking is ``for whoever's hungry.'' In Part III the narrator applies this distinction to her career: the closed restaurant becomes ``a first cooking, meat set out to be looked at,'' and because the dish needs its boiling stage, she can choose the borrowed stove while counting the restaurant as part of the same work. Two details from earlier scenes reappear in the resolution: the sign over her father's rice barrel, \emph{Rice, as much as you want}, becomes the last line of her first menu, and Hao's ``Leeks are fine,'' first said about her substitute for garlic shoots, closes his reply to her refusal.

\paragraph{Query 189 extended response excerpt.}
\emph{Wuxia match: learning to carry the Pearl, the rescue of Yang Kang, and the withheld force behind a losing finish;} \path{189/story.md}.
\begin{quote}
\small
Ouyang Feng extended his left hand and drove the Pearl toward the cliff with a low, swelling thrust of the Toad Skill. It fled before the pressure until it reached the stone, where the same force rebounded and held it quivering. His right hand closed around it and let go almost at once, two knuckles whitened with frost.

The thing had lain in water that never saw the sun, and it was yin through and through; force driven out of a palm is yang. It fled such force as fish flee a torch, and where a pushing hand closed on it, it drank the heat out of the flesh. Ouyang Feng rubbed his knuckles against his sleeve, watching it sink.

Huang Rong had expected another rush, but he waited, his breathing quiet, the staff laid along his forearm. Then he cupped his uninjured hand beneath the Pearl, and it settled there.

``Jing-gege,'' she called. ``He has it!''

Her father's flute shot toward Ouyang Feng's wrist. The Western Venom moved his open palm aside and parried with the staff, and the Pearl stayed on his skin. A master who kept his breath contained could carry it as readily as a child carries an egg; Ouyang Feng had learned that from one frozen touch, and now he had only to cross the field. But he could not strike. A blow with everything behind it rises through the whole body, through the carrying arm as well, and the Pearl would have felt it go. So every thrust of the flute he turned with the staff alone, and the force that might have broken Huang Yaoshi's guard stayed locked in his belly.
\end{quote}
\medskip
\begin{quote}
\small
The half-platform's blow had thrown Yang Kang off the sloping rope, and he vanished below Guo Jing. Guo Jing heard him hit the hanging frame and heard the breath knocked out of him. A scrap of white silk fluttered away from his shoulder, and when Guo Jing looked down, Yang Kang was hanging by one hand from a splintered crosspiece, his legs kicking over sixty feet of empty air.

The Pearl was coming down within reach of Guo Jing's right hand. He turned under the rope and caught his brother's wrist instead.

Their names had been chosen on the same snowy night, before either of them was born, by a Taoist who wanted both boys to remember the Jingkang years, when the Jurchen took Kaifeng and carried two Song emperors off to the north: Jing for the first half of that era's name and Kang for the second. Guo Jing had never once thought of letting go of the second half.

The weight wrenched his shoulder, and his left forearm, crooked over the hemp, took both of them. He tried to bring a knee up, but the broken frame was still swinging, and Yang Kang's sash, caught in its lashings, dragged him outward.

The Pearl sank past them both, a foot from Yang Kang's free hand, and Guo Jing saw the hand change. The fingers hooked and stiffened into the Nine Yin White Bone Claw, which Yang Kang had learned from a blind witch in the Jin capital, and snatched. The Pearl fled the claw as if it had been struck. Frost bloomed on three fingertips, the hand jerked back, and the jerk went through Yang Kang's whole body into Guo Jing's arm. Guo Jing's sleeve tore, and the hemp ground into the flesh beneath it.

``Your belt,'' Guo Jing said. ``Kang-di. Cut it.''
\end{quote}
\medskip
\begin{quote}
\small
Ouyang Feng came down onto the lowest rope ten feet from Guo Jing and walked along it, unhurried, the staff trailing. He could not take the Pearl by force, which would only drive it away from both of them, and he could not strike Guo Jing, since one drop of the young man's blood would hand Peach Blossom the match; but Guo Jing could still be made to lose the Pearl himself. He laid the staff along the rope behind him and held out his left palm, open, at the height of Guo Jing's chest.

Guo Jing understood him. Years ago, on this same island, Ouyang Feng had put him off the top of a pine tree with that palm. The palm had pushed, and Guo Jing had pushed back with everything he had, and in the instant of his pushing the force in front of him had vanished and he had gone over into the air after it. If that happened now he would have to catch himself with his hands, and the Pearl would be gone.

He put up his left hand, which still would not close; for this it did not need to.

Their palms met. Ouyang Feng's force came in like water under a door, rising, and Guo Jing answered with the first of the Eighteen Palms. He bent his knees, drew his arm round in a circle, and pushed: the Overweening Dragon Has Regret. In a pine wood, a long time ago, Hong Qigong had told him that the heart of the move lay in the regret, not in the overweening. Any oaf with a few hundred catties of strength can be fierce. The overweening dragon has cause for regret, for what is full cannot last, and so whatever goes out must be able to come back. Send out ten parts of force and keep twenty in reserve; the day you can taste the regret, you will have learned three-tenths of the move. It is like old wine, the old beggar had said: no bite on the tongue, and the strength comes after.

Ten parts went out through his left hand and twenty stayed in his belly and his heels; none at all went into his right arm, where the Pearl lay.

Ouyang Feng's force vanished, and this time Guo Jing did not go after it. He swayed on the rope and kept his feet, and the Pearl rocked in his palm and was still, as if the dragon under it had turned over in its sleep.
\end{quote}
\medskip
\begin{quote}
\small
Guo Jing's right hand was full, and his left could bear very little. The nearest firm landing was the referees' ledge, six feet above and to his side. He could open his right hand, let the Pearl go and catch the rope with both; the Pearl would drift, and Ouyang Feng, above him, was waiting for exactly that.

Another strand gave, and Guo Jing called, ``Rong'er.''

She had turned at the blind man's shout and was already coming down the ropes from the White Camel net, faster than he had ever seen her move.

``Come up!''

He bent his knee and sprang. For the short ascent he kept the Pearl on his open palm, his arm rising with his body, no force in it at all. His left hand caught the stone lip, but the injured fingers opened under his weight.

Huang Rong reached her staff beneath his armpit, braced it across the edge and pulled, and her father, landing beside her, took Guo Jing by the collar. His knees scraped onto granite. Below them the rope uncoiled from the iron eye and whipped away across the face of the cliff.

``You're off the field, boy,'' said Hong Qigong quietly. ``Cage it or give it to the wind.''

Ouyang Feng stood on the platform above with his frostbitten hand held against his chest, watching. If the Pearl went back to the wind, it would go to him.

Guo Jing crawled the last yard to the cage with the Pearl resting cold in his palm and tipped it through the little door. Hong Qigong rang the bell.
\end{quote}
\medskip
\begin{quote}
\small
Ouyang Feng came ashore by the surviving high rope. He looked at the tally bowls before he accepted the silver bowl and the length of brocade that went to the winning captain. The fingers of his right hand were still white at the joints and would not close on the bowl, and when Yang Kang came up, his master handed it to him to carry.

Yang Kang bowed, then turned and bowed to Guo Jing as well. A strip of Guo Jing's torn sleeve was still clenched in his fist.

``Give that here,'' said Huang Rong.

He obeyed, smiling. She knelt and bound Guo Jing's forearm with it, winding the cloth above the worst of the rope burn. He flinched when she pulled it tight.

``I couldn't hold on long enough,'' he said.

She tied the knot, then laid her hand in his open palm. ``Close your hand.''

He tried, and two fingers moved. Huang Rong kept hold of them and called Yideng over, and the monk knelt and set one finger on the inside of Guo Jing's wrist. The warmth of the One Yang Finger went up the arm like sunlight going down through cold water.

Hong Qigong brought out the dish of white powder and held it toward the losing captain with an expression of great innocence. Huang Yaoshi dipped two fingers in it and drew a stripe down each cheek, as calmly as if he were painting a plum branch, and went to the edge of the ledge and stood looking down at the ruined platforms with the powder still on his face.

Hong Qigong picked up the bowl that held his abandoned chicken, found it full of bamboo splinters, and set it down again.

In the cage the Pearl turned slowly, and its thread of gold caught the afternoon sun, bright enough to be seen from the highest terrace.
\end{quote}

\emph{Commentary.} The story gives the Pearl one physical rule: it flees force driven out of a palm, and it drinks the heat from a pushing hand that closes on it. The passages turn that rule into a constraint on the body of whoever reaches for the Pearl. Ouyang Feng can carry it only on an open palm with his breath contained, so he parries Huang Yaoshi with the staff alone; Yang Kang's Nine Yin White Bone Claw frosts his fingertips, and the jerk tears Guo Jing's sleeve; when Guo Jing meets Ouyang Feng's palm, Hong Qigong's teaching on the Overweening Dragon becomes a division of force across his body, with ten parts sent out, twenty kept in reserve, and none in the arm that holds the Pearl. This division realizes the controlling idea recorded in the conception artifact, that mastery shows in what a hand holds back, and the ending registers what the rescue cost through the two fingers that move, the strip of torn sleeve, and the white powder on Huang Yaoshi's cheeks.

\paragraph{Query 204 extended response excerpt.}
\emph{Character design: a three-layer insecurity, an interview that tests it, and the festival choice;} \path{204/story.en.md}.
\begin{quote}
\small
\textbf{2 $\cdot$ What he wants and what unsettles him}

There is a wish he has not yet put into words, even to himself: to be chosen first. When groups are formed, teams picked, or plans made, he wants to be the first person someone thinks of, even on a day when he is the least available.

His insecurity has three layers. The deeper the layer, the less he can see it.

\textbf{Hearing last.} This layer he can joke about. A meeting place changes in the group chat and the message happens to pass him by; he declines an invitation and someone else is found immediately. What bothers him most is a particular version of this: people he connected start getting along without him. He introduced Kentaro to Kuroda, and at their first meeting it was Haruto who steered the talk from basketball to a drummer's wrists. Last month the two of them went to a professional basketball game together, and Haruto only learned they had made plans when the group photo appeared in the chat. In that moment he felt himself fade to the color of his bedroom wall.

\textbf{Fine anywhere.} This layer he half sees. ``Saeki, join the smaller group,'' says the teacher. ``Haruto's fine with any shift, right?'' says Misaki. Three bands short a bassist all think of him. He takes ``fine anywhere'' as a compliment, and it is one. What frightens him is the other side of it: if he fits anywhere, no place in particular is missing him. So every refusal feels to him like handing back a membership card. He agrees too quickly. When class duties clash with rehearsal, he imagines he can catch up by being a little late to both. Asked whether he is busy, he says he can squeeze it in, until something is actually overdue. Part of this fear is accurate: without him, a band really can rehearse with a backing track, and it doesn't sound bad.

\textbf{Wanting is a burden.} This layer he cannot see. Asked ``What do you want?'' or ``Which song do you want to play?'', his mind goes blank for a moment and his mouth says, ``Anything's fine.'' Later he works out what the other person hoped he would want and quietly supplies that answer. His skill at reading people, the thing that makes him popular, comes from the same habit: as a child he checked whether the adults had any energy left before mentioning a disappointment; now he walks into a classroom and knows at once who is tired and who has been left out.
\end{quote}
\medskip
\begin{quote}
\small
\textbf{Okamoto Suzuka: someone he is working with}

While editing the interview recordings, she notices that across five recordings, interviewees ask him back, ``What about you, Haruto?'' seven times, and seven times he turns the question into a joke about someone else. Her first conclusion is that he doesn't take the magazine seriously. When he repeatedly promises material and delivers it late, she decides he enjoys talking more than following through. This hurts him, and she has evidence for it. Once he explains the actual scheduling conflict, she realizes what she missed. Her print deadline still matters, and she can refuse to finish all his work for him.

Those seven deflected questions give her an idea for a feature: a bassist who plays support for three bands. The magazine needs a proper lead article, and she wants to interview Haruto himself. For him this is the hardest invitation of all. He would have to tell a voice recorder which band he belongs to and what he wants to play, and the answer would be printed.
\end{quote}
\medskip
\begin{quote}
\small
\textbf{Volume 1: the school festival (April to October)}

In the week before the festival, his late arrival at rehearsal costs the band a run-through, and his unfinished table cards consume time Misaki had set aside for her friend. He starts with a joking apology. Misaki asks, ``Are you coming tomorrow or not? I need to put someone down.''

Now he has to choose, and someone will be disappointed whatever he picks. He stays with Before the Last Train, apologizes in person to the other two bands, and recommends Ono, the first-year he used to practice with, as their bassist. He gives up a café shift, hands over the work he has completed, and personally asks another classmate to cover for him. Misaki is still annoyed, and Ono sounds perfectly fine in someone else's band. What he has to resist that evening is the urge to take everything back on so that everyone will smile at him again right away.

Suzuka finally records her interview. When she asks which band he belongs to, he almost says ``public infrastructure'' again; what he actually says is, ``Before the Last Train. I arranged the bass for our closing song.'' That sentence will be printed in the magazine.

While they move equipment, Kentaro helps and complains that Haruto has ignored his messages. Haruto finally says, ``You didn't invite me to the game, either.'' Kentaro stares for a moment, then grins. ``Finally. I thought you'd never get mad at me.'' They agree that next time, whoever has plans will say so first.

The first volume can end by the stairs. Suzuka is carrying unsold magazines. Haruto has packed up his bass and had been planning to go help at the café.

``I can't carry anything else today.'' He looks at her box. ``Can we leave it in the clubroom? I'll borrow a trolley.''

``Yes. Hold the door for me.''

After she puts the box down, he asks whether she heard the performance. Suzuka says she was minding the magazine stall and only caught the first song.

``The last one had my arrangement in it,'' he says. ``I'll send you a recording.''

``Do. Fair warning, I'll tell you if it's bad.''

``Then I'll wait until you're in a good mood.''

He laughs again and starts telling her about the arrangement. Next time the group chat goes unanswered, he may still worry. Today he has made something he cares about, and he wants her to give it a proper listen.
\end{quote}
\emph{Commentary.} The design gives each layer of Haruto's insecurity a Volume 1 scene that tests it. Haruto tells Kentaro ``You didn't invite me to the game, either,'' which confronts the \emph{Hearing last} layer. He keeps one band and gives up the other two; because Ono sounds perfectly fine in someone else's band, the \emph{Fine anywhere} fear partly comes true and Haruto still makes the choice. Suzuka's count of seven deflected questions turns the hidden \emph{Wanting is a burden} layer into an interview, so the turning point is one printed sentence naming his band and his arrangement, and the closing scene by the stairs repeats that move when he tells her which song carried his arrangement.

\paragraph{Query 377 extended response excerpt.}
\emph{Peking Opera (Chinese verse given by its English surtitles; stage terms in pinyin): Xiao Qiao's farewell, the agreed flogging of Huang Gai, and the landing where Zhou Yu makes the plan public;} \path{377/script.md}.
\begin{quote}
\small
\textbf{A1 --- The Boat at the Steps}

\emph{Xiao Qiao enters with her water sleeves gathered, carrying a folded winter cloak, in the small quick steps of the dan (suibu). Zhou Yu meets her at the courtyard threshold.}

\textbf{XIAO QIAO:}
The boatman has loaded the household chests. He asks which bank we are going to.

\textbf{ZHOU YU:}
He should wait.

\textbf{XIAO QIAO:}
He has waited since morning. The families at the landing saw our baggage go down, and now they are loading theirs.

\textbf{ZHOU YU:}
I left orders for a boat to be ready for you.

\textbf{XIAO QIAO:}
It is ready. Shall I put the children aboard?

\textbf{ZHOU YU:}
Our lord has not decided.

\textbf{XIAO QIAO:}
And you?

\textbf{ZHOU YU:}
I have asked for the command.

\emph{She lifts the cloak to his shoulders and keeps one edge of it in her hand.}

\textbf{XIAO QIAO --- jiaoban:}
Husband --- ah!

\emph{The jinghu enters with the long introduction of the Erhuang manban. Her sleeve and gaze turn together as she sings toward the unseen landing (touxiu, the sleeve thrown outward), toward Zhou Yu (the sleeve folded back to the wrist), then toward the house behind her.}

\textbf{XIAO QIAO --- Erhuang manban:}

\emph{English surtitles: A boat waits below the steps, our chests already aboard. The child asks where home will be; the old servant leans by the landing and says nothing. Last night by the lamp I sewed your battle sleeves, and every stitch pulled at my heart. Now you wear the armour, and the words I prepared will not come.}

\emph{Zhou Yu reaches to fasten the cloak, and she lets him take its edge. The orchestra plays through the interlude while she turns once toward the landing and shakes her sleeves loose (douxiu): she has made up her mind.}

\textbf{XIAO QIAO --- Erhuang yuanban:}

\emph{English surtitles: You will meet the wind on the river; I shall watch for boats at our gate. If our family leaves first, the neighbours are left alone with their fear. Empty half the boat for medicine and bandages, and send the winter clothes down to the landing. When wounded men knock at our door, they will find a lamp.}

\textbf{XIAO QIAO --- Erhuang sanban:}

\emph{English surtitles: Let me fasten your cloak before you go. Come home and take it off yourself.}

\emph{She closes the clasp and lets go of it on the last note.}

\textbf{ZHOU YU:}
If the fighting reaches the landing, take the children inland. Promise me.

\textbf{XIAO QIAO:}
Send me a man who knows where we are. An order shouted across a river never reaches a house.

\textbf{ZHOU YU:}
I will send one.

\emph{Xiao Qiao calls offstage to the household: ``Half the chests come ashore. The cloth and the medicine go upstream with the supply boats.'' The council drum sounds. She steps to the edge of the courtyard; she does not enter the military assembly.}
\end{quote}
\medskip
\begin{quote}
\small
\textbf{The proposal}

\emph{[xiaokaimen]. Zhou Yu and Lu Su stand at the tent's edge, looking north across the river. Huang Gai enters with the heavy square step of the singing jing, stops and strokes his beard (lüran).}

\textbf{HUANG GAI:}
Commander, look at their fleet. Their numbers are great and ours are few; we cannot hold out against them for long. But they have joined their ships bow to stern. Ships like that can be burned, and the army behind them routed.

\emph{Surtitle: They are many and we are few; we cannot last. But their ships are joined bow to stern: burn them, and they will run.}

\textbf{ZHOU YU:}
Fire needs a way in. Wen Pin patrols every channel.

\textbf{HUANG GAI:}
Then I go in as a deserter, and Cao Cao opens the channel for me himself.

\textbf{ZHOU YU:}
You have served three generations of the Sun family. You are the last man he would believe.

\textbf{HUANG GAI:}
That is why he will believe it, if he hears of a quarrel before my letter reaches him.

\emph{An attendant enters.}

\textbf{ATTENDANT:}
Commander, a gentleman from Jiujiang has crossed the river. Jiang Gan, styled Ziyi. He says he is your old schoolfellow.

\textbf{ZHOU YU} \emph{(three laughs, rising in pitch: ha ha, ha ha, a ha ha\ldots)}

\textbf{LU SU:}
Commander, why are you laughing?

\textbf{ZHOU YU:}
Cao Cao has sent a persuader, on the very night we need a witness.

\textbf{HUANG GAI:}
Then let the persuader see something with his own eyes.

\textbf{ZHOU YU:}
At my table he is my friend, and at my table I will tell him the truth. Outside it, he is Cao Cao's eyes. Let the eyes have work to do. \emph{(He speaks into Huang Gai's ear; the audience sees only the old man's nod.)} At the second watch, come quietly. At the third, come loudly.

\textbf{LU SU:}
And what shall I do?

\textbf{ZHOU YU:}
Drink, Zijing, and look happy.

\textbf{A5 --- The Command Seal}

\textbf{ZHOU YU --- Xipi yuanban:}

\emph{English surtitles: The watch drum beats the night away; the command seal is cold as frost in my hand. My old friend sleeps and snores. How could he know tomorrow is a battlefield? We read poems together as boys; tonight I am borrowing his eyes. The old general offers his plan and his back, and gives his loyal heart to the river.}

\emph{During the instrumental interlude Huang Gai appears at the tent door with a lantern, as arranged for the second watch. He sets it down, takes off his outer armour and lays it across the foot of the bed. Zhou Yu watches and does not help him.}

\textbf{ZHOU YU --- Xipi erliu:}

\emph{English surtitles: I hold the authority to order it. Who will bear the wound for Gongfu? Tomorrow every officer will watch, and the blame must fall on me. Let news of the rod cross the river; behind the fire-ships, tie fast boats for the escape. When the fire rises, everyone leaves the ships together, and I will count the men home boat by boat.}

\textbf{ZHOU YU --- Xipi sanban:}

\emph{English surtitles: Gongfu, receive my bow. Come back from the river, and we will drink together.}

\emph{Zhou Yu bows low. Huang Gai returns the bow and speaks in a whisper.}

\textbf{HUANG GAI:}
Tomorrow, whatever they say, keep your place behind that table. When they carry me past you, let them carry me.

\textbf{ZHOU YU} \emph{(whispering)}:
And the boats?

\textbf{HUANG GAI:}
Fast boats tied behind every fire-ship. And a good oarsman in mine.

\emph{He picks up his armour and lantern and slips out. Zhou Yu lies down again and goes back to snoring.}
\end{quote}
\medskip
\begin{quote}
\small
\textbf{The morning muster}

\emph{[jijifeng]. The four attendants and four armoured guards enter and stand in files. Lu Su, Cheng Pu and the officers stand below the desk. Zhou Yu enters in full armour, takes the command seal from the desk and sits.}

\textbf{ZHOU YU --- dingchang shi:}

\emph{English surtitle: Drums and horns ring at the camp gate; heroes stand guard on either side.}

\emph{An attendant places Jiang Gan, with his bag, at the outer edge of the assembly, where he can see and hear everything.}

\textbf{ZHOU YU:}
Cao Cao's camp stretches three hundred li along the river. He will not be broken in a day. Each captain will draw three months' grain and prepare to hold.

\textbf{HUANG GAI:}
Three months? Three years would make no difference! If we can break him this month, break him. If we can't, we should take Zhang Zibu's advice: throw down our armour and kneel to the north.

\emph{Lu Su starts, and the cymbal marks it with one sharp stroke.}

\textbf{ZHOU YU:}
I hold our lord's order to destroy Cao Cao. Whoever speaks of surrender again shall die.

\textbf{HUANG GAI:}
Hold your tongue! I followed General Sun Jian across the whole southeast. I have served three generations of this house. Where were you then, yellow-haired boy?

\textbf{ZHOU YU} \emph{(rising; he strikes the desk)}:
Old fool! With two armies face to face, you dare say this and shake our soldiers' hearts? If I do not take your head, I cannot command this army. Take him out and behead him!

\emph{Two executioners seize Huang Gai. Cheng Pu steps forward and kneels.}

\textbf{CHENG PU:}
Hold the blade! Commander, Huang Gai has served three generations of the Sun family. Spare him.

\textbf{ZHOU YU:}
General Cheng, would you break my law?

\textbf{CHENG PU:}
I would break my own knees first. Spare him.

\emph{Lu Su kneels beside him; then every officer kneels. Zhou Yu looks along the kneeling line, and his hand closes over the command seal.}

\textbf{ZHOU YU:}
For the officers' sake, his life is spared. Remove his command token. One hundred strokes of the rod!

\textbf{HUANG GAI:}
Strike, then! I'll count them myself.

\textbf{The flogging --- acted to percussion}

\emph{Two soldiers take off Huang Gai's outer armour and guide him to kneel at centre, facing the audience, his back to the tent. The rod swings through empty air a forearm's length behind him; each stroke is a crack of the clappers and a cymbal. No rod touches his body.}

\emph{Huang Gai counts aloud in a rough voice: ``One! ... Two! ... Three! ...'' At ``ten'' the drummer takes over the count, the strokes falling faster than any voice could follow, every tenth one marked by a single stroke of the large gong. With each gong Huang Gai's shoulders drop lower, and he forces them back up through the following phrase.}

\emph{At the fifth gong, the fiftieth stroke, he falls forward onto his palms, and his long beard swings out and down. Cheng Pu takes off his helmet and kneels before the desk; Lu Su throws himself across Huang Gai's back, and the rod stays raised above them.}

\emph{Zhou Yu half rises, his hand reaching toward the edge of the desk. He stops, closes his hand on the seal, and sits again.}

\textbf{ZHOU YU:}
For the officers' sake, the other fifty are held over. If he is insolent again, they will fall. Take him out!

\emph{Two soldiers lift Huang Gai under the arms and carry him past Jiang Gan, who clears the way with unusual speed. Huang Gai looks once at the empty peg where his command token used to hang. Jiang Gan follows the look.}

\textbf{HUANG GAI:}
Take me to my own quarters.

\emph{They go, and the officers follow without saluting the commander. Cheng Pu goes last, still carrying his helmet in his hand.}

\textbf{JIANG GAN:}
Gongjin, I shall trouble you no further.

\textbf{ZHOU YU:}
My boat will take you back.

\textbf{JIANG GAN:}
I have one waiting.

\emph{He hurries out before an escort can be offered. Left alone, Zhou Yu reaches toward Huang Gai's confiscated token on the desk, stops short of it, and leaves it there. [saotou].}
\end{quote}
\medskip
\begin{quote}
\small
\textbf{A8 --- Calling the Roll}

\textbf{CAO CAO --- Erhuang sanban:}

\emph{English surtitles: When we came, our banners covered the river; as we go, mud covers our saddles. I call for the forward camp: who answers? Only the cold wind past my ears. I saw Huang Gai's pennant with my own eyes, and I ordered the channel opened. Wen Pin told me, and I would not listen; now the river is on fire, and regret comes too late. The men who filled the ditch were the sick and the weak, and in the mud I do not see them rise. Tomorrow I will open the tent and call the officers. How many names on the roll will have no one to answer?}

\emph{He waits after the last phrase, as though an answer might still come from the rear.}

\textbf{CAO CAO:}
Ziyi. Take your brush. A letter to Sun Quan.

\emph{Jiang Gan kneels and spreads a scroll on his knee.}

\textbf{CAO CAO} \emph{(dictating)}:
``At Red Cliffs there happened to be sickness in my army. I burned my own ships and withdrew, and so let Zhou Yu win a name he did not earn.''

\emph{Surtitle: At Red Cliffs my army was sick; I burned my ships and withdrew, and let Zhou Yu win an empty name.}

\textbf{CAO CAO:}
I am not ashamed to retreat.

\emph{Jiang Gan writes, then stops with the brush in the air.}

\textbf{JIANG GAN:}
Chancellor. The men in the mud. Shall I write them as ``sickness'' too?

\emph{[sijitou]. Cao Cao rounds on him, beard flying. Then he slowly raises his long sleeve and covers his face with it (yanmian), hiding the white face. The orchestra stops playing; he holds the sleeve there through the silence, then lowers it.}

\textbf{CAO CAO:}
Write it as I said. Seal it. \emph{(Quieter.)} And give me the roll. I will read the names myself.

\emph{Jiang Gan hands him the bamboo slips. Cao Cao turns to Wen Pin.}

\textbf{CAO CAO:}
Ride, Zhongye. You warned me twice.

\emph{He gives the whip to the wounded Wen Pin and walks beside the horse, reading the slips, toward the north. Jiang Gan follows with the sealed letter, looking back once more at the place where the two soldiers knelt. The road rhythm thins into the creak of oars.}

\textbf{The southern landing}

\emph{Dawn. Xiao Qiao stands at the landing with a lantern, bandages and a pot of warm water; two women of the household work beside her. Wounded men sit along the edge of the stage. Xiao Qiao unfolds the winter cloak she fastened on her husband at Chaisang and spreads it over two of them who are shivering. Old Shen comes in from upriver, wet to the knees, and scans every face. Shen An enters, supported by a soldier, with a cut on his arm.}

\textbf{OLD SHEN:}
You! \emph{(He swats the boy with his hat, then pulls him close.)} I told you: the place of danger!

\textbf{SHEN AN:}
General Huang went into the water. I couldn't reach him.

\emph{Two soldiers carry in a soaked old man wrapped in a torn sail, an arrow shaft standing from his shoulder, his face turned to the floor. No one knows him. They lay him at the far edge, among the unknown men taken from the river.}

\textbf{XIAO QIAO --- Erhuang yuanban, as she moves among the wounded:}

\emph{English surtitles: Boats come back to the steps again. Help the wounded up first.}

\emph{Cheng Pu enters fast from the river, his armour blackened, searching.}

\textbf{CHENG PU:}
Gongfu! Has anyone seen Huang Gongfu?

\emph{No one answers until a weak voice comes from the far edge.}

\textbf{HUANG GAI:}
Demou ...

\emph{[sibian]. Cheng Pu stops dead.}

\textbf{CHENG PU:}
That is Gongfu's voice!

\emph{He runs, kneels and turns the old man over. It is Huang Gai. Cheng Pu breaks into a sobbing cry (kutou): ``Gongfu, ah!'' He strips off his own war-robe and wraps Huang Gai in it, then tears a strip from a banner to bind the shoulder. Xiao Qiao kneels with the warm water.}

\textbf{HUANG GAI:}
The last boat?

\textbf{OLD SHEN:}
It reached the lower steps. Two men need carrying.

\textbf{XIAO QIAO:}
Take the litter down.

\emph{Zhou Yu enters with Lu Su. Huang Gai tries to rise.}

\textbf{ZHOU YU:}
Sit, Gongfu.

\textbf{LU SU:}
Cao Cao has burned the rest of his ships and fled by the Huarong road. Liu Bei's men are following him toward Jiangling with ours. The captains say more than half his army is gone: fire, water, hunger and fever.

\emph{A soldier comes in carrying a box.}

\textbf{SOLDIER:}
Silver from the northern stores, Commander. The men thought it should go to your tent.

\textbf{ZHOU YU:}
To the army treasury, with the captains as witnesses. The wounded have a claim before my tent does.

\emph{The soldier bows and goes. Cheng Pu draws Huang Gai's command token from his belt and lays it in the old man's palm. Huang Gai looks at it and does not close his hand.}

\textbf{HUANG GAI:}
Will the men know?

\emph{Zhou Yu steps in front of him, where every soldier on the landing can see him.}

\textbf{ZHOU YU:}
Huang Gai proposed the fire attack. His surrender was our plan. He took the rod with my agreement. Let every ship hear who brought us through tonight.

\emph{The soldiers salute Huang Gai. He tries to return it from where he sits; Cheng Pu holds the robe off his arm so that he can finish the gesture. Then Cheng Pu stands, puts on his helmet, and turns to Zhou Yu.}

\textbf{CHENG PU:}
Gongjin. Being near you is like drinking good wine: a man is drunk before he knows it.

\emph{Surtitle: Keeping company with Zhou Gongjin is like drinking mellow wine; you are drunk before you notice.}

\emph{Zhou Yu bows to him. This time Cheng Pu returns the bow in full.}

\textbf{HUANG GAI:}
You promised me a cup.

\textbf{XIAO QIAO:}
Warm water first.

\emph{Everyone on the landing laughs, and an attendant brings wine. Zhou Yu takes the first cup without drinking, goes to the edge of the stage and pours it slowly onto the river.}

\textbf{ZHOU YU:}
The first cup for those who did not come back. Theirs and ours.

\emph{Everyone on the landing bows toward the river.}

\textbf{HUANG GAI --- answering Xiao Qiao's phrase, Erhuang yuanban:}

\emph{English surtitles: Where the battle flags are furled, we know each other's faces. Comrades are still waiting at the landing.}

\textbf{ZHOU YU --- Erhuang yuanban:}

\emph{English surtitles: When the victory report is written, keep the lamp burning. Go down to the water and ask after one more boat.}

\emph{Old Shen lifts his oar toward an offstage call and goes down to guide the next boat in. Xiao Qiao takes up the lantern and goes with him.}
\end{quote}

\emph{Commentary.} The script carries the cost of the stratagem through objects that pass between scenes. At the muster Zhou Yu closes his hand on the command seal while the officers kneel and stops short of Huang Gai's confiscated token; at the landing Cheng Pu returns that token, Huang Gai leaves his hand open around it and asks whether the men will know, and Zhou Yu answers by naming the plan before every soldier present. The arias of the two commanders share a counting motif: Zhou Yu's Xipi erliu ends with counting the men home boat by boat, Cao Cao's Erhuang sanban ends with names on the roll that no one will answer, and the closing couplet ends by going down to the water to ask after one more boat. The cup Zhou Yu promises Huang Gai in A5 returns as the first cup poured onto the river ``for those who did not come back.''

\paragraph{Query 535 extended response excerpt.}
\emph{Stage play: a disputed comma, a conditional offer, and the finished poem reaching a reader;} \path{535/script.md}.
\begin{quote}
\small
\textbf{Scene 1 --- The correction}

\emph{April 1929. WANG's commercial printing shop off Henan Road. The desk becomes a proofing table; the gramophone gives way to the rhythm of a press. Wedding cards printed with the double-happiness character lie on one bench, a dentist's circular in English on another. A pile of newly folded copies of SPRING MUD awaits wrappers. LIN compares his returned manuscript with the printed page.}

LIN
Mr. Wang. A comma has appeared after ``thirteen.''

WANG
Mr. Su's pencil. Saturday.

LIN
It makes ``thirteen'' hang there without its flowers.

WANG
Then they can wait. He paid for the correction.

LIN
There was nothing to correct.

WANG
You should have told him that on Saturday.

\emph{LIN shows him both pages. WANG looks, with the patience of a craftsman being asked to account for another man's decision.}

WANG
Same type body, same spacing. I took out a blank and put in the comma. If I take it out now I must unlock the forme. The dentist is next. He is English and pays on Fridays, which is more than poetry does.

LIN
How many copies?

WANG
Three hundred. A hundred have gone to the distributor. Two hundred are here. The paper alone cost---

LIN
I meant how many copies can be corrected.

WANG
All two hundred, if you wish to buy my afternoon.

\emph{LIN searches his pocket and finds only a few coins and a tram ticket.}

WANG
Mr. Lin, I don't argue with poets. They can argue all day. My men go home at six.

LIN
Have you been paid for this issue?

WANG
Half.

\emph{LIN puts the coins away and, with his pen, makes a small correction in one copy.}

LIN
Will Mr. Su be here?

WANG
After four. I've left his marked proof in the drawer. Tell him I need it back if he takes it. He says ``I never changed that'' when the bill comes.

\emph{WANG brings the proof. The pencilled comma is unmistakable, and lower on the page there is another small alteration; LIN compares the two marks.}

LIN
He has moved a stop here as well.

WANG
Editors like pairs. One correction looks accidental.

\emph{The press resumes. LIN stacks his corrected copy on five clean copies and wraps the bundle to take to the café.}

\emph{The café counter lights up. LARISSA takes the bundle while LIN stays beside the counter, reading SU's pencil marks. ZHU enters carrying school exercise books.}

ZHU
The new \emph{Spring Mud}, please.

LARISSA
You are in luck. It has just come in.

\emph{ZHU takes the top copy---the one LIN corrected. She lays her exercise books beneath it, opens at the poem, and reads. Her hand returns to the crossed-out comma.}

ZHU
Has this copy been damaged?

LIN
Improved.

\emph{Only now does she look at him.}

LARISSA
The author.

ZHU
I know. I use some of your poems with my older pupils.

LIN
Then I apologise to your older pupils.

ZHU
They're very hard to injure with a poem. Why did you change it?

LIN
I restored it. The editor supplied a breath where I had enough already.

ZHU
Do all the copies have your correction?

LIN
Only this one. Mr. Wang put a price on the others.

\emph{ZHU sets down the marked copy and takes a fresh one from the bundle. She checks the comma, then the stop lower on the page. LIN watches her hands.}

LIN
If you want my signature, you needn't buy two.

ZHU
I would like a clean one for class.

LIN
You prefer the editor's flowers.

ZHU
I prefer a page the whole class can read alike.

\emph{She pays. LIN holds out the corrected copy.}

LIN
Take it as well. Show them that authors can make mistakes about their own poems.

\emph{She hesitates, then accepts it.}

ZHU
May I ask you to leave the shop copies as they are?

LIN
For your pupils?

ZHU
For the people who have already bought theirs.

\emph{She puts both magazines into her satchel. LARISSA counts out change.}

LARISSA
You have given me half a yuan.

ZHU
Keep enough for his tea.

LIN
I can pay for my own tea.

LARISSA
Today, perhaps.

\emph{ZHU smiles despite herself. LIN almost returns it.}

ZHU
Then put it towards tomorrow's.

\emph{She leaves with the exercise books. LIN reaches for another magazine, but stops before marking it.}

LARISSA
You frightened her.

LIN
Did I?

LARISSA
She came to buy something she liked. You told her it was wrong.

LIN
Does she come often?

LARISSA
For the magazine. She sometimes stays to mark books. She drinks one cup and never asks me to turn down the music.

\emph{LIN folds the marked galley into his coat.}

LARISSA
Should I put the copies on the shelf?

\emph{He looks towards the door through which ZHU left.}

LIN
Wait until this evening. I'll come back.

\emph{He goes without his tea. LARISSA moves the magazines away from the samovar's steam.}
\end{quote}
\medskip
\begin{quote}
\small
\textbf{Scene 5 --- The statement}

\emph{March 1931. The drawing room of a Western-style house on Route Stanislas Chevalier, in the French Concession, which Du Yuesheng lends to his adviser. French chairs, a tiled fireplace, a gilt clock under a glass dome; on the mantelpiece beside it a small porcelain Guanyin and a dish of Buddha's-hand citrons. Scrolls in the host's own calligraphy. On a side table, a game of Chinese chess abandoned halfway.}

\emph{BAI, in a tailored suit, studies a scroll. YANG DU, in a padded scholar's gown with a string of wooden beads round his wrist, sits by the fire with a rug over his knees. He coughs now and then and makes nothing of it. A SERVANT shows in SHUYI and LIN, and brings tea.}

YANG DU
Mrs. Shen. And the poet. Forgive me for not getting up; my lungs have stopped taking instructions. Mr. Du asks you to consider the house your own. So do I, since it isn't mine either.

SHUYI
It's kind of you to lend it, Mr. Yang.

YANG DU
I lend it to everyone. That is my profession now.

BAI
\emph{(turning from the scroll)} We recited your ``Song of Young Hunan'' at school, sir. ``If China today is Greece, Hunan shall be Sparta.''

YANG DU
Everyone recited it. I was a young man who believed one province could save the country. Then I believed an emperor could. Then I believed the Buddha could.

LIN
And now?

YANG DU
Now I believe in mustard plasters. \emph{(He looks at LIN.)} You are the one who wrote about the tide.

LIN
Before April twenty-seven.

YANG DU
Ah. In April twenty-seven Mr. Du's men wore armbands with the character for ``worker'' on them. You need not say it; I can see you would like to. In a time of changing winds a man may be four or five people in one life. I have been most of them. The art is in choosing which of your selves you sign for.

\emph{He gets up, with effort. The SERVANT comes to his elbow.}

YANG DU
I am writing my own funeral couplet. At my age it is only prudent. I have the second line: ``To heal the people and save the country, others will rise after me.'' The first line is harder. It has to account for what I was. \emph{(Passing LIN, lightly.)} Old men read the back of a document first. Then the front. Then they rest.

\emph{He goes out on the SERVANT's arm. BAI offers LIN his hand, and LIN takes it. BAI speaks the easy Japanese of their student years.}

BAI
\emph{Hisashiburi da na,} Lin-kun.

LIN
Seven years.

BAI
You kept the blue gown.

LIN
It's a different gown.

BAI
Then you're making progress.

\emph{They sit. While SHUYI pours, BAI moves the chessboard aside to make room for an envelope.}

SHUYI
The school, please, Mr. Bai. My husband says you have been very good.

BAI
Your husband is generous. \emph{(He takes out a letter and a printed form.)} Yip Yin Academy, Kowloon. A private school. They want a teacher of modern Chinese literature. Housing included, passage paid. The first year's salary is here.

\emph{LIN reads the letter.}

LIN
That is more than I have earned in the last two years.

BAI
It is a real post. I haven't hired actors to sit at desks for you.

SHUYI
There are girls as well as boys. They sent a prospectus. Your poems are in one of the classes already.

LIN
You've read it?

SHUYI
I wanted to know where they expected you to sleep.

LIN
They want a collection under the school imprint.

BAI
You have enough poems. Mr. Du would take the first hundred copies. There would be a reading before you sail, and a short announcement in the \emph{North-China Daily News}. Something for the parents to see.

\emph{He gives LIN the printed form.}

LIN
This is longer than an announcement.

BAI
The first paragraph is the announcement.

\emph{LIN reads silently. SHUYI keeps her eyes on the paper and away from his face.}

LIN
``During my association with \emph{Spring Mud}, political matter was introduced into the publication without my knowledge.''

BAI
A Chinese text will be attached. Mr. Xu's own secretary drafted it. He writes better Chinese than any of us.

LIN
``Without my knowledge.''

BAI
When did you first know?

\emph{SHUYI sets the teapot down before it can overflow the cup.}

LIN
Is this an interview?

BAI
It could remain a conversation. If you sign, you have a useful explanation of your work. Your editor made changes; you were occupied with poetry. A great many people will find that easy to believe.

LIN
And Su?

BAI
Mr. Su's affairs are his own.

LIN
Except for the sentence I sign about them.

BAI
You haven't named him.

LIN
His name is on every masthead.

\emph{BAI takes a cigarette from his case and offers one. LIN refuses.}

BAI
You'll have heard about February.

LIN
The papers had nothing in them.

BAI
The papers are sensible. Twenty-four, at Longhua, on the night of the seventh. Most of them were taken at a meeting in the Eastern Hotel in January. Five of them wrote poems or stories. One was twenty-one; he had translated a life of Pet\H{o}fi for Lu Xun's magazine. I read their files. I did not sign them. I should like never to read yours.

LIN
``What is hope? A harlot.''

BAI
Pet\H{o}fi died on a Cossack lance at twenty-six. Poets make wretched conspirators, Lin-kun. They want to be read. \emph{(He taps the proof under the form.)} Do you remember Lu Junyi?

LIN
The Jade Unicorn.

BAI
A fortune-teller gets him to write four lines on his own wall. Read the first character of each line downwards: Lu, Jun, Yi, Rebels. He hadn't the faintest idea. His own steward read it out in the magistrate's court. Jin Shengtan writes in the margin that the four lines are used three times: to lure him out of his house, to push him into the water, and to keep him from ever going home. Your magnolia has managed the first two.

SHUYI
My husband quoted the \emph{Water Margin} too, last night. ``When one man rebels, nine generations of his clan are cut down.'' He meant our boys.

BAI
Mrs. Shen, I am offering your brother the third chapter with a different ending.

\emph{LIN turns the form over.}

LIN
There's a paragraph on the back.

BAI
A routine undertaking.

LIN
``To assist, when required, in establishing the circumstances of the alterations.'' Who put the comma in. Who carried the proof. Which printer kept it.

BAI
You needn't supply information you don't possess.

LIN
You don't know what I possess. You want my agreement before you ask.

\emph{SHUYI holds out her hand for the form. LIN gives it to her.}

SHUYI
Strike that paragraph.

BAI
I can ask for a revision.

SHUYI
Ask whom? You brought it. You're here.

BAI
The people who will take responsibility for letting your brother travel.

SHUYI
And you take none?

BAI
Mrs. Shen, I have put my name to his application. That isn't nothing in my office.

\emph{She studies him and sees that he means it.}

SHUYI
Then write beside his name that this will be the last question.

\emph{She moves the pen towards BAI. He leaves it on the table.}

BAI
I cannot bind an investigation that has not ended.

LIN
You can bind a poet who hasn't begun his teaching.

\emph{BAI rises and goes to the sideboard. He pours water rather than brandy.}

BAI
In Tokyo you used to say our country's trouble was that a man couldn't cross a room without knowing everyone's grandfather. I'm offering you a room where nobody knows yours. You won't have it.

LIN
You have put my grandfather in the contract.

BAI
Your sister is here because she asked for you to have a future. I am here because I remember you helping me down a stair with a broken ankle. I can remember that and still believe your friends are ruining this country.

LIN
You've read their minds?

BAI
I have read the instructions they send men who can't afford to refuse them. I have read the lists afterwards. You think every signature at the foot of a page means willingness?

LIN
You've just offered me a signature of that kind.

BAI
Yes. One I think you can survive.

\emph{SHUYI picks up the school letter and reads it from the beginning.}

SHUYI
The house has a kitchen. It doesn't mention servants. He would have to learn to cook.

LIN
Elder sister---

SHUYI
Let me finish reading something in this room that means what it says.

\emph{BAI comes back to his chair.}

BAI
Keep the letter. The post begins in September. I need an answer by the seventh of August, so the school has time to engage someone else.

LIN
Five months.

BAI
It is what I could arrange. I suggest you decide sooner.

\emph{A knock. The SERVANT enters.}

SERVANT
Inspector Li, sir. With the printer's papers.

\emph{BAI's attention goes at once to the door.}

BAI
The side room. I'll come presently.

\emph{LI appears behind the SERVANT with a small packet of proofs. He sees LIN.}

LI
Mr. Lin. I have admired your work.

\emph{LIN recognises the pale yellow cover under the packet's string.}

BAI
In the side room, Inspector.

\emph{LI withdraws. BAI waits until the door has closed.}

LIN
You made him come through here.

BAI
He was told to use the other door.

LIN
Is there always another door in a house like this?

BAI
Usually.

SHUYI
Where did he get those proofs?

BAI
A printer kept papers against an unpaid account. Some were forwarded with a statement. I can't discuss the witness.

LIN
Wang?

\emph{BAI's face does not change. SHUYI puts the pen down sharply.}

SHUYI
Brother. No more names.

\emph{BAI closes his cigarette case.}

BAI
I have an appointment. Take the papers. You can return them through Mr. Du's secretary. I will not call at your room.

\emph{He goes into the side room. SHUYI gathers the letter and the form separately, gives LIN the form and keeps the school letter a moment longer.}

LIN
I'm sorry.

SHUYI
For which part?

LIN
For bringing you here.

SHUYI
I brought you. My husband thinks he is clever because he found a classmate willing to help. I thought I was clever because I asked for a real school.

LIN
It is a real school.

SHUYI
Yes. That's the worst of it. I could see you in that kitchen, burning rice. I was going to send you a proper pot.

\emph{She folds the prospectus into its envelope.}

SHUYI
If you go, will you tell them what they want?

LIN
I don't know what they'll want next.

SHUYI
You know something now. I watched you almost give them a name.

LIN
He already had it.

SHUYI
You didn't know that before you said it.

\emph{Beyond the side door BAI and LI exchange a few indistinct words. LIN lowers his voice.}

LIN
I have to warn Su.

SHUYI
Then we leave separately. I'll keep Mr. Yang's secretary at the chessboard. I am very slow at chess.

LIN
Like the brothers at the---

SHUYI
I don't know any brothers but you. \emph{(A beat.)} You read the salary twice.

LIN
Yes.

SHUYI
Good. I don't want to go home with the only lie they haven't written for me.

LIN
Would Father have wanted me to sign?

SHUYI
Father wanted you home for the New Year. He died wanting it. I have no other message from him.

\emph{She smooths the back of his gown where it has caught under his collar.}

SHUYI
I'll go first. Wait until you hear me lose a chariot.

\emph{She goes. Soon her voice can be heard in the hall, politely insistent, asking the secretary to explain again how the cannon jumps. LIN takes the envelope and goes out by the other door. The SERVANT clears three cups. YANG DU comes back in, slowly, looks at the abandoned game on the side table, moves one piece, and sits down by the fire with the rug over his knees.}
\end{quote}
\medskip
\begin{quote}
\small
\textbf{Scene 8 --- The page that remains}

\emph{Before dawn, Friday 7 August. Pages cover the desk; the ferry poem has gone back into the notebook. A small lamp is enough to work by; the rest of the room is dark. LIN reads from a draft.}

LIN
``The brass cools in the hand of the dead.''

\emph{He crosses the line out, lifts the paperweight off the next page and sets it on the floor by his chair.}

LIN
He never had it. I had it.

\emph{He looks at the crossed-out line.}

LIN
When they executed Ji Kang, his friend Xiang Xiu walked past the empty house, heard a neighbour playing the flute, wrote a few lines about it and stopped. When I was young I thought Xiang Xiu was lazy.

\emph{He reads further down.}

LIN
``I held the door---''

\emph{He stops. His hand grips the edge of the desk as it gripped the back door in Zhabei.}

LIN
``I held the back door open. / She pulled my fingers from it.''

\emph{He writes. There is a knock, and SU comes in with a parcel of food and a printer's envelope.}

SU
Four o'clock. The printer sent these. He wants the copy written clearly, and nothing added after six.

LIN
There's nothing in it he doesn't already know.

\emph{SU puts the food on the bed and starts to gather the pages. LIN covers them with his arm.}

LIN
Listen first.

\emph{SU takes the chair opposite. LIN sorts the manuscript into two sections and a short last leaf.}

LIN
``Errata.''

SU
You can't call a poem ``Errata.''

LIN
It's the only notice of mine the \emph{Shenbao} ever printed.

\emph{He reads the first section. The light opens a little on the empty café counter.}

LIN

\begin{verse}
For \emph{spring}, read \emph{March, and the rent unpaid}.\\
For \emph{a woman left me}, read \emph{I let her go, and called it freedom}.\\
For \emph{thirteen}, read \emph{thirteen}. The tree was not the error.\\
In the café a Russian girl counted my coins twice\\
and wrote what I owed her very small.\\
Beyond the glass a boy ran with the evening paper,\\
selling a death, the price of cotton, rain.\\
I folded the paper small enough to hide the death\\
and wrote about a tree.\\
For \emph{the poet stood across the road}, read \emph{hid}.\\
My sister kept a place for me among the bearers;\\
the coffin was the length of my arm.\\
For \emph{Tides}, read \emph{a boy of seventeen on Baoshan Road}\\
\emph{who learned my lines backwards, setting them in lead,}\\
\emph{and said them forwards, walking towards the church}.\\
\end{verse}

\emph{SU looks up from the page.}

SU
You've put the waitress in.

LIN
She was there.

SU
She'll know herself.

LIN
She'll know I still owe her. That much can survive an inquiry.

\emph{SU turns to the second section.}

SU
This is longer.

LIN
It takes the space of two stanzas on page seventeen. I counted it against your layout. The advertisement goes to the back.

SU
The advertiser paid for seventeen.

LIN
Pay him back out of my fee.

\emph{SU reads the first line silently, and stops.}

SU
It's my sentence.

LIN
Read it aloud.

\emph{SU reads. He begins as an editor checking copy; the labour of saying it changes his voice.}

SU

\begin{verse}
For \emph{our comrade at the case}, read \emph{Qiao},\\
forty-eight, who promised his son in Wuxi a bicycle\\
every winter, and every winter wrote \emph{next year}.\\
For \emph{completed his duty}, read:\\
he put the names in a bucket and they would not sink;\\
he pressed them under with his thumb, and they rose,\\
and he pressed them under.\\
For \emph{before his death}, read \emph{while I stood on the back step}\\
\emph{with my hand on the frame, and a woman pulled my fingers from it}.\\
His shoulder held the door a little longer.\\
I did not hear what he said. I have tried\\
every sentence a man might say at a door,\\
and none of them is his.\\
\end{verse}

\emph{He stops. LIN waits until he can go on.}

SU

\begin{verse}
Let that line stay empty.\\
For \emph{a traitor}, read \emph{a printer whose son came home}.\\
For \emph{a printer}, read also \emph{a poet who said his name}\\
\emph{in a warm room, over very good tea}.\\
For \emph{we}, wherever it appears, read \emph{he}, and \emph{I}.\\
\end{verse}

\emph{SU puts the page down. Dawn has begun at the window.}

SU
``For \emph{we}, wherever it appears.'' That's my editorial.

LIN
It's every editorial.

SU
``A woman pulled my fingers from it.'' They could ask which woman.

LIN
They've been asking since before we wrote this.

SU
And you've named him.

LIN
I've named him what the bench called him. His son has another name for him.

SU
If it reaches the boy, he may hate us.

LIN
Will you print it?

\emph{SU sets the two sections together without answering yet, and turns to the short leaf.}

SU
And this?

LIN
Below the imprint. The closing lines. They go with the poem when anyone copies it out.

SU
There's room in five-point type.

LIN
Make it readable. Take out the list of forthcoming titles. There won't be any.

\emph{SU marks the layout. He strikes out the advertisement and writes a refund in the margin.}

SU
I'll print it. We can read the proof together at the shop.

LIN
I won't be at the shop. Bai told them I sail on the eighth, and I do: to Ningbo on the night boat, with a letter to a cloth merchant who needs his complaints put into English. My sister found him. I shall begin with damaged bolts.

SU
You always hated commercial translation.

LIN
I may still hate it. \emph{(A beat.)} I'm not a monk, Su. I haven't the freedom to sit down and die where they find me. I like being alive. So I'm going. Read the proof against these pages, all the way to the end.

SU
I will.

\emph{LIN writes his name beneath the whole poem. SU sees it.}

SU
Your full name.

LIN
It's been on the magazine for six years.

\emph{He signs the printer's authorisation as well. SU wraps the manuscript, leaving the copy LIN has made in his notebook, and puts the parcel on the bed beside the untouched food.}

SU
And Bai?

\emph{LIN takes the statement and the school letter. On a separate sheet he has already written a reply.}

LIN
``I cannot attest to this account. Please withdraw my application.'' My sister will leave it with the secretary after I've gone.

\emph{He reads BAI's note once more. Then he tears the statement slowly: in half, in quarters, smaller. He says, to no one in the room:}

LIN
Thank you.

\emph{He drops the pieces into the cold stove, folds the school letter back into its envelope and packs it in his case.}

SU
Wanyin sent word last month, through Wen's people. She asked whether you were writing.

LIN
Is she well?

SU
She said she was teaching. That's what I know.

\emph{LIN takes the pressed petal from the notebook's endpapers. It has broken at one edge, and he puts it back carefully.}

LIN
If you can answer her, tell her I've written something I saw.

\emph{SU takes the manuscript. At the door he turns, as if the old farewell between friends might still be possible.}

SU
Lin---

LIN
You have to reach the printer.

\emph{SU goes. LIN packs the dictionary, the notebook and the brass paperweight, then takes the notebook out once more. From the street comes the first call of a vendor. Standing by the case, he reads the closing lines.}

LIN

\begin{verse}
Reader, if this page has reached you,\\
the night was longer than we thought.\\
I wrote once that despair is as vain as hope.\\
It is. Neither of them carried the bag.\\
The night is long, and the road is long;\\
it would be easier to forget, and say nothing.\\
But someone, if not I, will remember them\\
and say their names again. Let it be you.\\
Where we had no time to breathe, put in the comma.\\
Where we lied to stay alive, write the truth in the margin.\\
Do not die for this page. Live, and correct it.\\
And when no one needs it any more, let it grow old.\\
\end{verse}

\emph{He closes the notebook and puts it in the case. He eats a mouthful of the food SU brought, wraps the rest for the journey, and goes.}

\textbf{Epilogue}

\emph{Darkness. The NEWSBOY's voice, as at the beginning.}

NEWSBOY (off)
\emph{Shenbao! Shenbao!} Japanese troops in Mukden! Mukden taken! \emph{Shenbao!}

\emph{Light rises slowly on Café Renaissance, a morning in late September 1931. The plane trees have begun to yellow. The chairs are still upturned on every table but one, by the window. LARISSA, in her apron, sets the chairs down one by one. A copy of the last SPRING MUD lies beside the till. A second copy lies on the window table, next to a glass of tea nobody has drunk.}

\emph{A SCHOOLGIRL of seventeen comes in with a satchel, out of breath.}

SCHOOLGIRL
Is there still a copy of the last \emph{Spring Mud}?

LARISSA
The last one?

SCHOOLGIRL
Our teacher told us to read it. Miss Zhu. Before she went.

LARISSA
Where did she go?

SCHOOLGIRL
They said her mother was ill.

\emph{LARISSA looks at her, then gives her the copy from beside the till. The girl holds out coins. LARISSA counts them twice and gives half of them back.}

LARISSA
Half. \emph{(She nods towards the window table.)} Sit there. Nobody uses it.

\emph{The SCHOOLGIRL sits at the window table. She opens the magazine at the back, below the imprint, and reads; her lips move. She takes a pencil from her satchel and, after a moment's thought, makes a small mark in the margin.}

\emph{The NEWSBOY's cry passes the window and goes away down Avenue Joffre. A tram bell. LARISSA turns over the record on the gramophone. The waltz from the beginning plays while the girl reads.}

\emph{Blackout.}

\textbf{End of play.}
\end{quote}

\emph{Commentary.} The play stages its central choice through the procedures of proofreading. In Scene 1 Lin objects that the editor's comma ``supplied a breath where I had enough already''; in Scene 5 Bai's statement asks him to sign that political matter entered \emph{Spring Mud} ``without my knowledge,'' and Lin reads its back paragraph as a request to name ``Who put the comma in. Who carried the proof. Which printer kept it.'' Scene 8 answers that form with an errata notice: each ``For \emph{X}, read \emph{Y}'' line in ``Errata'' replaces a public phrase with the person or act it covered, and Lin signs the poem with his full name before tearing up the statement. The closing lines reverse his first objection (``Where we had no time to breathe, put in the comma''), and the Epilogue shows a reader taking up that instruction when the schoolgirl makes a small mark in the margin.

\paragraph{Query 365 extended response excerpt.}
\emph{Zombie-cultivation outline: the premise and tier list, the first herd fight, the Cartographer's trap, and the confinement of the indestructible Immortal;} \path{365/story.md}.

\begin{quote}
\small
\textbf{Premise}

Eight years after the Upwelling, Yiyang keeps two million people alive inside a ring of ash wards laid along its railways and storm drains. The wards work like levees: they push the corpse-qi that seeps out of the ground away from the inner districts and let it pool in the outer ones, which the Reclamation Council calls spillways and lists as empty.

Lin Han, twenty-five, is a bounty hunter at the tenth level of Qi cultivation. He takes contracts in the spillways to pay for his own advancement and for his sister's treatment. His sister, Lin Yue, nineteen, repairs the wards for the Survey Office. A salvage run sends them after the same anomaly from opposite ends: there are too many strong corpses outside the walls. By the ladder's own odds, Yiyang's twelve million dead should have produced one cyan-pupilled corpse, but hunters have filed reports of three. The wards are breeding them, and someone outside has started stacking bodies on the ward anchors to push the pooled qi back toward the inner city.
\end{quote}
\medskip
\begin{quote}
\small
\textbf{The ladder}

``Black eyes'' and ``white eyes'' refer to the eyeball around the iris; the second colour is the pupil. Hunters read the pattern first and the terrain second, because each rung adds a different kind of problem.

Tier \textbar{} Eyes \textbar{} Rarity \textbar{} Equivalent human stage \textbar{} What the rung adds.

\textbf{Basic} \textbar{} Black eyes, black pupils \textbar{} The herd \textbar{} None; ordinary people can kill one \textbar{} Hunger and herd movement. Weak alone; in numbers they climb walls on each other's backs.

\textbf{Level 1} \textbar{} Black eyes, red pupils \textbar{} 1 per 10,000 basic \textbar{} Ability Level 1; Qi Condensation, tenth level \textbar{} Reinforced bone; the habits of the living body.

\textbf{Level 2} \textbar{} Black eyes, yellow pupils \textbar{} 1 per 300,000 \textbar{} Ability Level 2; Foundation Building \textbar{} The first projected ability: force pushed through matter.

\textbf{Level 3} \textbar{} Black eyes, green pupils \textbar{} 1 per million \textbar{} Ability Level 3; Late Foundation Building \textbar{} Regrowth while rooted in water, soil or flesh.

\textbf{Level 4} \textbar{} Black eyes, cyan pupils \textbar{} 1 per 10 million \textbar{} Ability Level 4; Core Formation \textbar{} A solid core; shapes stone at scale; compulsions left over from a trade.

\textbf{Level 5} \textbar{} White eyes, red pupils \textbar{} One among many Level 4s \textbar{} Ability Level 5; Yuan Ying \textbar{} A corpse-infant that abandons a ruined body and takes another.

\textbf{Level 6} \textbar{} White eyes, yellow pupils \textbar{} Mutates from Level 5 \textbar{} Ability Level 6; Soul Exit \textbar{} Awareness sent along the qi network, speaking with the voices it finds there.

\textbf{Level 7, S-class} \textbar{} White eyes, green pupils \textbar{} Rare; the first ``Demon Corpse'' \textbar{} Ability Level 7; Soul Division \textbar{} Mature hunting intelligence; runs several bodies at once; bargains and studies prey.

\textbf{Level 8, SS-class} \textbar{} White eyes, cyan pupils \textbar{} Extremely rare \textbar{} Ability Level 8; Integration \textbar{} Fuses with the ground and the dead around it; draws on any connected anchor.

\textbf{Level 9, SSS-class} \textbar{} White eyes, black pupils, like a living person's \textbar{} A supreme ruler \textbar{} Ability Level 9; Great Achievement \textbar{} A body that passes for human; lesser corpses flee or obey; rules territory and collects tribute.

\textbf{Zombie King} \textbar{} Vertical pupils \textbar{} 1 per 100 million Level 9s \textbar{} Ability unknown; Tribulation Crossing \textbar{} Calls down the lightning tribulation.

\textbf{Earth Demon Immortal} \textbar{} The King's true form after the lightning \textbar{} Only a King that survives \textbar{} Beyond the ladder \textbar{} Cannot be harmed.
\end{quote}
\medskip
\begin{quote}
\small
\textbf{Chapter 1 $\cdot$ The Depot --- Basic Zombies (black eyes, black pupils)}

Late summer, Year 8. Pumping Station 3 will fail without the batteries in a flooded bus depot in the Ninth District, a spillway. Han and Wei escort fourteen civilian loaders. Basic corpses are the rung ordinary people can kill, and the loaders do the killing: they work in pairs from doorways with hooked poles while the hunters watch the flanks.

Then Han spots drums of diesel in the service yard and goes for them: the fuel bounty would buy his Foundation pills. With the flank open, the herd finds the loaders' line. The corpses climb the bus sheds on each other's backs and spread across the corrugated roofs like mould. Wei pulls them around the sheds with the truck's horn. Han gets back in time to cut the return route open, but the truck stays behind, nose-down in floodwater with its engine still warm. The batteries reach Yiyang on handcarts, and the lost truck costs the crew its next contract.

Logging the depot for the Survey Office, Yue finds wet black sand in a maintenance passage that should end at a concrete footing. Her count of the herd is three times what the district's own dead can account for.

\textbf{Cliffhanger:} That night the abandoned truck starts up beyond the ward. In the gate floodlights, the corpse in the driver's seat has red pupils. It works the gears with the economy of a man who drove that route for twenty years, and it turns toward the clinic that took the batteries.
\end{quote}
\medskip
\begin{quote}
\small
\textbf{Chapter 8 $\cdot$ The Cartographer --- Level 7, S-class Demon Corpse (white eyes, green pupils)}

The Cartographer is the first opponent with mature hunting intelligence. It holds the map archive of the municipal library, where it reads the Council's ward plans and steers lesser corpses toward weak routes. It learned the survey shorthand from the notebooks of a team lost in Year 6. At Soul Division it runs three bodies at once: a reader at the archive desk, a hunter in the stacks and a negotiator in a librarian's lanyard. It has studied Han's habit of saving the nearest victim.

The negotiator brings terms: withdraw the survey teams and leave the outer districts alone. The hunters agree to a meeting, and meanwhile Yue, free since the depot evacuation, organises the outer shelters the Council says do not exist. The shelters move out along several routes, each group carrying enough to continue without Han. The Cartographer detects the evacuation and releases its prisoners into two separate killing grounds, betting that Han will run to the nearer one. Han stays where he is and trusts the teams already assigned there.

The fight takes down the library's lower floors. Han pins one of the three bodies while Yue's team cuts the qi line feeding the reading room, and Shen destroys the core after a second hunting party blocks its escape. Most of the prisoners survive, but three hundred people die in a shelter that appears on no list. In the archive the crew finds the Cartographer's own survey: a boundary around the entire Ninth District, walked and marked at five posts with claw strokes in a script nobody can read.

\textbf{Cliffhanger:} The Cartographer's working map of the lost shelter is a Council planning sheet, with a note beside the shelter in Shen's handwriting: \emph{Occupied, approx. 300. Proceed with spillway.}
\end{quote}
\medskip
\begin{quote}
\small
\textbf{Chapter 12 $\cdot$ Five Pillars --- The Earth Demon Immortal}

It keeps the ledger's law: a ruler owns what it walks. At dusk it sets out to pace Yiyang's boundary and claim the city and everyone in it. It hunts Han along the way, because the half-tribulation in his body would complete its crossing. Yue's plan uses what the survey has paid two years to learn: underground, corpses navigate by the currents along rail, pipe and wet stone, and her crews can bend those currents. Overnight they re-lay the ash lines so that every current curves back on itself. Han runs ahead as bait, burning his stolen stage to stay out of reach. All night the Immortal believes it is pacing the edge of the city. At what it takes for the last corner it finds five pillars holding up a haze of qi, scores its claim on them in the grave-sea script and leaves its mark at the foot of the first. Then it returns to its starting point to close the boundary.

Han, nearly burned out, shows it where it has been: the five pillars are the anchor columns of the Chorus Hall basement, three hundred metres from where it began. It has claimed one city block, and a claim binds both ways: it can draw only on what lies inside. Before it can walk a new boundary, the crews drop the five columns and the hall comes down on it in a mound. Yue sets the last ward plate on the top stone, and the rubble takes root and closes its seams. Han cuts the line beneath his feet and falls into the net Wei's crew has rigged below.

\textbf{Closing hook:} Han takes the keeper's post. Every month he carries the copper down the shaft, and every month the Immortal talks to him through the stone, as patient as the driver on its route. It offers to restore his circulation and his ascent, and to settle the half-tribulation it says he still owes the sky. Ke Lan finishes translating the ledger's last entry on sealed Kings: such a King is freed by one who goes down to it of his own will. The next spring, Han goes down with the copper and does not come up until morning.
\end{quote}

\emph{Commentary.} The outline treats the trait each rung adds in the tier list as the condition its chapter's fight must meet. Basic corpses are the rung ordinary people can kill, so in Chapter~1 the loaders do the killing, and Han's detour for the fuel bounty opens the flank the herd comes through; in Chapter~8 the Cartographer, from the rung that ``bargains and studies prey,'' builds its trap on Han's habit of saving the nearest victim, and the chapter turns on Han staying where he is. The last rung ``cannot be harmed,'' and Chapter~12 resolves it by confinement: the ledger's law that a ruler owns what it walks, combined with the crews' ability to bend the underground currents, leads the Immortal to claim a single block, and the claim lets it draw only on what lies inside before the hall is brought down on it.

\paragraph{Query 856 extended response excerpt.}
\emph{Chapter outline: a father's care misread and later recognized, a fault the painter shares with his kindred spirit, and recognition negotiated against a publicity story;} \path{856/story.md}.

\begin{quote}
\small
\textbf{Story background}

Over the same period, the yardstick by which people measured a painting changed several times: the mine's bulletin boards wanted festivity; the dealers of the 1990s wanted what would sell; Beijing's avant-garde circles said that realist painting of mining districts was outdated; among critics it was customary to accept a ``brush fee'' for writing favorable reviews; and the old people of the housing quarter asked only one question: ``Who is it a painting of?'' Between these yardsticks, the protagonist's paintings are rejected and misread again and again. At first he is convinced that the world does not understand; only later, through the people in his paintings and a kindred spirit who cannot see paintings, does he find his own standard: a painting must do right by the people in it, must show real skill of the hand, and must hold up under the gaze of strangers.
\end{quote}
\medskip
\begin{quote}
\small
\textbf{Chapter 1. The Painting Turned to the Wall (Autumn 1992)}

\emph{Main plot.} Lin Zhiyao returns to Beizhang with his graduation work. At the exhibition for the mine's fortieth anniversary he hangs \emph{Father's Cigarette}: his father, off shift and still in his work clothes, sits on a bench outside the bathhouse, coal embedded in his wrinkles, the cigarette tip the brightest point in the whole painting. On opening day the union chairman objects that the man in the painting ``looks as if he had just been fished out of the pit'' and has the painting turned around, leaning face to the wall in a corner. Passing miners tease Lin Dashan: ``Your son has painted you as a ghost.'' The old painter Pang Dehou makes a different remark: the face and the clothes are crowded into the same patch of gray, and from a few steps away they cannot be made out. Lin Zhiyao remembers the first remark and takes the second as politeness.

His father carries the painting home on his shoulder, finds old calendar paper, wraps the four corners of the frame, and hangs it on the wall where he changes clothes every day. He only asks his son why he chose to paint him when he looked his worst. The son says he paints the truth. The father says that winning a game of chess yesterday was also the truth, and tells him to paint that too when he has time. The son takes it as just a joke.

\emph{Psychological change.} The certainty he had at graduation is wounded. He concludes that ``they can't understand it,'' and from then on he neither explains nor asks anyone to look at his paintings. Every day his father reaches around the frame for the clothes hook; Lin decides that his father is ashamed to face himself in the painting, and does not notice the four corners wrapped in paper. At this stage his loneliness still carries pride.
\end{quote}
\medskip
\begin{quote}
\small
\textbf{Chapter 4. The Palette Knife (2003--2007)}

\emph{Main plot (third paragraph).} Back in Beizhang, he does not touch a brush for more than twenty days. One night in the last month of the lunar year, he picks up a palette knife and scrapes the seventeenth version of \emph{Smoke} and two other paintings down to the bare canvas, then goes downstairs to his father's wall. As he raises the knife to \emph{Father's Cigarette}, he sees the old calendar paper wrapped around the four corners of the frame, still printed with the months of 1992, and his hand stops in midair.
\end{quote}
\medskip
\begin{quote}
\small
\textbf{Chapter 8. Not That Version (2010)}

\emph{Main plot.} The two collaborate on ``Portraits and Songs of the Mining District.'' Su Wei insists on keeping the old songs as they are and chooses the take in which Old Zhou's breath is weakest and he coughs twice in the middle; Lin Zhiyao paints a portrait of Old Zhou that emphasizes his illness, with an oxygen tube trailing in from the lower left corner. Having heard the recording and seen the painting, Old Zhou refuses both, saying it is like inviting everyone to come and pity him. He wants the version he sang to an accordion when he was young, and he wants the portrait to have a different shirt.

Privately, both think the old man is vain. As they argue, they discover that they are giving the same reasons: each is deciding for someone else what is most precious. Su Wei carries the accordion on her back to record again, accepts Old Zhou's changes to the lyrics and asks him to tell where the changes came from; Lin Zhiyao goes several days later, sits in Old Zhou's home and looks at the Dacron shirt ironed until it shines. The old portrait stays in the studio; the new one goes into a later exhibition.

\emph{Psychological change.} In Su Wei, Lin Zhiyao sees his own old failing, and this works better than anyone's criticism to his face. The word ``truth'' loosens in his mind: how a person is willing to be seen is also that person's truth. Su Wei changes her view too: the person making the recording should not presume to understand the song better than the person singing it.
\end{quote}
\medskip
\begin{quote}
\small
\textbf{Chapter 12. The Paintings Leave Beizhang (2015)}

\emph{Main plot (first two paragraphs).} The commercial curator Jiang Bing reads the reviews of the small exhibition and comes to his door with a proposal for a three-city tour, setting its publicity theme as ``the miner's son no one understood for twenty years.'' Lin Zhiyao is tempted: he has waited twenty years, and what he most wants to hear is someone introducing him this solemnly. Su Wei reads the first draft of the publicity and asks him why he cut the exhibitions he has held, and cut Old Zhao and the nurse who bought his paintings. He quarrels with her, saying she does not understand how bitter his wait has been.

After the quarrel he rereads the draft and finds that, on this account, his father's ledger, Old Zhou's shirt and the two hundred yuan his senior classmate left on the table would all cease to exist. He goes back and negotiates with Jiang Bing point by point: he restores his real history, accepts a smaller scale, agrees to sell several works, and insists only that \emph{Ever Spring} be shown together with all its studies. The critic who named a price for a ``brush fee'' at 798 in 2007 now writes a long essay unasked, full of ``suffering'' and ``genius''; Lin replies courteously and asks him not to publish it.
\end{quote}
\medskip
\begin{quote}
\small
\textbf{Chapter 14. Painting Again Next Year (2017)}

\emph{Main plot.} \emph{Ever Spring} is exhibited at the provincial art museum. Former neighbors charter a bus to come and see it. Some misremember their own seats, some insist that they are the one serving food in the painting, and some say that life in the new apartment blocks is more comfortable than the days in the painting. Lin Zhiyao argues with them for a while, then simply asks everyone to write the meal as they remember it on a card and post it beside the painting. Young visitors read these short memories, turn back to the painting, and see the people around the same table looking in different directions. New exhibition invitations arrive one after another; some people come only to buy a catalogue.

He and Su Wei return to Beizhang, each keeping a studio in their new apartment. His father's old chair occupies the best spot in the painting studio; Su Wei, complaining that he always uses it as a model, pulls it over and sits in it to choose records. Lin Zhiyao picks up his sketchbook. She asks how long she has to sit, saying she must go out in half an hour. He says she need not wait; what he does not finish, he will paint tomorrow. Hearing a page turn, she reminds him that today she is wearing the shirt he thinks too loud. He says he knows; that shirt is exactly what he is drawing. Outside the window the new apartment blocks have no chimneys, and every home has its lights on.

\emph{Psychological change.} He no longer waits for the world to come and warm itself at his fire. It is good when people look at the paintings; when no one does, he will still paint tomorrow.
\end{quote}

\emph{Commentary.} The outline marks the painter's changes through objects that return in later chapters and are read differently on their return. In Chapter 1 Lin takes his father's reaching around the frame as shame and overlooks the calendar paper on its corners; in Chapter 4 the same paper, still printed with the months of 1992, stops the palette knife. Chapter 8 states the principle that governs the later choices, that how a person is willing to be seen is also that person's truth. In Chapter 12 Lin restores his real history to the tour publicity once he sees that the ``no one understood'' story would erase his father's ledger, Old Zhou's shirt and the two hundred yuan, and in the final scene, twenty-five years after treating his father's chess game as a joke, he draws the shirt Su Wei is wearing although he finds it too loud.

\paragraph{Query 990 extended response excerpt.}
\emph{Wuxia stage play: the unfinished character on Mo's silk, the archer's test in the market, and the duel that the completed character ends;} \path{990/script.md}.
Character names are romanized; Chinese characters in the script are given in pinyin with an English gloss.

\begin{quote}
\small
\textbf{WHERE THE BRUSH MEETS THE BLADE}

\textbf{A Wuxia Stage Play in Three Acts}

\textbf{Company and setting (selected)}

\textbf{SU HENG, twenty-four.} Known in the Lin'an market as the Brush of Cui Feng. Six years in Master Mo's bamboo grove have made him an exceptional calligrapher and martial artist. He fights with a long brush carried in a cinnabar sash and has never drawn a sword in earnest.

\textbf{MO LIAORAN, seventy.} His teacher. By the market's account, his brush once held a gateway against a lord's guards; this morning he cannot tie his own sandal. He has told his student many times to leave, and has made sure he stayed.

\textbf{BAI LIANGZHOU, twenty-six.} Called Three-Cup Bai. A wandering swordsman who busks and drinks in the market. Su Heng's sworn brother: both learned the stick, the sword and their first characters in Old Pei's yard, and only Su was admitted to Mo's grove. Bai reads a little and writes his own name badly.

\textbf{YAN WUJIU, thirty.} A hired archer, called Still-Water Yan. Patient, economical and careful about the ground beyond his target. He carries six dark arrows and a seventh with a white shaft.

\textbf{HAN JINGLUE, forty-five.} Administrator of Lin'an. His private martial association is known in the tea shops as the Shadow Pavilion. He intends to bring the independent schools under his protection and command. He writes a faultless official hand.

\textbf{THE STORYTELLER.} A woman who tells stories to the clack of pear-wood clappers on a platform beside the tea shop. She also carries broth, medicine and news between the city and the grove.

\textbf{ENSEMBLE.} A stallkeeper; MOTHER CHEN, who keeps the tea shop; a silk merchant named Zhang; porters, customers, Han's aide and two guards. The company later becomes six representatives of the schools at the peak: an old scholar of the West Lake Academy, the elder of a sword hall, the master of the Iron Oar ferrymen, a garrison captain, a woman of the Twin Saber Hall and a young recorder who keeps the schools' register.
\end{quote}
\medskip
\begin{quote}
\small
\textbf{The art of the brush (selected paragraphs)}

Every student begins with yong, \emph{eternal}, because its strokes contain the eight movements of regular script. In Mo's grove each movement is also a movement of the body. Ce, the dot, \emph{the tilting stone}: a dropped weight, a stamp of the heel. Le, \emph{the rein}: a horizontal check. Nu, \emph{the crossbow}: a vertical drive. Ti, \emph{the leap}: the hook that kicks back at the foot of a vertical. Ce, \emph{the whip}: a short rising flick. L\"{u}e, \emph{the sweep}: a long fall to the left. Zhuo, \emph{the peck}: a short, quick one. Zhe, \emph{the rend}: the broad stroke falling to the right, which the old masters called the golden knife.

A character is a run of these movements in a fixed order. Its force reaches as far as the writer's breath and travels in the direction of its strokes. The end of a stroke decides what it does: taken to a point like a hanging needle, the force leaves the writer; rounded off like a drop of dew, it stays. Su fights in regular script, each stroke whole and in sequence, so an opponent who knows the character can tell where the next stroke will go. Bai's sword moves like cursive, which joins strokes and breaks their order.

\textbf{zhi --- stop:} two verticals and two horizontals; a short barrier, sustained by the writer's planted stance and breath. \textbf{feng --- edge:} a cutting force directed through the last descending line. \textbf{jing --- still:} a slower form that settles vibration when the writer can hold the line long enough. Each form reaches only so far, lasts only as long as the writer's breath and costs him strength; none of them takes away the opponent's freedom to move and choose.

\textbf{fang --- release:} the unfinished work on Mo's old silk. Its left half, fang (square), is written. The right half, pu (striking hand), has four strokes and ends with the golden knife. What the completed character does is found out in the course of the play.
\end{quote}
\medskip
\begin{quote}
\small
\textbf{Act I, Scene 1. Morning in the Bamboo Grove (selected passage)}

\emph{Mo lifts the cloth from the old silk. On it is fang (square), the left half of fang (release), in dark, dry ink. The right half has never been written.}

\textbf{SU}
Do you want me to try again?

\textbf{MO}
Do you want to?

\emph{Su lays a fresh practice sheet beside the silk, copies the left component, then writes the four strokes of pu (striking hand). Fang (release) fills the screen, complete and ordinary, and nothing on the stage moves.}

\textbf{SU}
There. I can write the word. I've written it a thousand times.

\textbf{MO}
I know.

\textbf{SU}
Tell me what you are waiting to see.

\emph{Mo takes the finished practice sheet, folds it around the damp rag and puts it aside.}

\textbf{MO}
This old silk was meant for an archway. A patron asked me to write ``release'' above a gate to his estate. He said debtors would go free beneath it.

\textbf{SU}
And they didn't?

\textbf{MO}
Some did. They praised him. I started the inscription. Then a woman asked him to release her son from the guards. He said the boy had stolen a knife. I believed him. The woman came up the steps; his guards came after her. I wrote zhi (stop) across the gate.

\emph{He traces the shape on the tabletop with a fingertip.}

\textbf{MO}
It held. She couldn't get out. They took her away while I finished telling them to be gentle.

\textbf{SU}
You tried to stop a fight.

\textbf{MO}
I chose the doorway she needed.

\emph{Su looks at the silk.}

\textbf{SU}
Is that why you never finished?

\textbf{MO}
For a while I thought I could write a form that would know better than I did. I kept trying. People came to admire my silence.

\emph{He nods toward the mound of brush handles.}

\textbf{MO}
The brush I used at that gate is in there. Every other brush in that mound wore out first. I buried that one with its hair on.

\textbf{SU}
You should have told me.

\textbf{MO}
Yes.

\emph{Outside, a vendor calls faintly from the road. Su rolls up the contest notice.}

\textbf{SU}
I'll speak to Liangzhou. And the administrator. I'll come back before the broth.

\textbf{MO}
She makes enough for two.

\emph{Mo takes the long brush from the rack and holds it out. The handle is worn pale where his thumb rested.}

\textbf{SU}
I have a brush.

\textbf{MO}
This one has reach. Take it.

\emph{Su accepts it, examines the binding below the bristles and secures the brush in his sash.}

\textbf{SU}
If he wants a display, he can watch me write a notice withdrawing my name.

\textbf{MO}
Try speaking first.

\emph{Su bows and leaves. Mo starts to cover the old silk, changes his mind, and leaves it open. He takes the rag wrapped in Su's completed fang (release) and begins drying the table.}
\end{quote}
\medskip
\begin{quote}
\small
\textbf{Act I, Scene 2. The Marketplace of Lin'an (selected passage)}

\textbf{BAI}
He'll find you a stool among the scholars. You can explain my sword while I use it.

\emph{Su starts to rise. A bowstring snaps, his tea bowl bursts, and a dark arrow stands in the post behind it.}

\emph{The crowd scatters under the carts. BAI draws his sword. Above the market, on a narrow roof platform, stands YAN WUJIU, a shallow bowl of tea balanced on the forearm that holds his bow. He draws a second arrow from his quiver, leaving five there, one of them white.}

\textbf{YAN}
Stay clear of the table.

\emph{Mother Chen is crouched behind it. Su moves between her and the roof.}

\textbf{SU}
She is clear now.

\textbf{YAN}
Step into the lane.

\textbf{BAI}
It's a test. Don't give him the ground.

\emph{Su looks up past the archer toward a hanging sign and, beyond it, the place where Han has just disappeared.}

\textbf{SU}
For whose benefit?

\emph{Yan waits. A porter cut off from the crowd edges toward the shelter of a cart; Yan follows the man with his eyes and keeps his point lowered until the porter is under cover.}

\textbf{YAN}
Yours, if you learn.

\emph{He releases. Su cannot step aside; Mother Chen is crouched at his back. He meets the shaft with feng (edge), driving the brush down through its last stroke, and the character flares across the street and turns the arrow from his chest. In the same movement he tries to send the cutting stroke back up the roof.}

\emph{Bai sees the silk merchant getting up behind Su and pulls the man down. Su catches the movement, twists his brush and forces the stroke toward an empty cart. The arrow goes with it and buries itself in the cart, and a bolt of silk unrolls in two severed lengths. Su's wrist gives, and he nearly drops the brush.}

\emph{With another shaft half drawn from the quiver, Yan watches Su recover and scans the crowded lane before putting the arrow back. He lifts the bowl from his forearm and drinks; none of the tea has spilled.}

\textbf{BAI}
Come down here!

\textbf{YAN}
The roof holds me well enough.

\textbf{BAI}
Then stay for the next one.

\emph{Bai steps onto the tea table. Su catches his coat.}

\textbf{SU}
Mother Chen is underneath.

\emph{Bai looks down and jumps off beside the table instead. Yan retreats along the roof and is gone.}

\emph{Su opens his right hand and cannot close it again without pain. Bai takes a strip from the cut silk, then looks at its owner.}

\textbf{BAI}
Put this on my account.

\textbf{SILK MERCHANT}
You haven't got an account.

\emph{Bai takes out Han's purse and throws it to him.}

\textbf{BAI}
Apparently I have now.

\emph{He wraps Su's wrist.}

\textbf{SU}
You know him.

\textbf{BAI}
Yan Wujiu. They call him Still-Water Yan. He shoots with a bowl of tea on his arm and drinks it afterward. He waits longer than most men can bear. Never chase an arrow back to him. By then he's chosen where you'll put your foot.

\textbf{SU}
Han hired him?

\textbf{BAI}
I didn't see the payment.

\textbf{SU}
You recognized the test.

\textbf{BAI}
I've had one. So has every man paid to stand beside a lord. He wants to know what he bought.

\textbf{SU}
He hasn't bought me.

\emph{Su winces as Bai pulls the bandage tighter than necessary.}

\textbf{BAI}
Hold that end.

\emph{Su holds it while Bai ties the knot. Bai leaves the purse with the merchant and goes. The Storyteller helps Mother Chen to her feet.}

\textbf{MOTHER CHEN}
He said he wouldn't interfere with trade.

\textbf{STORYTELLER}
He didn't break the bowl himself.

\emph{The merchant counts coins against the ruined silk. Su goes to him.}

\textbf{SU}
Is it enough?

\textbf{SILK MERCHANT}
For the cloth. The cart is another matter.

\emph{Su puts his paper money beside the purse and sets his six new sheets on the stall.}

\textbf{SU}
Keep these until I can pay again.

\textbf{STALLKEEPER}
Take them. Tell Master Mo I want his next couplet.

\emph{Su takes the paper. As the market resumes, the Storyteller climbs back onto her platform and knocks her clappers together twice.}

\textbf{STORYTELLER}
Two arrows in a morning market! \\
One took a bowl, one took a bolt. \\
The bowl belongs to Mother Chen; \\
The silk is Zhang's, and Zhang wants payment. \\
Who owns the man upon the roof?

\emph{She holds out her fan to the listeners for the answering line, as tellers do, but no one supplies it. A guard's black sleeve passes at the edge of the crowd.}

\emph{Su heads toward the grove. The merchants right the tea table beneath the fading city light.}

---
\end{quote}
\medskip
\begin{quote}
\small
\textbf{Act III, Scene 5. Where the Brush Meets the Blade (climax)}

\emph{Su unwinds the outer bandage and binds it higher over the cut. He draws the silk from his robe, removes the folded strip and tucks it deep inside his robe. Bai lowers his blade enough to see the silk.}

\textbf{BAI}
I told you to leave that until tomorrow.

\textbf{SU}
Will you give me time to wet the brush?

\emph{Bai looks at the sword in his hand, then at the older swordsman.}

\textbf{OLDER SWORDSMAN}
You can allow it.

\textbf{BAI}
Wet it, then.

\emph{Su goes to the case. The aide opens it, lifts out the inkstone and shuts the lid again. Su loads the long brush, then stretches the old silk over the flat lid, with fang (square) to the left and the unwritten space to the right. The camera enlarges the dry half-character above them.}

\textbf{HAN}
My case is at your service.

\emph{Su pushes it away from Han's side of the platform toward the neutral space beneath the pine. The aide steps back.}

\emph{Su writes the first stroke of pu (striking hand) on the silk, and a thin line extends from the image toward Bai's sword arm. Bai feels it pull and moves to test it. Su follows with the second stroke; the line tightens and draws Bai's blade toward the case.}

\emph{Su bears down, and the force comes back through the brush into his bandaged wrist until the bandage darkens.}

\textbf{BAI}
Is this the part where I stand still for you?

\emph{Bai lets the pull take him a step, then drives forward with it, and the line Su has drawn brings the blade straight at Su. Su pulls the brush clear and breaks the form. Bai's blade stops at the writing case and splits a corner of its lacquer lid.}

\emph{The silk slips, and Su catches it before it falls. The screen holds the two new strokes, cut off from the old half.}

\textbf{BAI}
You won't get the sword out of me that way.

\textbf{SU}
I see that.

\emph{Su sets the silk back down and takes his stance between the case and Bai, his brush shaking.}

\emph{Han closes his fan and looks toward Yan. The archer takes out the white shaft and nocks it without drawing.}

\textbf{HAN}
Master Bai. They have seen enough of the grove's unfinished work.

\emph{Bai ignores him. He salutes Su once more, and this time he names the cut before he makes it, as Pei used to before the last bout of the day.}

\textbf{BAI}
Clouds Break, Moon Rises.

\emph{He takes the narrow length of the platform. The blade circles high and parts Su's guard, then drops and comes up from below. Su's stop can catch it only by holding ground near the broken case. Su writes zhi (stop) in the air, all four strokes this time, and Bai meets it with his whole weight, locking blade and luminous line above the silk. On the screen the slant of the sword stands beside the stop.}

\textbf{OLD SCHOLAR}
Stop, and a dagger-axe beside it. Wu (martial).

\emph{The young recorder writes it down.}

\emph{As Bai presses and Su's breath runs out, the bottom of the projected zhi (stop) flickers. Beyond Bai, the narrow stair and Han's men lie directly along the path of a returning feng (edge).}

\emph{Su glances at them and begins the edge anyway, gathering the weight through his shoulder. Bai sees the angle and braces to turn it. On his rock above the pine, Yan brings the white arrow to half draw.}

\emph{Su stops before the last stroke of feng (edge) and lowers the brush to the old silk instead.}

\emph{The last two strokes of pu (striking hand) leave the brush as one movement, the whole stalk at once, and Bai, who has read every character Su has made since they were boys, has no order to read. The third stroke runs under the pressure of the blade, and Su turns his whole body with it, giving up the center. The fourth, the golden knife, runs out to a needle's point in the empty air beside the pine. The completed fang (release) spreads across the screen.}

\emph{Su lets his stop go. The force held between sword and brush rushes out the way the golden knife points, and a long cloth stroke flies outward over the drop. Bai, still driving forward, is carried after it. Su goes with him instead of trying to hold the form, and the brush tears out of his hand.}

\emph{Bai's front foot leaves the platform. For an instant his whole weight hangs over nothing.}

\emph{Su catches his coat with both hands, but the injured arm gives at once. He falls to one knee, hooks his sound arm beneath Bai's and stops him on the lower ledge. Bai's sword strikes the edge of the platform; he can haul himself up only by letting it go.}

\emph{He opens his hand, and the sword falls into the cloth below.}

\emph{Su pulls, and Bai climbs against him until both of them are back on the platform. They lie beside the case, alive and breathing hard, as the completed character fades above them.}

\emph{Han is on his feet.}

\textbf{HAN}
Archer. Bai has lowered his weapon.

\emph{Yan has the white arrow at full draw, aimed toward Bai. The witnesses turn toward the rock.}

\emph{Su sees the point. His brush lies out of reach; he tries to rise between the arrow and Bai, but his arm will not hold him.}

\emph{Bai gets up first, pulls Su upright and puts himself squarely between the arrow and his brother.}

\textbf{BAI}
I yield.

\emph{Yan holds at full draw.}

\textbf{HAN}
You heard him. Release.

\textbf{YAN}
I heard him yield.

\textbf{HAN}
You have your instruction.

\textbf{YAN}
He has no sword.

\textbf{HAN}
He has deceived the gathering. They arranged this on the path. They mean to mock us before the schools.

\emph{The older swordsman steps onto the first stair to the arena.}

\textbf{OLDER SWORDSMAN}
I accepted the terms. So did you. Once a contest is decided, no hand is raised again. You said it yourself, an hour ago.

\emph{Yan lets the bow down without releasing and keeps the arrow in his hand.}

\textbf{HAN}
Your fee bought an arrow.

\textbf{YAN}
It bought a shot at a man breaking those terms. Find him.
\end{quote}
\medskip
\begin{quote}
\small
\textbf{Scene 5, continued (the white arrow and the finished silk)}

\emph{Yan descends from the rock and slides the white arrow back among the four dark shafts while Bai watches him.}

\textbf{BAI}
Who paints a last arrow white? Do you want us all to count it?

\textbf{YAN}
My teacher put a bowl of water on my arm until I could shoot without spilling it. Then he walked me to a drop like that one, stood with his heels over the edge and told me to shoot. I lay flat on the rock with the sweat running into my boots.

\textbf{BAI}
And?

\textbf{YAN}
He said I had learned the archery that shoots.

\emph{He touches the white shaft.}

\textbf{YAN}
This one is for the other kind.

\textbf{BAI}
That's all?

\textbf{YAN}
A shaft that straight is expensive.

\emph{Yan starts toward the broad stair.}

\textbf{SU}
The market. Why the bowl first?

\textbf{YAN}
You were beside the keeper. I needed you in the lane.

\textbf{SU}
You could have asked.

\textbf{YAN}
You could have stayed under the table.

\emph{He goes. Bai waits until he is off the platform, then lets out a breath and sits heavily beside Su.}

\textbf{BAI}
I liked him better on a roof far away.

\emph{Su unfolds the silk. Its right-hand strokes differ from Mo's old work: the first two dark and cramped, the last two run together in one long cursive pull, drawn out by the falling weight.}

\textbf{BAI}
That isn't his hand.

\textbf{SU}
No. It's the way you swung a stick in Pei's yard. He spent six years trying to get it out of my wrist.

\textbf{BAI}
I can barely write my own name.

\textbf{SU}
It didn't stop you teaching.

\textbf{BAI}
So that's the word the old man couldn't finish.

\textbf{SU}
He kept trying to make it take the sword out of someone else's hand. It only lets go of your own.

\textbf{BAI}
Mine went over the edge as well.

\textbf{SU}
You opened your hand. I saw you do it.

\emph{He folds the silk.}
\end{quote}
\medskip
\begin{quote}
\small
\textbf{Scene 5, continued (ending)}

\emph{The STORYTELLER takes her place on it. The sound of the city returns: bowls set out, a cart wheel, people asking who won.}

\textbf{VOICE IN THE MARKET}
Which of them was left standing?

\textbf{STORYTELLER}
By the end, both. With some assistance.

\textbf{ANOTHER VOICE}
Then which was the master?

\textbf{STORYTELLER}
If I knew that, I'd put it on the bill and charge you twice.

\textbf{A THIRD VOICE}
Whose name are they cutting in the cliff?

\textbf{STORYTELLER}
Nobody's. The mason packed his chisels and came down for his supper.

\emph{She opens her fan. Behind her, in the dim grove where the lamp still burns beside Mo, Su and Bai come in at the open gate. Bai stops in the gateway with his recovered, dirty sword, and Su waits beside him.}

\textbf{STORYTELLER}
At Cui Feng Peak the pine has kept \\
A scrap of white upon its root. \\
The cliff has kept its empty stone. \\
The road is steep. Two men come down; \\
One takes the other's weight.

\emph{Bai lays the sword beside the bench. Su takes the folded strip from his robe and holds it out. Bai opens it and reads slowly, one character at a time, his lips moving.}

\textbf{BAI}
Bai. Liang. Zhou.

\textbf{SU}
His last inscription. He said you were his last student.

\emph{Bai looks at the bench, then back at the gate he has just come through.}

\textbf{BAI}
He could have said it at the gate.

\textbf{SU}
He was afraid to. Last night he held his arm in a bucket of well water until it stopped shaking, so he could write it.

\emph{Bai kneels, and Su kneels with him. Bai lays the strip on the bench beside Mo; Su draws the split long brush from his sash and pushes it upright into the mound among the others.}

\emph{The Storyteller strikes her clappers once.}

\textbf{STORYTELLER}
Honored listeners, here the tale is told. The plate comes round.

\emph{She closes the fan as the grove light settles on the three figures. The screen remains blank.}

\emph{Blackout.}

\textbf{END}
\end{quote}

\emph{Commentary.} The script fixes the rules of its brush combat before the first fight: a character's force travels in the direction of its strokes, a stroke taken to a needle's point sends that force away from the writer, and Su's regular script lets an opponent who knows the character read the next stroke. Mo's story in Scene~1 attaches a moral cost to these rules, since his \emph{zhi} (stop) across the gate closed ``the doorway she needed.'' The climax resolves the duel within them: with Han's men on the return path of \emph{feng}, Su stops before its last stroke, writes the last two strokes of \emph{fang} as one movement that leaves Bai no order to read, and points the final stroke into the empty air beside the pine, so the held force leaves through an open direction. The release costs both men their grip, as the brush tears out of Su's hand and Bai opens his hand to let the sword fall, which gives Su's later remark that the character ``only lets go of your own'' a referent the audience has just watched.

\paragraph{Query 196 extended response excerpt.}
\emph{Absurdist campus play: the procedure that exchanges the seniors' clothes for a standard shell, the graduation speech that collapses into a children's rhyme, and the freshmen's replay of the opening;} \path{196/script.md}.
\begin{quote}
\small
\textbf{Act I, Scene 1}

In the dark, Mr.~System's voice.

\textsc{Mr.~System.} Next.

The basement of the student activity center: the drama club's rehearsal room. A ring of fill lights comes up in a corner; Zhou Xiaobo stands before the lights, livestreaming to a phone on a tripod. He is in his patchwork jacket, with five badges pinned on the chest.

\textsc{Zhou Xiaobo.} (waving extravagantly) Fam---i---ly---! Xiaobo's back! First day of the last semester of senior year, and here's the big one: before graduation, I'm shooting a documentary that truly belongs to this generation! (mimes shouldering a film camera) I've already got the title. It's a secret for now.

[\ldots]

\textsc{Zhou Xiaobo.} (sitting down, picking up his cup) Ah, ten o'clock. Tonight we ate spicy hotpot, fried chicken, half a portion of cold noodles, and drank four milk teas: one full sugar, one no ice, one less sugar, one cold brew. We ate very well tonight. This is because we live at Z University, and we are seniors.

[\ldots]

The loudspeaker on the wall clicks.

\textsc{Broadcast.} Dear students, the spring campus recruitment fair opens in the gymnasium at nine tomorrow morning. All seniors please arrive on time and complete the entry procedure as prompted. Also: with the approval of the pricing authority, undergraduate tuition at this university will rise by thirty percent next academic year, to further improve the quality of talent cultivation.

\textsc{Zhou Xiaobo.} (to the loudspeaker) Leading by a mile.

\textsc{Broadcast.} (the voice slowly becomes Mr.~System's) Next.

The loudspeaker falls silent. Lin Zhiyuan turns his milk-tea cup half a turn.

\textsc{Lin Zhiyuan.} (quietly) Tell me \ldots\ when we go to the recruitment fair tomorrow, who exactly are we going there to become?

Every hand stops. Su Meng's hands freeze mid-stroke with the hand cream. Chen Mo stares at her thermos. Zhou Xiaobo opens his mouth and closes it again. Silence.

\textsc{Zhou Xiaobo.} (laughing) Holy crap, a literary youth's opening line.

\textsc{Lin Zhiyuan.} I wasn't trying to---

\textsc{Zhou Xiaobo.} Anyway, we're not going there to become ourselves, that's for sure. Dig-a dig-a dig-a dig---

He makes a clownish digging motion.

\textsc{Su Meng.} (laughing along) Xiaobo, that's called getting into character early.

\textsc{Chen Mo.} (quietly) Makes sense.

Lin Zhiyuan does not laugh. He closes his book.

\textsc{Zhou Xiaobo.} (raising his cup) Come on! To tomorrow, when we---(glancing at Lin Zhiyuan)---become someone else---cheers!

Four milk-tea cups knock together with a brittle plastic click. Lin Zhiyuan's cup is the last to be drawn back. Lights fade. The loudspeaker repeats, very softly: ``\ldots please complete as prompted \ldots\ please complete as prompted \ldots''
\end{quote}
\medskip
\begin{quote}
\small
\textbf{Act I, Scene 3}

Twelve noon. The center of the gymnasium. Two rows of chairs: six gray folding chairs in front, and in back four chairs, the same four from the rehearsal room. On each chair lies a folded gray suit-raincoat. The Chorus tear off white tape and mark a square on the floor in front of each chair, ten in all, then stand before the front-row chairs. The four stand before the back-row chairs. Cold white light. A bare synthesized tone, like a computer alert.

\textsc{Mr.~System.} Final entry procedure. Please complete as prompted. Step one: remove your student outerwear.

Chorus One immediately takes off her gray T-shirt (underneath is an identical one), folds it, and lays it on the chair. Chorus Two through Six follow in sync, at the same speed.

\textsc{Mr.~System.} Please complete as prompted.

Lin Zhiyuan watches them. He has worn this white shirt for three days.

\textsc{Lin Zhiyuan.} (to himself) They say Kong Yiji could never take off his long gown.

He starts undoing the buttons slowly from the first one; after each button he lines the two front panels up again.

\textsc{Lin Zhiyuan.} I can take mine off.

He takes off the shirt and folds it into a square block, his fingertips barely trembling. Su Meng watches him and her eyebrows lift.

\textsc{Su Meng.} (barely audible) He really \ldots

She takes off her blazer too, quickly, and folds it neatly. Chen Mo stands still, looking down at her hoodie.

\textsc{Chorus.} Please complete as prompted. Please complete as prompted. Please complete as prompted.

Chen Mo closes her eyes, takes off the hoodie, and folds it. Her eyes stay closed.

\textsc{Zhou Xiaobo.} (looking around, a dry laugh) Family, this is the server-launch ceremony for us work-mules---(no reaction from the Chorus) Dig-a dig-a dig---

\textsc{Chorus.} (in unison, mechanically taking it up) ---plant tiny little seeds, grow tiny little flowers---

Zhou Xiaobo freezes.

\textsc{Zhou Xiaobo.} How do you know---

\textsc{Mr.~System.} Please complete as prompted.

Zhou Xiaobo looks down at his jacket; his fingers find the badge on his chest that reads ``I Won't Go Ashore'' and press on it for a moment.

\textsc{Zhou Xiaobo.} (quietly) Just for a minute. It's not like I'm never wearing it again.

Looking away, he takes off the jacket, folds it, and lays it on the chair; its corners do not line up.

\textsc{Mr.~System.} Step two: put on your standard shell.

The Chorus put on the gray raincoats in sync. Dressed, the six have the same shoulder line, the same length, the same gray; there is no telling who is who. Su Meng puts hers on and adapts fast, straightening the collar as she goes. Lin Zhiyuan puts his on and tugs the cuffs flat. Chen Mo puts hers on; the raincoat is too big, and she shrinks a little inside it. Zhou Xiaobo puts his on and pulls a face at an invisible mirror in the air. The face does not hold; it stiffens where it is.

\textsc{Mr.~System.} Step three. (Pause.) Step into the square.

The Chorus step forward together, each into a square. The four look down at the squares on the floor.

\textsc{Chorus.} Step in. Step in. Step in.

Su Meng steps into her square first. Lin Zhiyuan follows close behind, his foot landing right inside the square. Chen Mo steps in with her eyes closed. Zhou Xiaobo is last. He looks down at his feet.

\textsc{Zhou Xiaobo.} (quietly) That's it?

He steps in.

\textsc{Mr.~System.} Entry procedure complete. Welcome, standard applicants of this season.

\textsc{Chorus.} Thank you for your cooperation. We wish you every success.

All the lights snap on at once, glaringly bright; everyone squints. Lin Zhiyuan's hand touches something hard in his raincoat pocket. He takes it out: his fountain pen. He does not remember putting it there. He puts the pen back; his hand grips it inside the pocket and does not come out.

\textsc{Mr.~System.} Please proceed to your presentation venues. (Pause.) Next.

The four walk off in four directions. The Chorus march off in step. Only the two rows of chairs remain. The last light rests on a back-row chair, falling on the patchwork jacket, its corners not lined up. A badge glints in the white light.

Just as the lights are about to go out, Zhou Xiaobo runs back on from the wings, snatches up the jacket, stuffs it inside his raincoat, and runs off again. Blackout.
\end{quote}
\medskip
\begin{quote}
\small
\textbf{Act III, Scene 14}

\textsc{Mr.~System.} The final item. The graduate representative will now speak.

\textsc{Chorus.} (in unison) Will---Zhou Xiaobo---come to the stage---

Lin Zhiyuan, Su Meng, and Chen Mo all look up, but Zhou Xiaobo stays seated, both hands pressed on his knees, his fingers trembling.

\textsc{Chorus.} Zhou Xiaobo---to the stage---

He stands and walks to the rostrum. The two sleeves of the empty suit hang from the chair back. Chorus Six hands him a microphone and a card.

\textsc{Chorus Six.} Your script. Just read it.

Zhou Xiaobo looks at the card. He raises the microphone to his mouth. Silence. He opens his mouth; no sound. He tries again; a faint sound in his throat, as if something were stuck there. Again: still no words, though the shape of his open mouth is perfectly standard. Silence. Lin Zhiyuan leans forward in his seat, listening. Su Meng's fingertips tremble on her knees. Chen Mo keeps watching Zhou Xiaobo's face.

\textsc{Zhou Xiaobo.} (from the throat) \ldots Dig.

The sound comes out.

\textsc{Zhou Xiaobo.} Dig-a \ldots

\textsc{Chorus.} (mechanically, in unison, taking it up) \ldots dig-a dig.

\textsc{Zhou Xiaobo.} Dig-a dig-a \ldots

\textsc{Chorus.} \ldots plant tiny little seeds \ldots

Zhou Xiaobo wants to say something else. Only this sound will come out.

\textsc{Zhou Xiaobo.} Dig-a dig-a dig-a dig---

\textsc{Chorus.} \ldots grow tiny little flowers.

\textsc{Zhou Xiaobo.} (into the microphone, louder and louder) Dig-a dig-a dig-a dig-a dig-a dig-a---

He stops. The microphone slips from his hand and hits the table with a thud, which the speakers make enormous.

[\ldots]

\textsc{Mr.~System.} (calmly) Thank you for your excellent speech. (Pause.) Next.

Everything resumes. The Chorus sit back down. Chorus One comes up, takes the microphone from the table, and gently turns Zhou Xiaobo toward the seats. Zhou Xiaobo walks back and sits down. He does not cry.

[\ldots]

Chen Mo rises, the chair on her back. Passing Zhou Xiaobo's seat, she stops.

\textsc{Chen Mo.} (very faintly) Why didn't you tell any jokes today?

Zhou Xiaobo looks up at her. He opens his mouth; it moves twice, without a sound. Chen Mo stands there for a while, then begins to tell her first joke in four years.

\textsc{Chen Mo.} (very earnestly, very clumsily) Heisenberg gets pulled over for speeding. The officer asks: Do you know how fast you were going? Heisenberg says: No. But I know where I am.

Zhou Xiaobo looks at her. He does not laugh.

\textsc{Chen Mo.} (nodding) \ldots Not funny. I know.

She turns and walks out with the chair on her back. Zhou Xiaobo sits alone in the follow spot. After a long time, his shoulders move.

\textsc{Zhou Xiaobo.} (very softly) \ldots Pfft.

He laughs, but by now no one left on stage can hear it. He stands and walks out. The stage is empty. Blackout.
\end{quote}
\medskip
\begin{quote}
\small
\textbf{Epilogue}

September. The rehearsal room. One chair, with the patchwork jacket hanging on its back. Where the badges were, only a few small pinholes remain. A top light. From the corridor, the rumble of suitcase wheels, voices, laughter. On the door, a freshly printed poster: ``Z University Drama Club: Now Recruiting.'' Four freshmen (Chorus One to Four, in brightly colored T-shirts, pulling suitcases) push the door open and come in.

\textsc{Freshman A.} This is the drama club?

\textsc{Freshman B.} Just one table and four chairs?

\textsc{Freshman C.} (seeing the jacket) Whose is this?

\textsc{Freshman D.} No idea. A senior said it's been hanging here for months. Nobody ever came for it.

Freshman C takes the jacket off the chair back, shakes it out, and puts it on. It fits exactly. He goes to the corner, sets his phone on the tripod, and switches on the fill light.

\textsc{Freshman C.} (waving extravagantly) Fam---i---ly---!

The other three sit at the table: one leafs through a book, one rubs in hand cream, one memorizes vocabulary from a phone. Freshman A picks up a milk tea.

\textsc{Freshman A.} Ah, ten o'clock. Tonight we ate spicy hotpot, fried chicken, half a portion of cold noodles, and drank four milk teas: one full sugar, one no ice, one less sugar, one cold brew. We ate very well tonight. This is because we live at Z University, and we are freshmen.

\textsc{Freshman B.} (keeps leafing through the book; turns a page)

\textsc{Freshman D.} (lips barely moving) Ubiquitous: present everywhere.

The bell in the corridor rings seventeen times, the way university bells do. The loudspeaker on the wall clicks.

\textsc{Mr.~System.} Welcome, new students. Please complete as prompted.

Freshman C flashes a V sign at the phone camera.

\textsc{Freshman C.} Family, I'm going to shoot a documentary, one that truly belongs to this generation---

\textsc{Mr.~System.} Next.

Blackout.

(The End)
\end{quote}

\emph{Commentary.} The play carries its argument through repeated material whose owner changes. In Act I, Scene 3, the children's rhyme is Zhou Xiaobo's joke until the Chorus take it up mechanically; in Act III, Scene 14, it is the only sound he can produce at the microphone, and Mr.~System thanks him for an ``excellent speech'' and says ``Next.'' Chen Mo's Heisenberg joke, told after the ceremony to a classmate who has lost his voice, answers that procedure with a personal act, and Zhou's laugh comes only after a long time, when no one left on stage can hear it. The epilogue then repeats Scene 1's livestream greeting and dinner speech almost word for word, with ``seniors'' changed to ``freshmen'' and a newcomer wearing Zhou's jacket, now stripped of its badges, so Mr.~System's closing ``Next'' places the four seniors' year inside a recurring cycle.

\section{Serial-Fiction Extension}
\label{app:serial}

Serial writing and serial distillation extend the harness of \S\ref{sec:harness} to open-ended novels. In this setting, the work persists across sessions and each chapter advances the project state.

\paragraph{Workspace.}
The workspace separates persistent series design (the series bible), current chapter work, and append-only published text. Each chapter uses the complete pipeline of Appendix~\ref{app:pipeline}, and a manifest records the published sequence. Publication proceeds through gate precheck, ledger promotion, file copying, and manifest update as an idempotent transaction. If a session reopens mid-chapter, a recovery matrix identifies where to resume.

\paragraph{Theory remapped onto serial units.}
Serial writing retains the scene-level value shift of \S\ref{sec:schema} and organizes larger units around chapter publication. A chapter card specifies a closing hook through suspense, reversal, emotional charge, or an information drop; review checks the chapter ending against this design. Three to five chapters form a small three-act arc, with complete arc closure at volume level. The top-level spine records the story's direction and the limits of its world without fixing a distant climax. Only the current volume is materialized, so later volumes can be added as the work develops.

\paragraph{Persistent state in three ledgers.}
Three append-only ledgers preserve state across chapters; they are supplied before writing and validated afterward. The \emph{threads} ledger tracks foreshadowing, promises, and mysteries through open, advance, and payoff events, each with an intended horizon; overdue entries identify abandoned threads. The \emph{world-facts} ledger stores typed entity--attribute--value facts and records which facts supersede earlier ones. It also records ability limitations and who knows each secret at chapter $N$; an errata channel applies post-publication corrections only to future chapters. The \emph{biography} ledger records each character's growth across multiple segments and the changes in each volume.

\paragraph{Writer context at chapter $N$.}
Context is more detailed for material closer to chapter $N$: one line per distant volume, unit digests, recent chapter recaps, ledger entries selected by the chapter card, then scene cards and the preceding chapter's tail. Summaries are prepared at publication, keeping per-chapter context assembly cost flat as the work grows.

\paragraph{Self-corpus retrieval.}
When the writer retrieves earlier scenes from the same work, a publication-order filter admits only those at or before the current chapter position. This prevents future material from entering the prompt.

\paragraph{Serial distillation.}
The distillation package analyzes serialized works in publication order because their endings may still be unwritten. It produces either a reference corpus or an importable project state. The reference corpus contains paradigm cards for volume structure, hooks, and pacing, plus a sparse selection of scene excerpts. For continuation, the project state passes through an exchange-format import gate. On a 538-chapter, 1.63-million-character serial, distillation reconstructed six pipeline files, thirty scene slices, five character packages, and the paradigm cards.

\paragraph{Genre as data.}
Seven families of worldbuilding references, from apocalypse survival to xianxia, accompany a profile whose axes cover reader contract, world chassis, narrative engine, serialization topology, state burden, and evidence burden. The workflow remains shared across genres; profiles, references, and state modules carry the differences.

\section{Ethics and Disclosure}
\label{app:ethics}

\paragraph{Risks.}
Risks include reproducing copyrighted style and harmful stereotyping through persona files. Verbatim reuse of canon text requires an explicit directive, and a verifier checks the writer's reuse receipt against the story (Appendix~\ref{app:reference-machinery}). Canon excerpts and persona assertions carry line-anchored source locators (Appendix~\ref{app:canon-corpus}).

\paragraph{AI-assisted writing.}
The research ideas, manuscript outline, and key arguments were developed by the human authors. AI tools assisted with manuscript editing.

\end{document}